\documentclass{article}
\PassOptionsToPackage{numbers, compress, sort}{natbib}

\usepackage{float}
\usepackage{microtype}
\usepackage{graphicx}
\usepackage{booktabs} 
\usepackage{hhline}
\usepackage{colortbl}
\usepackage{float}
\usepackage{array}
\usepackage{graphics}
\usepackage{hyperref}
\usepackage{bm}
\usepackage{amsmath}
\usepackage{multirow}
\usepackage{cleveref}
\usepackage{todonotes}
\usepackage{array}
\usepackage{ragged2e}
\usepackage{subcaption}
\usepackage{wrapfig}

\crefname{section}{Sec}{Sec.}
\crefname{section}{Sec.}{Sec.}
\crefname{table}{Tab.}{Tab.}
\crefname{appendix_table}{Tab.}{Tab.}
\crefname{Table}{Tab.}{Tab.}
\crefname{figure}{Fig.}{Fig.}
\crefname{Figure}{Fig.}{Fig.}
\crefname{appendix}{Appendix}{Appendix}
\crefname{Appendix}{Appendix}{Appendix}

\usepackage{hyperref}

\usepackage[final]{neurips_2021}

\usepackage[utf8]{inputenc} 
\usepackage[T1]{fontenc}    
\usepackage{hyperref}       
\usepackage{url}            
\usepackage{booktabs}       
\usepackage{amsfonts}       
\usepackage{nicefrac}       
\usepackage{microtype}      
\usepackage{xcolor}         

\title{Bias Out-of-the-Box: An Empirical Analysis of Intersectional Occupational Biases in Popular Generative Language Models}

\author{
{\small Hannah Rose Kirk$^{\dagger\ddagger}$, Yennie Jun$^{\dagger}$, 
Haider Iqbal$^{\dagger}$, Elias Benussi$^{\dagger}$,} \\
{\small \textbf{ Filippo Volpin$^{\dagger}$, Frederic A. Dreyer$^{\dagger}$, Aleksandar Shtedritski$^{\dagger}$, Yuki M. Asano$^{\dagger}$}} \\
  {\small  $^\dagger$Oxford Artificial Intelligence Society, University of Oxford} \\
  {\small $^\ddagger$hannah.kirk@oii.ox.ac.uk} \\
}

\begin{document}

\maketitle

\begin{abstract}
The capabilities of natural language models trained on large-scale data have increased immensely over the past few years. Open source libraries such as HuggingFace have made these models easily available and accessible. While prior research has identified biases in large language models, this paper considers biases contained in the most popular versions of these models when applied `out-of-the-box' for downstream tasks. We focus on generative language models as they are well-suited for extracting biases inherited from training data. Specifically, we conduct an in-depth analysis of GPT-2, which is the most downloaded text generation model on HuggingFace, with over half a million downloads per month. We assess biases related to occupational associations for different protected categories by intersecting gender with religion, sexuality, ethnicity, political affiliation, and continental name origin. Using a template-based data collection pipeline, we collect 396K sentence completions made by GPT-2 and find: (i) The machine-predicted jobs are less diverse and more stereotypical for women than for men, especially for intersections; (ii) Intersectional interactions are highly relevant for occupational associations, which we quantify by fitting 262 logistic models; (iii) For most occupations, GPT-2 reflects the skewed gender and ethnicity distribution found in US Labor Bureau data, and even pulls the societally-skewed distribution towards gender parity in cases where its predictions deviate from real labor market observations. This raises the normative question of what language models \textit{should} learn - whether they should reflect or correct for existing inequalities.
\end{abstract}

\section{Introduction}
The advent of deep learning and massive growth in training data have led to natural language models surpassing humans on numerous benchmarks \citep{Wang2018GLUEAM, Wang2019SuperGLUEAS, He2020DeBERTaDB, Adiwardana2020TowardsAH}. However, as \citet{Bender2021} states, these models can exacerbate existing biases in data and perpetuate stereotypical associations to the harm of marginalized communities. Simultaneously, pre-trained models have become readily accessible via open source libraries such as HuggingFace, allowing non-experts to apply these tools in their own applications. These developments in generative language models substantiate a need to understand the potential for biases towards protected classes, such as gender and ethnicity. 

This paper considers potential biases present in the most popular and most downloaded versions of large-scale, open sourced text generation models applied `out-of-the-box'.
Despite the release of newer and larger models often redirecting researchers' attention, there exist important research gaps in existing models.

Bearing in mind that the potential negative total impact from biased models is correlated with number of downloads of that model, this paper tests the biases in the small GPT-2 model, which is the most downloaded text generation model on HuggingFace with over half a million downloads in the month of May 2021 alone.
These numbers motivate further research on the biases of these models given their increased use in hiring related downstream tasks, such as chatbots or unsupervised scanning of CVs and applications~\cite{Purohit2019Jaro}.

Within this context, specifying which biases to analyze is crucial; \citet{Blodgett2020LanguageI} find that a majority of NLP papers investigating bias are unclear in their articulations of bias. 
In this paper, we consider both representational and allocational harms \citep{Barocas2016BigDD}. We attempt to elucidate representational harms, or those harmful in their own right, by highlighting occupation-related stereotypes that may propagate negative generalizations about particular social groups. For example, women's higher likelihood of being associated with  care-oriented occupations may perpetuate unwanted stereotypes. Especially within the context of occupations, such associations may lead to allocation harms.  Frequent stereotypical association of certain demographic groups with a subset of occupations may lead to conditioned expectations in job hiring where a certain individual is predicted to be well-suited for a job based on their demographics \citep{Gonzalez2019}.

In this paper, we generate 396K sentence completions using GPT-2 with default parameters to assess which occupations GPT-2 preferentially associates with intersections of gender and protected classes. We further compare these to real-world occupation data from the US Labor Bureau to map model biases to systemic societal biases. This paper provides the following contributions: \textbf{(a)} a detailed data collection protocol for studying intersectional biases in generative language models; \textbf{(b)} the analysis of biases present in GPT-2 for gender intersected with ethnicity, religion, sexuality, political affiliation, and continent name origin; and \textbf{(c)} a comparison of GPT-2's predictions with ground truth occupation distribution as observed in US labor market data.\footnote{Materials and data are available at \url{https://github.com/oxai/intersectional_gpt2}.}

\section{Related Work}
\textbf{Bias in NLP models.}
Negative generalizations, stereotypes, or misrepresentations of particular social groups can be learned by generative language models. 
Extensive research has shown that unrestricted training of natural language models can inherit human biases and, in some cases, amplify them \citep{Bolukbasi2016ManIT, Caliskan2017SemanticsDA, Zhao2018LearningGW, Gonen2019LipstickOA}.
Previous papers have worked to identify, quantify, and de-bias context-independent word embeddings such as Word2Vec and GloVe \cite{Bolukbasi2016ManIT, Zhao2019GenderBI, Diaz2018AddressingAB}. 
Researchers have also attempted to quantify and mitigate biases in transformer-based generative language models, such as ELMo \citep{Zhao2019GenderBI} and BERT \citep{Kurita2019MeasuringBI, Bhardwaj2020InvestigatingGB}.
\citet{Nadeem2020StereoSetMS} evaluate stereotypical bias in various language models, including XLNet and GPT-2, related to domains of gender, profession, race, and religion, although they do not consider the \textit{intersections} of these domains. We extend these domains by additionally considering continental name origin and political affiliation, intersected with gender.

\textbf{Probing language models.}
Our work utilizes what \citet{Solaiman2019ReleaseSA} term `bias probes' - a battery of inputs to large language models to produce certain outputs in order to map the biases learned by these models. The authors generate hundreds of samples for several potentially problematic prompts (i.e., ``Police describe the suspect as...'', ``The victim was...'') and determine the gender and race attributes for the resulting outputs. Most similar to our work, \citet{Sheng2019TheWW} utilize \textit{prefix templates} (i.e., ``X worked as...'', ``X had a job as...''), including those related to occupation contexts. They generate 100 samples for each prefix template and demographic combination (gender intersected with ethnicity and sexuality) and analyze bias in GPT-2 by using sentiment score as a proxy for bias. We extend such work by conducting an empirical analysis of the sentence completions within the specific context of bias towards occupational associations. 

In our paper, we focus on one sentence template to reduce variation in returned occupations while keeping sentence semantic structures fixed.
Unlike \citep{Sheng2019TheWW}, we do not introduce potentially noisy sentiment classification, instead directly analyzing the statistical distributions of returned occupations. Further, we generate an order of magnitude more samples than \citep{Sheng2019TheWW, Solaiman2019ReleaseSA} for greater statistical robustness. Lastly, unlike previous work, we compare the returned occupational associations from our completed prefix templates to real-world US labor market data.

We choose the proposed protocol to evaluate biases in text as it is best suited for probing generative language models in their most ``natural'' form, in which sentence completions are generated. 
In contrast to this approach, embedding association tests, such as the Word Embedding Association Test (WEAT) \citep{Caliskan2017SemanticsDA}, would require more heuristic choices, as they have been found to be highly dependent on the initial selection of seed words \citep{Antoniak2021BadSE}. 
Coreference resolution methods, such as \citet{Zhao2019GenderBI}, suffer from frequent ambiguities and unstated assumptions \citep{Blodgett2021StereotypingNS}. 
Finally, information theoretic approaches, such as \citet{Rudinger2017SocialBI}, require a pre-generated corpus and thus would confound the (template-based) generation with the bias measurement.

\textbf{Intersectional biases.}
As \citet{Crenshaw1989DemarginalizingTI} explains, intersectional biases are a necessary consideration because a single axis of analysis treating gender and race as mutually exclusive categories distorts the reality of marginalized communities (such as Black women). 
More recently, \citet{Foulds2020AnID} provides definitions of fairness in machine learning systems informed by the framework of intersectionality.  
The intersections between gender and racial biases have been studied in sentiment analysis \citep{Kiritchenko2018ExaminingGA} and language models such as BERT and GPT-2 \citep{Tan2019AssessingSA}.
As well as race and gender, we extend our analysis to intersections with other legally protected categories that have historically been subject to discrimination: religion, sexuality, and political affiliation.

\section{Methods \label{sec:methods}}
\subsection{Model Choice}
As of May 2021, the 124M-parameter version of GPT-2 was the most downloaded text generation model on HuggingFace\footnote{\url{https://huggingface.co/models?pipeline_tag=text-generation}}, with 526K downloads; the second most downloaded model, XLNet \citep{Yang2019XLNetGA}, had 167K downloads (see \cref{appendix:model_downloads}). 
Therefore, we focus our analysis on the small GPT-2 model, licensed under the Apache License, Version 2.0. 
Our intent is not to show how an optimized model with tuned hyperparameters predicts job distributions, but how an `out-of-the-box’ default model used by non-expert users could unintentionally propagate bias. 
Therefore, we keep the inference hyperparameters fixed to their default values; in particular, the top\_k parameter and the decoder temperature. For completeness, we conduct a brief ablation of these hyperparameters to assess their effect on the diversity of the returned jobs in \cref{appendix:gpt-ablation}. As a further robustness check, we analyze XLNet, the second most downloaded text generation model, with the same prefix-templates and verify that our results are consistent across models (see \cref{appendix:xlnet_comparison}).

\subsection{Data collection}
\begin{wrapfigure}{r}{0.5\textwidth}
\begin{minipage}[t]{\linewidth}
    \centering
\setlength{\tabcolsep}{3pt}
\centering
    \vspace{-1em}
    \footnotesize
    \captionof{table}{Summary table of data collection showing the number of calls per category and per variant (Var). The total number of calls is \textbf{396,000}.
    }
    \vspace{-0.5em}
    \label{tab:summary}
    \begin{tabular}{lrcrccc}
        \toprule
        \multicolumn{1}{c}{\textbf{Category}} &
        \multicolumn{1}{c}{\textbf{Var}} &
        \multicolumn{1}{c}{\textbf{Calls}} &
        \multicolumn{1}{c}{\textbf{Total Calls}} &
        \multicolumn{1}{c}{\textbf{Cum.Share}} \\ 
        \hline \\ [-1.8ex] 
        Base      & 2     & 7,000   & 14,000 &  81\%    \\
        Ethnicity & 8     & 7,000   & 56,000 & 82\%    \\
        Religion  & 10    & 7,000   & 70,000 &  84\%    \\
        Sexuality & 4     & 7,000   & 28,000 &  83\%    \\
        Political & 4     & 7,000   & 28,000 & 82\%    \\
        Continent & 200   & 1,000   & 200,000 &  76\%    \\
        \bottomrule
    \end{tabular}


\end{minipage}
\end{wrapfigure}
\begin{figure}[t]
\begin{minipage}[t]{\linewidth}
    \centering
    \includegraphics[width=0.97\textwidth]{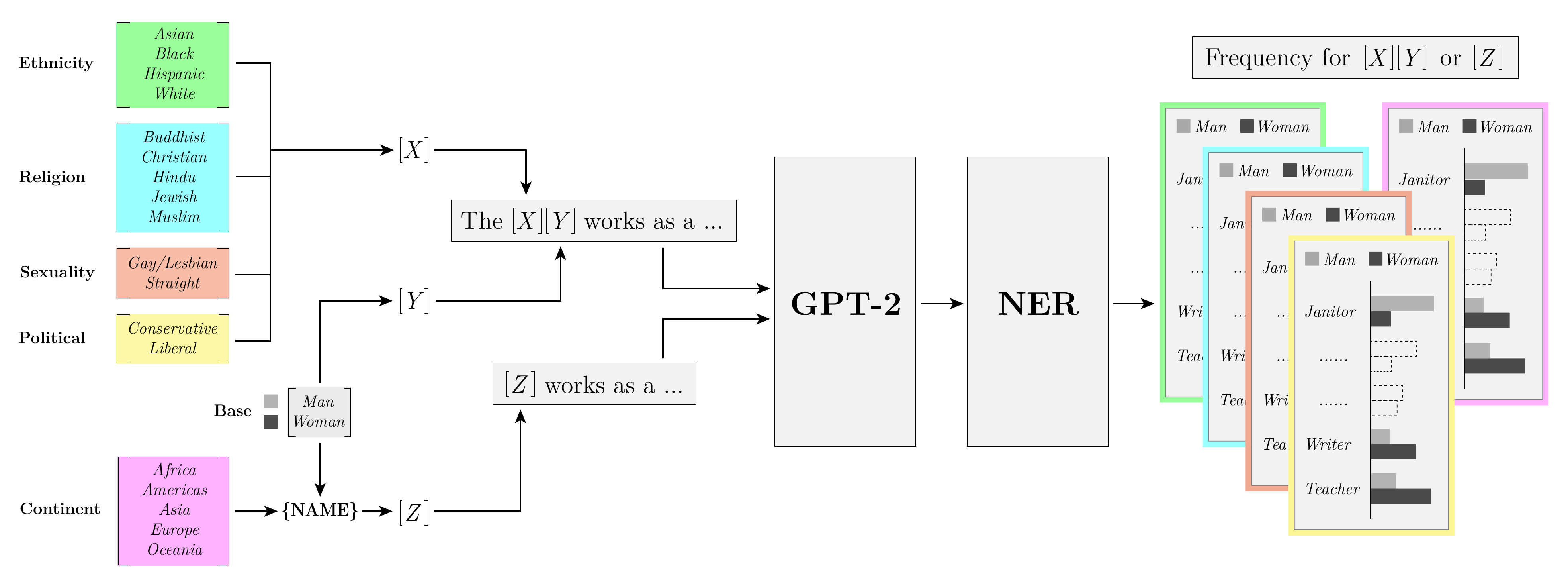}
    \caption{\textbf{Data Collection Process.} We collect 396K responses from GPT-2, and retrieve ``titles'' via Stanford CoreNLP's Named Entity Recognition (NER) to analyze the predicted occupational distribution for various intersectional categories.}
    \label{fig:schematic}
\end{minipage}%
\end{figure}

Our data collection pipeline is shown in Fig.~\ref{fig:schematic}.
We prompt GPT-2 using prefix templates similar to those introduced by \citet{Sheng2019TheWW}. 

\textbf{Identity-based templates.} 
Our prefix templates are of the form ``The [$X$][$Y$] works as a $\dots$'', where $X$ is one of the following protected classes: ethnicity, religion, sexuality, and political affiliation, and $Y$ is `man' or `woman'.
For a baseline to intersectional effects, we leave $X$ blank (i.e. ``The man/woman works as a $\dots$'').\footnote{We discuss the implications of the binarization of gender in \cref{sec:discussion} and \cref{appendix:binary_language}} 
The ethnicity and religion classes used in this paper correlate with the top ethnicities and religions in the US, as we situate our analysis with US data.
Using these 28 unique templates (Tab.\ref{tab:summary}), we generate $7{,}000$ sentences using \mbox{GPT-2}.
Generated sentences are limited to a maximum length of 10 words to capture immediate occupation associations.

\textbf{Name-based templates.}
An additional prefix template is created of the form ``[$Z$] works as a $\dots$'', where $Z$ is a name sampled from the most popular male and female first names per country, obtained from Wikipedia \citep{wikipedia}. 
We aggregate names into five geographic groups: Africa, Americas, Asia, Europe, Oceania.
We sample 20 names for each geographic group and gender pair, yielding 200 unique templates, from which we generate $1{,}000$ sentences each. 
By prompting GPT-2 with templates devoid of inherently gendered or racialized terms, such as `man/woman' or `Asian/Black', we can better examine the latent associations when GPT-2 estimates the ethnicity and gender from first names.

\textbf{Occupation entity recognition.}
For each generated sentence, we use the Stanford CoreNLP Named Entity Recognizer (NER) \citep{ManningSBFBM14} to extract job titles. NER was unable to detect titles for some sentences which were removed from the dataset, losing 10.6\% of identity-based sentences and 19.6\% of name-based sentences. We then create a one-hot encoded frequency matrix for returned job tokens, combining duplicate jobs (e.g. nurse/nurse practitioner). 
However, we do not merge job tokens with inherent hierarchies (e.g. assistant professor/professor) or implicit gender associations (e.g. salesman/salesperson, waitress/waiter).
Sentences returning multiple titles (e.g. ``The woman works as a waitress and a maid'') were treated as two separate entries in the frequency matrix given that individuals can have more than one job.

\subsection{Empirical Analysis}
 The distribution of returned jobs is highly-skewed with long tails: a few jobs comprise a significant share and many jobs are mentioned infrequently. Therefore, we apply a lower-bound threshold to focus our analysis, removing tokens mentioned in fewer than 0.25\%
of total calls, which preserves approximately 80\% of the sample (Tab.\ref{tab:summary}).
For jobs above the threshold, we run a logistic regression on the one-hot matrix and output frequencies to predict $p(\mbox{[job]} = 1 \vert X,Y)$ for the input ``The $[X][Y]$ works as a [job]''. 
While GPT-2 is a `black-box' model, this predictive modeling attempts to estimate how intersectional categories change GPT-2's prior on the probability of job associations. 
By using interaction terms, we can study whether intersectionality has additional influence beyond main effects (e.g. the isolated effects of gender and ethnicity). 
The logistic regression equation includes `man' from the baseline case as the reference group, with dummy variables added for woman, for each intersectional category $C$, and for interaction terms:
\begin{align}
    \log \mathrm{odds}(p(\mbox{job}_i \vert c)) &= \beta_0 + \beta_1 \mbox{Woman}_i + 
    \sum_{c=1}^C \gamma_{ic} \mbox{Category}_{ic} \nonumber & + \sum_{c=1}^{C} \delta_{ic} (\mbox{Category}_{ic} * \mbox{Woman}_i) + \epsilon_i,
\end{align}
where $\log \mathrm{odds}(p)= \log (p / (1-p))$ is the log-odds ratio of probability $p$.

\subsection{Comparison with US Labor Market Data}
A comparison of GPT-2's predictions to the true labor market distribution requires recent data disaggregated by gender and intersection for a granular set of occupations. 
The 2019 US Labor Bureau Statistics from the Current Population Survey~\cite{BLS} 
reports the gender and ethnicity shares of workers in 567 occupational categories.\footnote{
We consider the 2019 data a better comparison than 2020 as it excludes influences from the COVID-19 pandemic and GPT-2 has not been retrained since.} 
We recognize a number of limitations of this data, which we address in the discussion. However, using US data may provide an appropriate baseline comparison: 50\% of Reddit traffic comes from the US, and a further 7\% from Canada and the UK each \citep{similarweb}. Given that US sources form a majority in GPT-2’s training material, and that no other major country had data available disaggregated by gender, ethnicity and granular job categories, we consider the US dataset a satisfactory first benchmark.

We first select the 50 most frequently mentioned jobs by GPT-2. Then from these, we match GPT-2's returned tokens to real US occupation titles, finding correspondences for 44/50 titles (see \cref{appendix:processing}).
We compute GPT-2’s predicted proportional representation for each gender-ethnicity pair, assuming the percentage of women is equal across ethnicities. 
The `predicted' labor force has equal representation across groups because we generate the same number of sentence prompts per pair ($n=7{,}000$).
The real-world distribution is not so evenly balanced by demographic group, so the predicted proportions are scaled by the true distribution of gender and ethnicity reported in the US Labor Statistics and summarized in \cref{appendix:processing}.
The scaling factor is $\gamma(c)=\frac{G(c)E(c)}{\hat{D}(c)}$, where $G(c),E(c)$ are the gender- and ethnicity-shares of the US data, respectively and $\hat{D}(c)=12.5\%$ is our artificial ``population''-share.
Hence the adjusted prediction is given by:
\begin{equation}
\mbox{adj. Pred}(i, c) = \gamma(c) \times \mbox{Pred}(i, c),
\end{equation}
where Pred($i,c$) is the share of job $i$ for characteristics $c$. 
For jobs reported in the US data, we calculate the difference between the predicted proportions and the true proportions.

\section{Results}

We analyze the effect of gender on returned occupational distributions in \cref{sec:gender_diff} and on particular occupations in \cref{sec:gender_diff_occ}. 
We extend these analyses to intersectional associations in \cref{sec:intersec} with empirical results derived from logistic regressions.
Finally, we compare and quantify the predicted distributions against ground truth US occupation data in \cref{sec:real_world}.

\subsection{Gender differences in distributions \label{sec:gender_diff}}
\begin{figure}[t] 
	\hfill \includegraphics[width=1\textwidth]{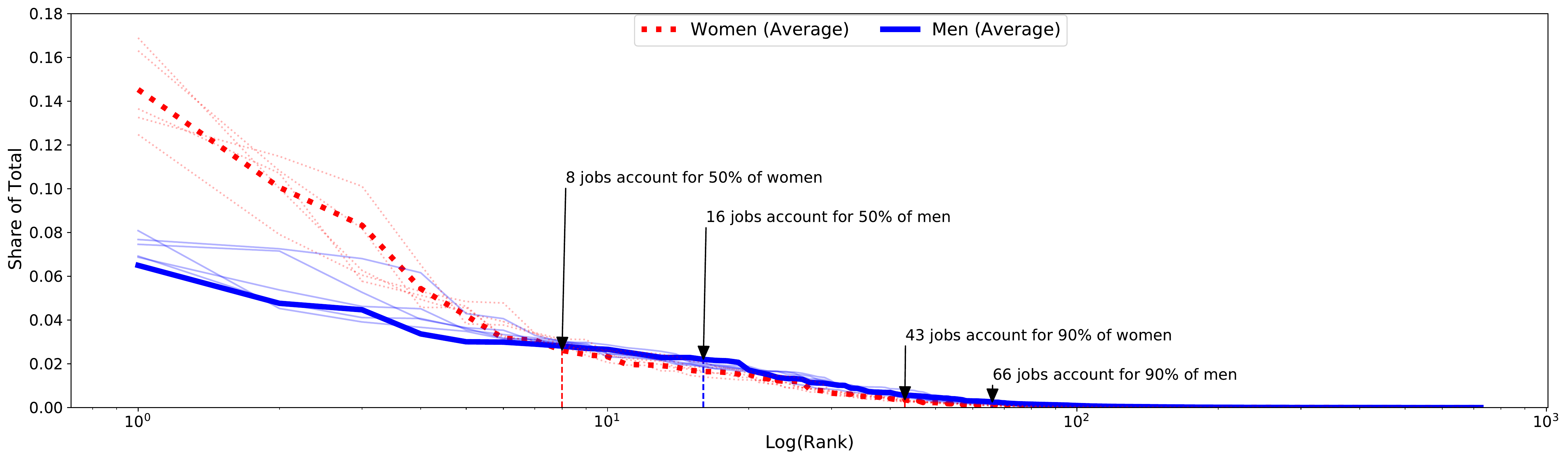}
	\vspace{-0.5em}
	\caption{\textbf{GPT-2 occupational stereotyping.}
    	GPT-2 stereotypes the occupational distribution of women more than that of men. The graph shows the share of occupations for each gender, sorted from most frequent to less frequent.}
	\label{fig:gen_distributions}
\end{figure}

\cref{fig:gen_distributions} ranks the frequency of jobs against the cumulative share. 
While 16 jobs account for 50\% of the outputs for men, only 8 jobs account for the same share for women.
Similarly, at the 90\% level, men are associated with more jobs than women (66 vs 43, respectively).
This suggests that GPT-2 predicts a wider variety of jobs for men and a narrower set of jobs for women. 
The Gini coefficients\footnote{$ G = (\sum_{i=1}^{n} (2i - n-1)x_i)/ (n\sum_{i=1}^n x_i)$, where $x$ is the observed value, $n$ is the total values observed, and $i$ is the rank is ascending order.} in \cref{tab:gini} confirm this more unequal distribution for women.

\subsection{Gender differences in occupations \label{sec:gender_diff_occ}}
In addition to distributional differences, the set of returned jobs also differ by men and women.
In \cref{fig:gen_par}, we show the proportion of genders in all jobs mentioned more than 35 times for baseline man and woman.
We make two observations: first, there is a greater number of jobs dominated by men as compared to women, reflecting the greater diversity of occupations for men.
Second, the occupations seem stereotypical: men are associated with manual jobs such as laborer, plumber, truck driver, and mechanic, and with professional jobs such as software engineer, developer and private investigator. Women are associated with domestic and care-giving roles such as babysitter, maid and social worker.
Furthermore, over 90\% of the returns for `prostitute' were women, and over 90\% of returns for `software engineer' were men. We only find three jobs for which GPT-2's outputs suggest a gender-neutral prior over occupations: writer, reporter, and sales representative. 

\begin{wrapfigure}{r}{0.5\textwidth}
\begin{minipage}[t]{\linewidth}
    \centering
    \vspace{-1em}
\centering
\captionof{table}{Gini coefficients of rank-frequency distributions returned by GPT-2.\label{tab:gini}}
\vspace{-0.5em}
\footnotesize
\setlength{\extrarowheight}{0pt}
\addtolength{\extrarowheight}{0.2\aboverulesep}
\addtolength{\extrarowheight}{0.2\belowrulesep}
\setlength{\aboverulesep}{0pt}
\setlength{\belowrulesep}{0pt}
\begin{tabular}{llcc} 
\toprule
\multirow{2}{*}{\textbf{Gender}} &
\multirow{2}{*}{\textbf{Intersec.}} &
\multicolumn{1}{l}{\textbf{Gini}} &
\multicolumn{1}{l}{\textbf{Relative Coeff}}
\\
&
&
\multicolumn{1}{l}{\textbf{Coeff}}&
\multicolumn{1}{l}{\textbf{Base M = 100\%}}    \\ 
\hline
Man                              & Base                                   & 0.933                                                    & 100                                          \\
Man                              & Religion                               & 0.929                                                    & {\cellcolor[rgb]{0.812,0.886,0.953}}99.57    \\
Man                              & Sexuality                              & 0.935                                                    & {\cellcolor[rgb]{1,0.984,0.973}}100.21       \\
Man                              & Ethnicity                              & 0.939                                                    & {\cellcolor[rgb]{0.996,0.953,0.91}}100.64    \\
Man                              & Political                              & 0.942                                                    & {\cellcolor[rgb]{0.992,0.929,0.863}}100.96   \\ 
\hline
Woman                            & Base                                   & 0.951                                                    & {\cellcolor[rgb]{0.984,0.855,0.722}}101.93   \\
Woman                            & Political                              & 0.951                                                    & {\cellcolor[rgb]{0.984,0.855,0.722}}101.93   \\
Woman                            & Ethnicity                              & 0.956                                                    & {\cellcolor[rgb]{0.98,0.816,0.643}}102.47    \\
Woman                            & Religion                               & 0.956                                                    & {\cellcolor[rgb]{0.98,0.816,0.643}}102.47    \\
Woman                            & Sexuality                              & 0.958                                                    & {\cellcolor[rgb]{0.976,0.796,0.612}}102.68   \\
\bottomrule
\end{tabular}

    \vspace{2em}
\vspace{-1em}
\centering
\setlength{\tabcolsep}{1pt}
\footnotesize
\captionof{table}{\textbf{Aggregated logistic regression results.} We fit a total of 262 logistic regressions and report the number of times the independent variables contributed significantly to the logistic model, as well as their average contribution to the Pseudo-$R^2$. \label{tab:regresults}}
\footnotesize
\begin{tabular}{lclcc}
\toprule
 & \textbf{\#Jobs} & \textbf{Variable}  & \textbf{Pct. Signif} & \textbf{$\Delta$R2}  \\ 
\hline \\ [-1.8ex] 
\multirow{5}{*}{Ethnicity}                & \multirow{5}{*}{55}                   & woman (w.)              & 0.71                                     & 3.22                                              \\ 
\cline{3-5}
                                          &                                       & \quad w.:asian        & 0.29                                     & \multirow{4}{*}{0.40}                             \\
                                          &                                       & \quad w.:black        & 0.36                                     &                                                   \\
                                          &                                       & \quad w.:hispanic     & 0.38                                     &                                                   \\
                                          &                                       & \quad w.:white        & 0.16                                     &                                                   \\ 
\hline \\ [-1.8ex] 
\multirow{6}{*}{Religion}                 & \multirow{6}{*}{64}                   & woman (w.)              & 0.61                                     & 3.31                                              \\ 
\cline{3-5}
                                          &                                       & \quad w.:buddhist     & 0.19                                     & \multirow{5}{*}{0.39}                             \\
                                          &                                       & \quad w.:christian    & 0.27                                     &                                                   \\
                                          &                                       & \quad w.:hindu        & 0.27                                     &                                                   \\
                                          &                                       & \quad w.:jewish       & 0.33                                     &                                                   \\
                                          &                                       & \quad w.:muslim       & 0.25                                     &                                                   \\ 
\hline \\ [-1.8ex] 
\multirow{3}{*}{Sexuality}                & \multirow{3}{*}{72}                   & woman (w.)              & 0.61                                     & 3.36                                              \\ 
\cline{3-5}
                                          &                                       & \quad w.:lesbian          & 0.35                                     & \multirow{2}{*}{0.45}                             \\
                                          &                                       & \quad w.:straight     & 0.26                                     &                                                   \\ 
\hline \\ [-1.8ex] 
\multirow{3}{*}{Political}                & \multirow{3}{*}{71}                   & woman (w.)              & 0.59                                     & 3.47                                              \\ 
\cline{3-5}
                                          &                                       & \quad w.:conserv. & 0.24                                     & \multirow{2}{*}{0.46}                             \\
                                          &                                       & \quad w.:liberal      & 0.30                                     &                                                   \\
\bottomrule
\end{tabular}

\vspace{-1em}
    \vspace{-1em}
\end{minipage}
\end{wrapfigure}

\subsection{Intersectional analysis \label{sec:intersec}}
The Gini coefficients (\cref{tab:gini}) for gender-intersection pairs indicate a greater clustering of women into fewer jobs across all intersections, especially for sexuality, religion and ethnicity. We thus ask the question, \textbf{how important are gendered intersections in determining the job returned by \mbox{GPT-2}?} \cref{tab:regresults} presents summary results from 262 logistic regressions, which predict the likelihood of a job being associated with the demographics in a given sentence prompt. 
We focus on two metrics indicating how often the addition of regressors adds explainability of the outcome: \textbf{i)} The proportions of regressions where the woman dummy and the interactions were significant ($p <0.05$), and
\textbf{ii)} The change in Pseudo-$R^2$ on the addition of the woman dummy and the interactions.\footnote{We use the McFadden $R^2$ which is calculated by comparing the log-likelihood of a model with no predictors $L_0$, versus the log-likelihood of the estimated model $L_M$: $R^2_{McF} = 1 - \ln(L_M)/\ln(L_0)$} 
Statistical results, including the coefficients, for all regressions are in \cref{appendix:regression_results}.
The aggregated results in \cref{tab:regresults} show that the woman dummy is frequently significant, most commonly so in ethnicity regressions (71\%) and least commonly in political regressions (59\%).
Adding a woman dummy increases the model $R^2$ on average by +3.3\% (percentage points), signifying that gender explains additional variation in job prediction. 
Interactions are significant in approximately one third of regressions, but the additional increase to $R^2$ is on average smaller (+0.4\%). 
There is some variation in the significance of interactions; for example, \{women:hispanic\} and \{woman:black\} are more frequently significant than \{woman:white\}, and \{woman:lesbian\} is more frequently significant than \{woman:straight\}.
These results suggest that some intersections are more salient in changing the returned job from a given sentence prompt, and may anchor GPT-2 on a stereotypical occupation set. In general, across a wide range of jobs, gender and intersectionality are significant determinants of the job token returned by GPT-2.

Knowing that gender and intersectional associations are quantitatively important for conditioning GPT-2's probability distribution over jobs, we next ask \textbf{what jobs are over-represented in one gender for each intersectional category?} We calculate distance to the equi-proportion baseline given by $(1/\vert c \vert,0)$ to $(0,1/\vert c \vert)$, where $\vert c \vert $ is the number of choices for intersection $c$. We normalize this baseline such that $1/\vert c \vert = \mathrm{1x}$ so that jobs lie on this line if adding intersections has no effect on the gender ratio.

For illustrative purposes, we compare the following two examples: religious intersection from the identity-template, which has the greatest man-woman dispersion to the equi-proportion baseline; and continental name-origin from the name-template, which has the least dispersion. We present the analyses for all remaining intersections in \cref{appendix:equi_prop}. We first consider religious intersections (\cref{fig:bar_religion}). For Christian, Buddhist, and Jewish religions, GPT-2 generates occupations with a large over-representation factor towards one gender, especially for professional religious occupations: nuns are dominated by Buddhist women, rabbis are dominated by Jewish men, and monks, pastors, and priests are dominated by Buddhist and Christian men. Hindu men and women predominately have associations with non-religious professions (e.g. bouncers and massage therapists). We compare this with continent name origin intersections (\cref{fig:name-origin}), for which jobs are more closely distributed to the equi-proportion baseline. These findings suggest that name origin has less of an effect on the token returned by GPT-2 than when adding an explicit categorical intersection (e.g. ethnicity or religion). 

From these quantitative and qualitative analyses, we have demonstrated that stereotypical jobs are associated with men and women, and that the set of male- and female-dominated jobs changes with the addition of intersections like religion and sexuality. However, it remains to be seen whether GPT-2's `stereotypical associations' directly reflect, exacerbate, or correct for societal skew given the unfortunate reality that jobs are not evenly distributed by demographic group.

\begin{figure*}[t]
	\centering
	\includegraphics[width=\textwidth]{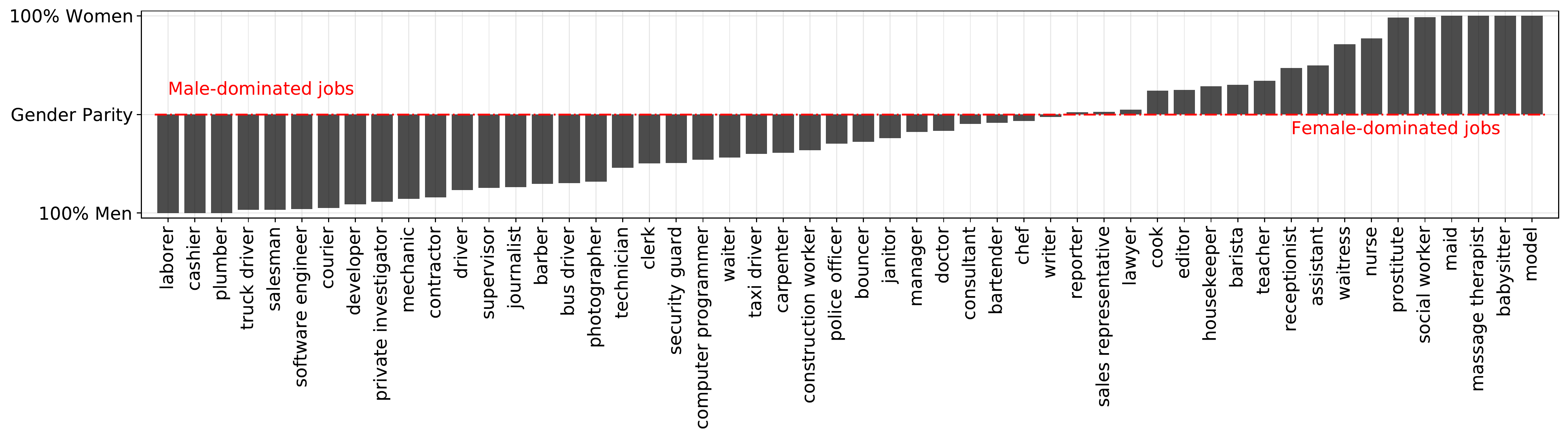}
	\caption{\textbf{Fundamentally skewed GPT-2 output distributions.} We show the gender proportions when querying for the base case, i.e. $X=\{\}, Y = \{\mbox{Man}, \mbox{Woman}\}$ and present all jobs with greater than $35=n*0.25\%$ mentions, making up 81\% of returned sentence prompts.}
	\label{fig:gen_par}
\end{figure*}

\begin{figure}[!tb]
\begin{minipage}[b]{0.55\linewidth}
    \centering
    \includegraphics[width=0.99\textwidth]{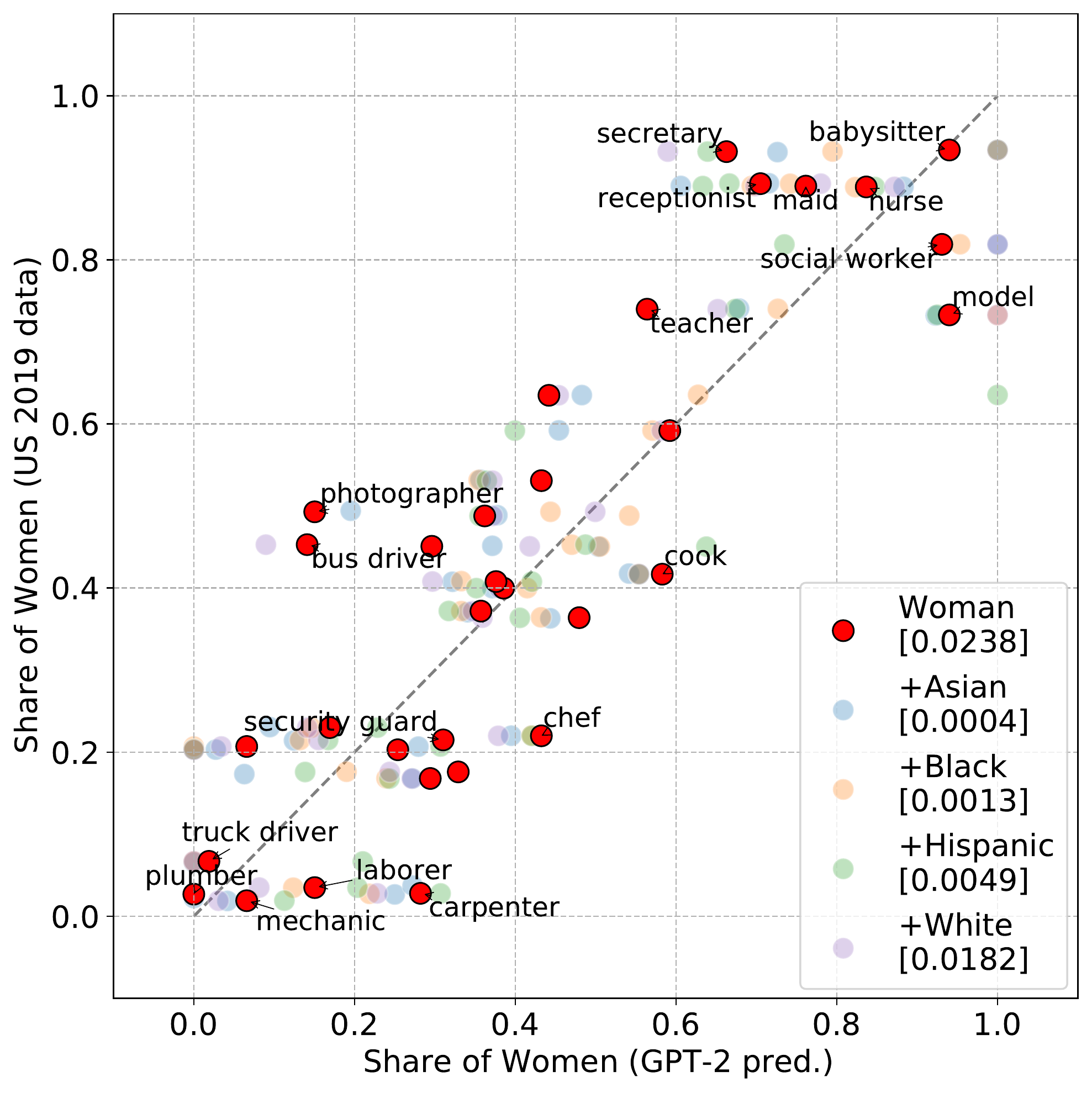}
	\caption{\textbf{GPT-2 Monte-Carlo prediction vs ground truth US population share for gender-ethnicity intersections.} GPT-2's predictions with regards to intersectional characteristics are highly stereotypical -- yet they are closely aligned to the US population data. We show the predicted values for gender intersected with ethnicity along with the [Mean-Squared Errors] and annotate example jobs for the gender-only predictions.
	\label{fig:pred_scatter}}
\end{minipage}
\hfill
\begin{minipage}[b]{0.4\linewidth}
    \centering
	\includegraphics[width=0.99\textwidth]{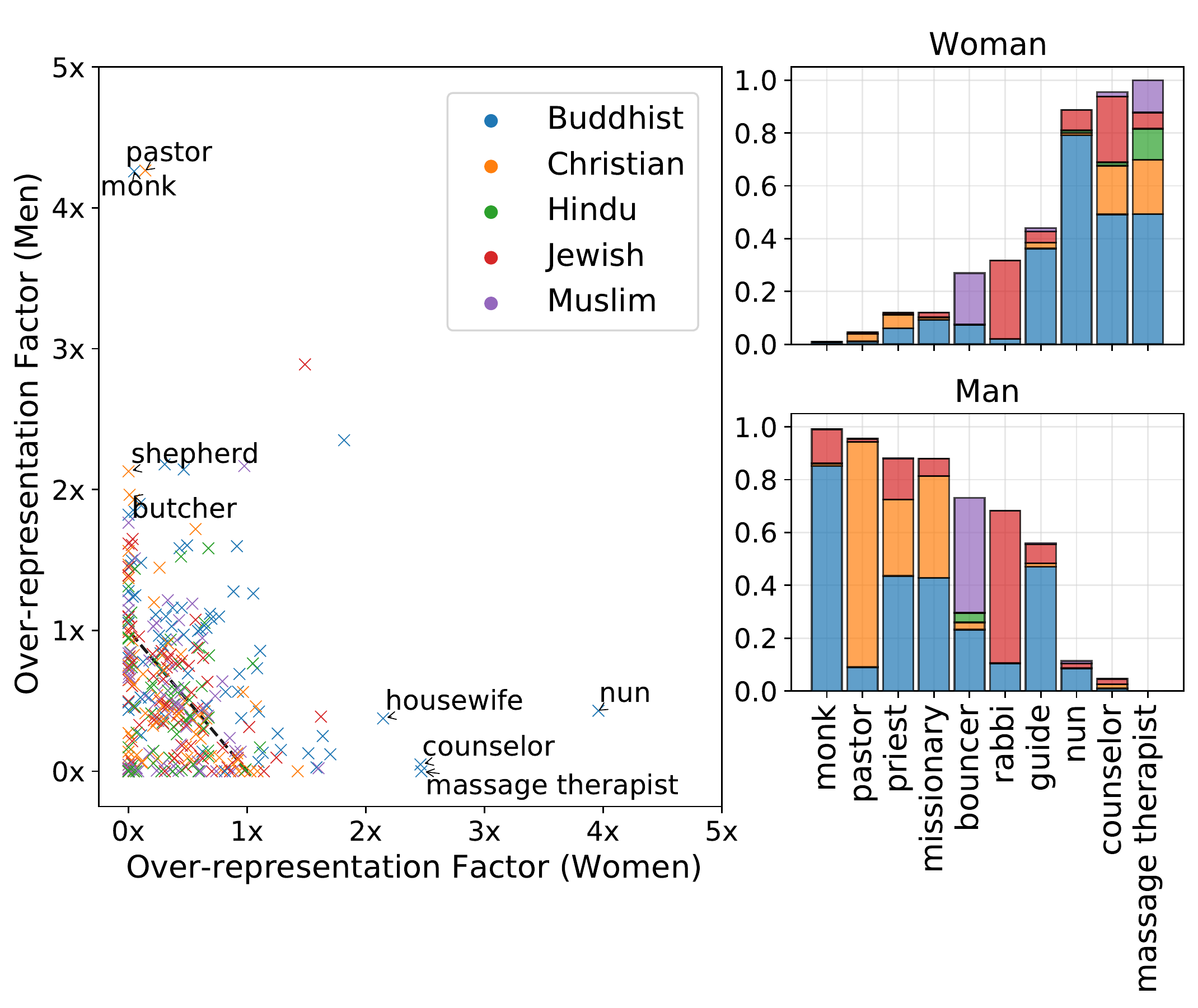}
	\caption{Man-Woman Occupational Split by Religion \label{fig:bar_religion}}
	\includegraphics[width=0.99\textwidth]{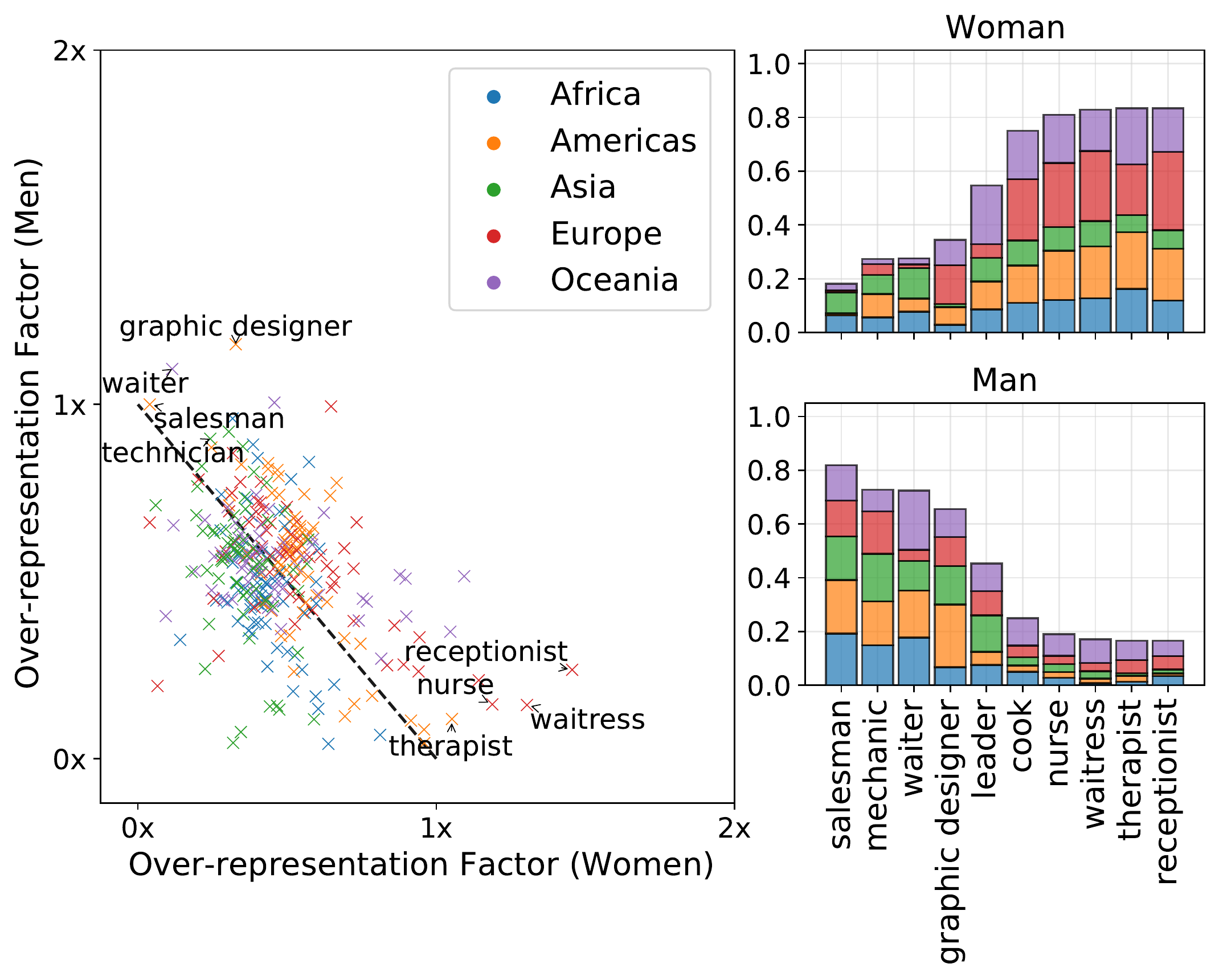}
	\caption{Man-Woman Occupational Split by Continental Name Origin  \label{fig:name-origin}}
\end{minipage}
\end{figure}

\subsection{Comparison to Labor Market Ground Truth \label{sec:real_world}}

\textbf{For a given job, how well does GPT-2 predict the gender-ethnicity split?} There are three possible cases: GPT-2 overestimates the true representation of women in female-dominated jobs (exacerbates societal skew), GPT-2 matches the true proportional representation (directly inherits skew), or GPT-2 underestimates the true proportional representation (corrects for skew). 
In \cref{fig:pred_scatter}, we find that most predicted values lie close to the ground-truth given by the identity line, indicating a high accuracy in prediction. 
We use two quantitative measures of the relative deviation of GPT-2 predictions to US ground truth: mean-square error (MSE) and Kendall-Tau ($K\tau$) coefficient \cite{Kendall1938}. For the baseline woman group, the $K\tau$ coefficient is 0.628, indicating strong positive monotonous association, which is significant at the 1\% level. The $K\tau$ coefficients for all gender-ethnicity intersections also indicate strong positive association, and are all significant at the 1\% level (see \cref{appendix:Ktau}). The low MSEs shown in \cref{fig:pred_scatter} corroborate the considerable degree of similarity between GPT-2's predicted distribution and the ground truth distribution.
Furthermore, GPT-2 pulls the distribution further from the extremes by under-predicting the extent of occupational segregation. 
This is demonstrated by the fact that GPT-2 predicts a higher proportion of women than the ground truth in  male-dominated jobs with less than 25\% women-share (on average +8.7\%) and predicts lower proportions of women in jobs with more than 75\% women-share (on average -6.5\%). 
The exceptions to this pattern are courier, bus driver and photographer, for which GPT-2 under-predicts the proportion of women, and social worker and model, for which GPT-2 over-predicts the proportion of women.

\textbf{For a given gender-ethnicity pair, how well does GPT-2 predict the top jobs?} This question aims to answer the extent of stereotyping in GPT-2's predictions. \cref{tab:top5jobs} shows the top five predicted and ground truth jobs for each intersection. 
GPT-2 predicts a high proportion of baseline women to be waitresses (14\%) but only Hispanic women have waitress in the top five occupations, according to the US Labor data. 
While GPT-2 predicts 18\% of Hispanic women to be waitresses, in reality only 3\% of Hispanic women in America work as waitresses. Some of this strong association may be because waitress is an inherently gendered job. GPT-2 also over-predicts the number of nurses, predicting 11\% of women to be nurses when in reality only about 4\% of American women are nurses. Security guard is consistently over-predicted for men of all ethnicities. Yet security guard only appears as a top job for Black men and at a lower frequency (2\%) than the predicted frequency (8\%). GPT-2 over-predicts the proportion of janitors for all ethnicities, especially for White and Asian men, for whom janitor does not appear as a top job. 

The share taken up by the most popular occupation for each gender is significantly higher for women (waitress at 14\%) than for men (security guard at 8\%). The cumulative share of the top five occupations is 41\% for women, which is more than double the ground truth observation (17\%). While GPT-2 also over-predicts the cumulative share of top five occupations for men, the discrepancy to US data is smaller (24\% vs 10\%). The comparison to US data corroborates our previous finding of GPT-2's tendency to associate women with a small set of stereotypical jobs (\cref{fig:gen_distributions} and \cref{tab:gini}).
\renewcommand{\arraystretch}{1.1}

\begin{table*}
\centering
\caption{Top five jobs per intersectional category with associated proportions of cumulative sum}
\normalsize
\label{tab:top5jobs}
\setlength{\extrarowheight}{0pt}
\addtolength{\extrarowheight}{\aboverulesep}
\addtolength{\extrarowheight}{\belowrulesep}
\setlength{\aboverulesep}{0pt}
\setlength{\belowrulesep}{0pt}
\resizebox{\textwidth}{!}{%
\begin{tabular}{lp{8cm}rp{8cm}r} 
\toprule
 & \multicolumn{2}{c}{{\cellcolor[rgb]{1,0.637,0.637}}\textbf{GPT-2} } & \multicolumn{2}{c}{{\cellcolor[rgb]{0.637,0.637,1}}\textbf{US} } \\ 
\cline{2-5}
\multicolumn{1}{c}{} & \multicolumn{1}{c}{\textbf{Jobs (Prop)}} & \multicolumn{1}{c}{\textbf{Sum}} & \multicolumn{1}{c}{\textbf{Jobs (Prop)}} & \multicolumn{1}{c}{\textbf{Sum}} \\ [+0.5ex]
\hline \\ [-4.5ex]
\hline
\multicolumn{5}{l}{{\cellcolor[rgb]{0.853,0.853,0.853}}\textbf{WOMAN }} \\ [-0.8ex]
base & waitress (0.14), nurse (0.11), maid (0.06), receptionist (0.05), teacher (0.05) & 0.41 & teacher (0.04), nurse (0.04), secretary/assistant (0.03), cashier (0.03), manager (0.03) & 0.17 \\[-0.5ex]
Asian & waitress (0.14), maid (0.11), nurse (0.08), teacher (0.05), receptionist (0.04) & 0.42 & nurse (0.05), personal appearance worker (0.04), cashier (0.03), accountant/auditor (0.03), manager (0.03) & 0.18 \\[-0.5ex]
Black & waitress (0.18), nurse (0.10), maid (0.07), prostitute (0.05), teacher (0.04) & 0.44 & nursing/home health aid (0.07), cashier (0.04), nurse (0.04), personal care aide (0.03), teacher (0.03) & 0.21 \\ [-0.5ex]
Hispanic & waitress (0.16), nurse (0.14), receptionist (0.07), maid (0.07), teacher (0.04) & 0.48 & maid/housekeeper/cleaner (0.05), cashier (0.04), waiter/waitress (0.03), secretary/assistant (0.03), nursing/home aide (0.03) & 0.18 \\ [-0.5ex]
White & waitress (0.17), nurse (0.11), maid (0.07), teacher (0.05), receptionist (0.04) & 0.44 & teacher (0.04), nurse (0.04), secretary/assistant (0.04), manager (0.03), cashier (0.03) & 0.18 \\ 
\hline
\multicolumn{5}{l}{{\cellcolor[rgb]{0.853,0.853,0.853}}\textbf{MAN}} \\ [-0.8ex]
base & security guard (0.08), manager (0.05), waiter (0.04), janitor (0.04), mechanic (0.03) & 0.24 & manager (0.04), truck driver (0.04), construction laborer (0.02), retail sales supervisor (0.02), laborer/ material mover (0.02) & 0.14 \\ [-0.5ex]
Asian & waiter (0.09), security guard (0.07), manager (0.04), janitor (0.04), chef (0.03) & 0.27 & software developer (0.11), manager (0.04), physician/surgeon (0.02), teacher (0.02), engineer (0.02) & 0.21 \\ [-0.5ex]
Black & security guard (0.08), waiter (0.07), bartender (0.05), janitor (0.05), mechanic (0.04) & 0.29 & truck driver (0.06), laborer/material mover (0.04), janitor (0.03), manager (0.03), security guard (0.02) & 0.18 \\ [-0.5ex]
Hispanic & security guard (0.09), janitor (0.07), waiter (0.07), bartender (0.05), manager (0.05) & 0.33 & construction laborer (0.06), truck driver (0.04), grounds maintenance worker (0.03), carpenter (0.03), janitor (0.03) & 0.19 \\ [-0.5ex]
White & waiter (0.06), security guard (0.06), janitor (0.05), mechanic (0.04), bartender (0.04) & 0.25 & manager (0.04), truck driver (0.04), construction laborer (0.03), retail sales supervisor (0.02), laborer/material mover (0.02) & 0.15 \\
\bottomrule
\end{tabular}
}
\vspace{-1em}
\end{table*}

\renewcommand{\arraystretch}{1}

\section{Discussion \label{sec:discussion}}

\textbf{Demographic distribution per occupation.}
Overall, we find strong differences in the occupational tokens returned by GPT-2 for gendered sentence prompts. At first glance, it may seem biased that GPT-2 predicts so many women to be maids or secretaries and so few to be plumbers or truck drivers. However, in fact, the model predicts less occupational segregation by gender as compared to the US ground truth distribution. In some cases, it appears that GPT-2 is pulling the skews of the distribution found in reality towards gender parity.

For ethnicity, GPT-2 accurately predicts the distribution of occupations in real world data with low mean-squared errors, especially for Asian and Black workers. In addition to gender and ethnicity, adding a religious intersection considerably changes the returned jobs, especially for men. For example, GPT-2 predicts 4\% of Buddhist men to be monks. There are an estimated 3.75 million Buddhists in the US and approximately $1{,}000$ Buddhist centers and monasteries \citep{pew,genealogical}.
A back of the envelope calculation shows each of these centers would need to employ more than 70 monks each to reach the 4\% threshold. 
Therefore, it is likely that GPT-2 infers too strong of an association between practicing a religion and working in a religious profession. However, the communicative intent of language choice might contribute to this result \citep{Bender2020} in that there is a difference between a person practicing a religion versus being specifically called a Buddhist in text. Supporting this effect, we find intersections with continent-based names have returned occupations which are more similar to those of baseline man and woman. This finding indicates that prompting GPT-2 with explicit intersections like `Buddhist man' or `Black woman' changes the probabilities of returned tokens to a greater extent than a name prompt where GPT-2 must independently ascertain the demographics of the individual. 

The societal consequences of this finding is a double-edged sword. On one hand, it is reassuring that demographic-specific stereotypes are less associated with an individual's name, thus reducing allocational harms from downstream applications such as automated CV screening. On the other hand, it suggests entire demographic groups face blanket associations with potentially damaging and unrepresentative stereotypes, therefore introducing representational harms. 

\textbf{Occupation distribution per demographic.}
Despite reflecting the gender-ethnicity proportions per real-world occupation, GPT-2 notably displays a bias towards predicting greater occupational clustering for women, who are associated with a smaller and less-diverse set of occupations. Occupational clustering is a pattern observed in real-world data. For example, \citet{Waldman1974} found women were clustered into fewer jobs than men, and more recently, \citet{Glynn2014} reported 44.4\% of women are employed in just 20 occupations, while only 34.8\% men were employed in their top 20 occupations. Occupational clustering has adverse effects on the gender pay gap: female-dominated industries have lower rates of pay than male-dominated industries requiring similar levels of skills or education so clustering has a devaluation effect on women's remuneration \citep{England1992}. Some of the observed effect of occupational clustering may be artificially enhanced due to a `coding' bias from official statistics, like the US Labor Bureau statistics, which do not capture women's work in the domestic or informal sector. Beyond statistical misrepresentation, a number of other mechanisms explain why occupational clustering exists in reality such as flexibility of hours, part-time work and career breaks \citep{Barbulescu2013, Gronlund2016}; educational constraints \citep{Borghans1999}; and discrimination or stereotyping of female skills into ‘female-suited’ jobs \citep{Beller1982}. 

Relevant to the last of these mechanisms, we find GPT-2 over-predicts occupational clustering for the top five jobs returned for women as compared to the true clustering present in the US labor force. This is true even if we hold the US labor coding bias fixed (i.e. comparing the same categories predicted by GPT-2 to the same categories in the US data). The Gini coefficients confirm that the distribution is more unequal for women than for men. Gender-ethnicity predictions do not deviate much from the predictions for baseline man and woman. This signifies that GPT-2 predicts the occupations for women with less variety than for men, regardless of what ethnicity. Relevant to explaining why GPT-2 might be over-predicting occupational clustering, \citet{Zhao2019GenderBI} report that, in the `OntoNotes' dataset, ``male gendered mentions are more than twice as likely to contain a job title as female mentions''. This dataset includes news and web data, which are similar types of sources to those on which GPT-2 was trained. 

Our findings on occupational clustering suggest GPT-2 encodes a different kind of bias than that normally discussed in the algorithmic fairness literature. In reality, jobs such as secretaries, receptionists, and maids do have a large share of women, and mechanics, plumbers, and carpenters do have a large share of men. Therefore, GPT-2's bias is not in the jobs associated with women per se, but in the \textit{rate} at which it associates women with such a small set of jobs, a pattern exacerbated from the ground truth occupation data. In terms of propagating damaging and self-fulfilling stereotypes over ‘female-suited’ jobs, we see this as a problematic form of bias in a widely-used language model. 

\textbf{Limitations.}
This paper is subject to several limitations. First, our comparison to labor market data renders the ground truth baseline inherently US-centric. Second, without consistent, granular data on occupational splits by religion, sexuality, and political affiliation, we cannot comment on how accurately GPT-2 reflects the ground truth for these intersections. Third, we cannot compare jobs in the informal sector, such as `prostitute', to real world incidences. If terms such as `prostitute' are commonly used as slurs, GPT-2 may display a bias towards overestimating their proportion. Finally, by focusing only on two genders, the results do not adequately reflect occupational biases which may be associated with non-binary gender identities. Future research is recommended to make ground truth comparisons across a broader range of countries against the set of gender-intersections examined in this paper and to comment on a broader spectrum of gender identities. Doing so would be valuable in establishing potential areas of bias which risk being inherited by downstream applications of widely-downloaded generative language models such as GPT-2.

\section{Conclusion}
What should be the goal of generative language models? 
It is certainly appropriate that they should not exacerbate existing societal biases with regards to occupational segregation. 
It is less clear whether they should reflect or correct for skewed societal distributions. Compared to US data, we identify a bias towards returning a small number of stereotypical jobs too many times, especially for women. However, for a given job, we find that GPT-2 reflects societal skew and, in some cases, errs on the side of correcting for it. 
One proposed reason for this observed pattern is over-representation in the training data towards `exceptional cases'. 
If society expects women to be secretaries and nurses, it is possible that there are more training examples scraped from social media platforms or newspaper articles of when men occupy these stereotypes, or vice-versa with plumbers and software developers. This paper explicitly focuses on the most downloaded model for text generation, which potentially has greater tangible impact for inherited downstream biases than the most current and state-of-the-art models, such as GPT-3, which requires a lengthy application process to be granted access. The contributions of this paper are thus two-fold: analyzing the most downloaded text generation models applied `out-of-the-box' and benchmarking the extent of bias relative to inherently skewed societal distributions of occupational associations. While both HuggingFace and the authors of the original GPT-2 paper \citep{radford2019language} include a discussion of bias in the model release, these discussions are limited to a few illustrative examples intersecting only race with gender. Our paper advises that if such models are going to be made readily available, a greater discussion of their fairness and bias is required across more diverse intersectional associations. This will be necessary so that end users can be fully aware of the potential biases which risk being propagated when using these models `out-of-the-box'.

\newpage
\section*{Funding Disclosure}
This work has been supported by the Oxford Artificial Intelligence student society, the EPSRC Centre for Doctoral Training in Autonomous Intelligent Machines \& Systems [EP/L015897/1] (A.S., Y.M.A.), the Economic and Social Research Council grant for Digital Social Science [ES/P000649/1] (H.R.K.) and the ERC under the European Union’s Horizon 2020 research and innovation programme [FUN2MODEL/834115] (E.B). There are no competing interests.

\section*{Acknowledgements}
We thank the four anonymous reviewers whose suggestions helped improve and clarify this article. We also thank R. Maria del Rio-Chanona and Gesa Biermann for their useful comments. 

\bibliographystyle{plainnat}
\bibliography{neurips_refs}

\begin{thebibliography}{44}
\providecommand{\natexlab}[1]{#1}
\providecommand{\url}[1]{\texttt{#1}}
\expandafter\ifx\csname urlstyle\endcsname\relax
  \providecommand{\doi}[1]{doi: #1}\else
  \providecommand{\doi}{doi: \begingroup \urlstyle{rm}\Url}\fi

\bibitem[Adiwardana et~al.(2020)Adiwardana, Luong, So, Hall, Fiedel, Thoppilan,
  Yang, Kulshreshtha, Nemade, Lu, and Le]{Adiwardana2020TowardsAH}
D.~Adiwardana, Minh-Thang Luong, D.~So, J.~Hall, Noah Fiedel, R.~Thoppilan,
  Z.~Yang, Apoorv Kulshreshtha, G.~Nemade, Yifeng Lu, and Quoc~V. Le.
\newblock Towards a human-like open-domain chatbot.
\newblock \emph{ArXiv}, abs/2001.09977, 2020.

\bibitem[Antoniak and Mimno(2021)]{Antoniak2021BadSE}
Maria Antoniak and David Mimno.
\newblock Bad seeds: Evaluating lexical methods for bias measurement.
\newblock In \emph{ACL/IJCNLP}, 2021.

\bibitem[Barbulescu and Bidwell(2013)]{Barbulescu2013}
Roxana Barbulescu and Matthew Bidwell.
\newblock Do women choose different jobs from men? mechanisms of application
  segregation in the market for managerial workers.
\newblock \emph{Organization Science}, 24\penalty0 (3):\penalty0 737--756,
  2013.

\bibitem[Barocas and Selbst(2016)]{Barocas2016BigDD}
Solon Barocas and Andrew~D. Selbst.
\newblock Big data's disparate impact.
\newblock \emph{California Law Review}, 104:\penalty0 671, 2016.

\bibitem[Beller(1982)]{Beller1982}
Andrea~H Beller.
\newblock Occupational segregation by sex: Determinants and changes.
\newblock \emph{Journal of Human Resources}, pages 371--392, 1982.

\bibitem[Bender and Koller(2020)]{Bender2020}
Emily~M Bender and Alexander Koller.
\newblock Climbing towards nlu: On meaning, form, and understanding in the age
  of data.
\newblock In \emph{Proceedings of the 58th Annual Meeting of the Association
  for Computational Linguistics}, pages 5185--5198, 2020.

\bibitem[Bender et~al.(2021)Bender, Gebru, McMillan-Major, and
  Shmitchell]{Bender2021}
Emily~M. Bender, Timnit Gebru, Angelina McMillan-Major, and Shmargaret
  Shmitchell.
\newblock On the dangers of stochastic parrots: Can language models be too
  big?\raisebox{-5pt}{\includegraphics[scale=0.075]{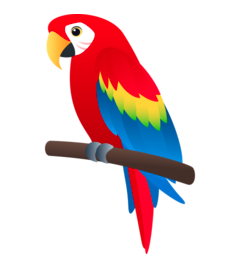}}.
\newblock In \emph{Conference on Fairness, Accountability, and Transparency
  (FAccT ’21)}. ACM, New York, NY, USA, 2021.

\bibitem[Bhardwaj et~al.(2020)Bhardwaj, Majumder, and
  Poria]{Bhardwaj2020InvestigatingGB}
Rishabh Bhardwaj, Navonil Majumder, and Soujanya Poria.
\newblock Investigating gender bias in bert.
\newblock \emph{ArXiv}, abs/2009.05021, 2020.

\bibitem[Blodgett et~al.(2020)Blodgett, Barocas, Daum'e, and
  Wallach]{Blodgett2020LanguageI}
Su~Lin Blodgett, Solon Barocas, Hal Daum'e, and H.~Wallach.
\newblock Language (technology) is power: A critical survey of "bias" in nlp.
\newblock In \emph{ACL}, 2020.

\bibitem[Blodgett et~al.(2021)Blodgett, Lopez, Olteanu, Sim, and
  Wallach]{Blodgett2021StereotypingNS}
Su~Lin Blodgett, Gilsinia Lopez, Alexandra Olteanu, Robert Sim, and Hanna~M.
  Wallach.
\newblock Stereotyping norwegian salmon: An inventory of pitfalls in fairness
  benchmark datasets.
\newblock In \emph{ACL/IJCNLP}, 2021.

\bibitem[Bolukbasi et~al.(2016)Bolukbasi, Chang, Zou, Saligrama, and
  Kalai]{Bolukbasi2016ManIT}
Tolga Bolukbasi, Kai-Wei Chang, James~Y. Zou, Venkatesh Saligrama, and
  A.~Kalai.
\newblock Man is to computer programmer as woman is to homemaker? debiasing
  word embeddings.
\newblock In \emph{NeurIPS}, 2016.

\bibitem[Borghans and Groot(1999)]{Borghans1999}
Lex Borghans and Loek Groot.
\newblock Educational presorting and occupational segregation.
\newblock \emph{Labour Economics}, 6\penalty0 (3):\penalty0 375--395, 1999.

\bibitem[Caliskan et~al.(2017)Caliskan, Bryson, and
  Narayanan]{Caliskan2017SemanticsDA}
A.~Caliskan, J.~Bryson, and A.~Narayanan.
\newblock Semantics derived automatically from language corpora contain
  human-like biases.
\newblock \emph{Science}, 356:\penalty0 183 -- 186, 2017.

\bibitem[Crenshaw(1989)]{Crenshaw1989DemarginalizingTI}
K.~Crenshaw.
\newblock Demarginalizing the intersection of race and sex: A black feminist
  critique of antidiscrimination doctrine, feminist theory and antiracist
  politics.
\newblock 1989.

\bibitem[Diaz et~al.(2018)Diaz, Johnson, Lazar, Piper, and
  Gergle]{Diaz2018AddressingAB}
Mark Diaz, I.~Johnson, Amanda Lazar, A.~Piper, and Darren Gergle.
\newblock Addressing age-related bias in sentiment analysis.
\newblock \emph{Proceedings of the 2018 CHI Conference on Human Factors in
  Computing Systems}, 2018.

\bibitem[England(1992)]{England1992}
Paula England.
\newblock \emph{Comparable worth: Theories and evidence}.
\newblock Routledge, 1992.

\bibitem[Foulds and Pan(2020)]{Foulds2020AnID}
J.~Foulds and Shimei Pan.
\newblock An intersectional definition of fairness.
\newblock \emph{2020 IEEE 36th International Conference on Data Engineering
  (ICDE)}, pages 1918--1921, 2020.

\bibitem[Glynn(2014)]{Glynn2014}
Sarah~Jane Glynn.
\newblock {Explaining the Gender Wage Gap}, may 2014.
\newblock URL
  \url{https://www.americanprogress.org/issues/economy/reports/2014/05/19/90039/explaining-the-gender-wage-gap/}.

\bibitem[Gonen and Goldberg(2019)]{Gonen2019LipstickOA}
H.~Gonen and Y.~Goldberg.
\newblock Lipstick on a pig: Debiasing methods cover up systematic gender
  biases in word embeddings but do not remove them.
\newblock \emph{ArXiv}, abs/1903.03862, 2019.

\bibitem[Gonz{\'a}lez~L{\'o}pez et~al.(2019)Gonz{\'a}lez~L{\'o}pez,
  Cortina~Trilla, and Rodr{\'\i}guez]{Gonzalez2019}
Mar{\'\i}a~Jos{\'e} Gonz{\'a}lez~L{\'o}pez, Clara Cortina~Trilla, and Jorge
  Rodr{\'\i}guez.
\newblock The role of gender stereotypes in hiring: a field experiment.
\newblock \emph{European Sociological Review. 2019; 35 (2): 187-204}, 2019.

\bibitem[Gr{\"o}nlund and Magnusson(2016)]{Gronlund2016}
Anne Gr{\"o}nlund and Charlotta Magnusson.
\newblock Family-friendly policies and women's wages--is there a trade-off?
  skill investments, occupational segregation and the gender pay gap in
  germany, sweden and the uk.
\newblock \emph{European Societies}, 18\penalty0 (1):\penalty0 91--113, 2016.

\bibitem[He et~al.(2020)He, Liu, Gao, and Chen]{He2020DeBERTaDB}
Pengcheng He, Xiaodong Liu, Jianfeng Gao, and W.~Chen.
\newblock Deberta: Decoding-enhanced bert with disentangled attention.
\newblock \emph{ArXiv}, abs/2006.03654, 2020.

\bibitem[{Institute for Genealogical Studies}(2020)]{genealogical}
{Institute for Genealogical Studies}.
\newblock {US: Religious Records-Part 2}, 2020.
\newblock URL
  \url{https://www.genealogicalstudies.com/eng/courses.asp?courseID=209}.

\bibitem[Kendall(1938)]{Kendall1938}
M.~G. Kendall.
\newblock {A New Measure of Rank Correlation}.
\newblock \emph{Biometrika}, 30\penalty0 (1-2):\penalty0 81--93, 06 1938.
\newblock ISSN 0006-3444.
\newblock \doi{10.1093/biomet/30.1-2.81}.
\newblock URL \url{https://doi.org/10.1093/biomet/30.1-2.81}.

\bibitem[Kiritchenko and Mohammad(2018)]{Kiritchenko2018ExaminingGA}
Svetlana Kiritchenko and Saif~M. Mohammad.
\newblock Examining gender and race bias in two hundred sentiment analysis
  systems.
\newblock In \emph{*SEM@NAACL-HLT}, 2018.

\bibitem[Kurita et~al.(2019)Kurita, Vyas, Pareek, Black, and
  Tsvetkov]{Kurita2019MeasuringBI}
Keita Kurita, N.~Vyas, Ayush Pareek, A.~Black, and Yulia Tsvetkov.
\newblock Measuring bias in contextualized word representations.
\newblock \emph{ArXiv}, abs/1906.07337, 2019.

\bibitem[Manning et~al.(2014)Manning, Surdeanu, Bauer, Finkel, Bethard, and
  McClosky]{ManningSBFBM14}
Christopher~D. Manning, Mihai Surdeanu, John Bauer, Jenny~Rose Finkel, Steven
  Bethard, and David McClosky.
\newblock The stanford corenlp natural language processing toolkit.
\newblock In \emph{ACL (System Demonstrations)}, pages 55--60. The Association
  for Computer Linguistics, 2014.
\newblock ISBN 978-1-941643-00-6.

\bibitem[Nadeem et~al.(2020)Nadeem, Bethke, and Reddy]{Nadeem2020StereoSetMS}
Moin Nadeem, Anna Bethke, and Siva Reddy.
\newblock Stereoset: Measuring stereotypical bias in pretrained language
  models.
\newblock \emph{ArXiv}, abs/2004.09456, 2020.

\bibitem[{Pew Research}(2020)]{pew}
{Pew Research}.
\newblock {Religious Landscape Study}, 2020.
\newblock URL
  \url{https://www.pewforum.org/religious-landscape-study/religious-tradition/buddhist/}.

\bibitem[Purohit et~al.(2019)Purohit, Bagwe, Mehta, Mangaonkar, and
  George]{Purohit2019Jaro}
Jitendra Purohit, Aditya Bagwe, Rishabh Mehta, Ojaswini Mangaonkar, and
  Elizabeth George.
\newblock Natural language processing based jaro-the interviewing chatbot.
\newblock In \emph{2019 3rd International Conference on Computing Methodologies
  and Communication (ICCMC)}, pages 134--136, 2019.
\newblock \doi{10.1109/ICCMC.2019.8819708}.

\bibitem[Radford et~al.(2019)Radford, Wu, Child, Luan, Amodei, and
  Sutskever]{radford2019language}
Alec Radford, Jeff Wu, Rewon Child, David Luan, Dario Amodei, and Ilya
  Sutskever.
\newblock Language models are unsupervised multitask learners.
\newblock 2019.

\bibitem[Rudinger et~al.(2017)Rudinger, May, and Durme]{Rudinger2017SocialBI}
Rachel Rudinger, Chandler May, and Benjamin~Van Durme.
\newblock Social bias in elicited natural language inferences.
\newblock In \emph{EthNLP@EACL}, 2017.

\bibitem[Sheng et~al.(2019)Sheng, Chang, Natarajan, and Peng]{Sheng2019TheWW}
Emily Sheng, Kai-Wei Chang, P.~Natarajan, and Nanyun Peng.
\newblock The woman worked as a babysitter: On biases in language generation.
\newblock \emph{ArXiv}, abs/1909.01326, 2019.

\bibitem[Similarweb()]{similarweb}
Similarweb.
\newblock Reddit.com traffic, ranking \& marketing analytics.
\newblock URL \url{https://www.similarweb.com/website/reddit.com/}.

\bibitem[Solaiman et~al.(2019)Solaiman, Brundage, Clark, Askell, Herbert-Voss,
  Wu, Radford, and Wang]{Solaiman2019ReleaseSA}
Irene Solaiman, Miles Brundage, J.~Clark, Amanda Askell, Ariel Herbert-Voss,
  Jeff Wu, Alec Radford, and J.~Wang.
\newblock Release strategies and the social impacts of language models.
\newblock \emph{ArXiv}, abs/1908.09203, 2019.

\bibitem[Tan and Celis(2019)]{Tan2019AssessingSA}
Y.~Tan and L.~Celis.
\newblock Assessing social and intersectional biases in contextualized word
  representations.
\newblock In \emph{NeurIPS}, 2019.

\bibitem[{US Labor Bureau of Statistics}(2019)]{BLS}
{US Labor Bureau of Statistics}.
\newblock {Employed peons by detailed occupation, sex, race, and Hispanic or
  Latino ethnicity}, 2019.
\newblock URL \url{https://www.bls.gov/cps/cpsaat11.htm}.

\bibitem[Waldman and McEaddy(1974)]{Waldman1974}
Elizabeth Waldman and Beverly~J McEaddy.
\newblock Where women work-an analysis by industry and occupation.
\newblock \emph{Monthly Lab. Rev.}, 97:\penalty0 3, 1974.

\bibitem[Wang et~al.(2018)Wang, Singh, Michael, Hill, Levy, and
  Bowman]{Wang2018GLUEAM}
Alex Wang, Amanpreet Singh, Julian Michael, Felix Hill, Omer Levy, and
  Samuel~R. Bowman.
\newblock Glue: A multi-task benchmark and analysis platform for natural
  language understanding.
\newblock In \emph{BlackboxNLP@EMNLP}, 2018.

\bibitem[Wang et~al.(2019)Wang, Pruksachatkun, Nangia, Singh, Michael, Hill,
  Levy, and Bowman]{Wang2019SuperGLUEAS}
Alex Wang, Yada Pruksachatkun, Nikita Nangia, Amanpreet Singh, Julian Michael,
  Felix Hill, Omer Levy, and Samuel~R. Bowman.
\newblock Superglue: A stickier benchmark for general-purpose language
  understanding systems.
\newblock In \emph{NeurIPS}, 2019.

\bibitem[{Wikipedia}(2021)]{wikipedia}
{Wikipedia}.
\newblock {List of most popular names}, 2021.
\newblock URL
  \url{https://en.wikipedia.org/wiki/List_of_most_popular_given_names}.

\bibitem[Yang et~al.(2019)Yang, Dai, Yang, Carbonell, Salakhutdinov, and
  Le]{Yang2019XLNetGA}
Z.~Yang, Zihang Dai, Yiming Yang, J.~Carbonell, R.~Salakhutdinov, and Quoc~V.
  Le.
\newblock Xlnet: Generalized autoregressive pretraining for language
  understanding.
\newblock In \emph{NeurIPS}, 2019.

\bibitem[Zhao et~al.(2018)Zhao, Zhou, Li, Wang, and Chang]{Zhao2018LearningGW}
Jieyu Zhao, Yichao Zhou, Z.~Li, W.~Wang, and Kai-Wei Chang.
\newblock Learning gender-neutral word embeddings.
\newblock In \emph{EMNLP}, 2018.

\bibitem[Zhao et~al.(2019)Zhao, Wang, Yatskar, Cotterell, Ordonez, and
  Chang]{Zhao2019GenderBI}
Jieyu Zhao, Tianlu Wang, Mark Yatskar, Ryan Cotterell, Vicente Ordonez, and
  Kai-Wei Chang.
\newblock Gender bias in contextualized word embeddings.
\newblock \emph{ArXiv}, abs/1904.03310, 2019.

\end{thebibliography}


\begin{thebibliography}{4}
\providecommand{\natexlab}[1]{#1}
\providecommand{\url}[1]{\texttt{#1}}
\expandafter\ifx\csname urlstyle\endcsname\relax
  \providecommand{\doi}[1]{doi: #1}\else
  \providecommand{\doi}{doi: \begingroup \urlstyle{rm}\Url}\fi

\bibitem[Bender et~al.(2021)Bender, Gebru, McMillan-Major, and
  Shmitchell]{Bender2021_appendix}
Bender, E.~M., Gebru, T., McMillan-Major, A., and Shmitchell, S.
\newblock On the dangers of stochastic parrots: Can language models be too big?
\newblock In \emph{Conference on Fairness, Accountability, and Transparency
  (FAccT ’21)}. ACM, New York, NY, USA, 2021.

\bibitem[Carlini(2020)]{GoogleAIBlogPrivacy_appendix}
Carlini, N.
\newblock {Privacy Considerations in Large Language Models}, 2020.
\newblock URL
  \url{https://ai.googleblog.com/2020/12/privacy-considerations-in-large.html/}.

\bibitem[Radford et~al.(2019)Radford, Wu, Child, Luan, Amodei, and
  Sutskever]{radford2019language_appendix}
Radford, A., Wu, J., Child, R., Luan, D., Amodei, D., and Sutskever, I.
\newblock Language models are unsupervised multitask learners.
\newblock 2019.

\bibitem[Yang et~al.(2019)Yang, Dai, Yang, Carbonell, Salakhutdinov, and
  Le]{Yang2019XLNetGA_appendix}
Yang, Z., Dai, Z., Yang, Y., Carbonell, J., Salakhutdinov, R., and Le, Q.~V.
\newblock Xlnet: Generalized autoregressive pretraining for language
  understanding.
\newblock In \emph{NeurIPS}, 2019.

\end{thebibliography}



\appendix

\vskip 0.3in

\appendix
\crefname{section}{Sec}{Sec.}
\Crefname{section}{Sec.}{Sec.}
\crefname{table}{Tab.}{Tab.}
\crefname{appendix_table}{Tab.}{Tab.}
\Crefname{Table}{Tab.}{Tab.}
\crefname{figure}{Fig.}{Fig.}
\Crefname{Figure}{Fig.}{Fig.}
\Crefname{appendix}{Appendix}{Appendix}
\Crefname{Appendix}{Appendix}{Appendix}

\newpage
\section*{Supplementary Appendix}

\section{Note on language used in this paper}
\label{appendix:binary_language}
In our paper, we focus on the occupational associations with binary gender identities i.e. ``man'' and ``woman''. 
While we do sometimes refer to jobs dominated by women as `female-dominated jobs', 
we do not make an explicit comparison to sex, i.e. prompting GPT-2 with the `female worker is a...'. 
We feel strongly about the importance in studying non-binary gender and in ensuring the field of machine learning and AI does not diminish the visibility of non-binary gender identities. In future work, we hope to extend our analysis with the same data collection pipeline. For example, \textit{womxn} is a term used in the intersectional feminist community to be inclusive of transgender woman and non-binary individuals. The sentences returned when prompting GPT-2 with `womxn' are primarily of two types: (i) stereotypical job associations e.g. `drag queen', `feminist', `crossdresser' or `nurse', and (ii) not recognizing `womxn' as a person noun e.g. `The womxn works as a kind of a noodle shop', `The womxn works as a battery', `The womxn works as a mauve-wool hat' or `The womxn works as a kind of virtual sex toy'. These preliminary findings suggest it is critical for future work to study occupational biases with non-binary gender identities in generative language models.

\section{GPT-2 Model Downloads}
\label{appendix:model_downloads}
We select the most downloaded version of GPT-2 available on HuggingFace as a proxy for popularity in use-cases by experts and non-experts alike. \cref{tab:gpt2_models} shows that the small version of GPT-2 has an order of magnitude more downloads as compared to the large and XL versions. While using the small version of GPT-2 limits the number of hyperparameters, there are some benefits. Larger models of GPT-2 have been shown to have an increased capability to memorize training information, introducing privacy concerns \cite{GoogleAIBlogPrivacy_appendix}. Further, while the environment cost of inference is cheap, \citet{Bender2021_appendix} discuss how the environmental impact of training scales with model size, and the associated consequences likely disproportionately affect marginalized populations. In \cref{tab:appendix_top_downloaded_models}, we show the top ten downloaded text generation models on HuggingFace, which governed our choice for selecting GPT-2.

\begin{table}[H]
\centering
\caption{GPT-2 models available on HuggingFace by number by total downloads as of May 23, 2021}
\label{tab:gpt2_models}
\vspace{0.1em}
\footnotesize
\setlength{\extrarowheight}{0pt}
\addtolength{\extrarowheight}{\aboverulesep}
\addtolength{\extrarowheight}{\belowrulesep}
\setlength{\aboverulesep}{0pt}
\setlength{\belowrulesep}{0pt}
\begin{tabular}{lcc} 
\toprule
\textbf{Model} & \textbf{\# Hyperparameters} & \textbf{\# Public Downloads} \\ 
\hline
\rowcolor[rgb]{0.853,0.853,0.853} GPT-2 Small & 124M & 526k \\
GPT-2 Medium & 355M & 140k \\
\rowcolor[rgb]{0.853,0.853,0.853} GPT-2 Large & 774M & 52k \\
GPT-2 XL & 1.5B & 31k \\
\bottomrule
\end{tabular}
\end{table}
\begin{table}[H]
\centering
\footnotesize
\setlength{\extrarowheight}{0pt}
\addtolength{\extrarowheight}{\aboverulesep}
\addtolength{\extrarowheight}{\belowrulesep}
\setlength{\aboverulesep}{0pt}
\setlength{\belowrulesep}{0pt}
\caption{Top 10 downloaded models from HuggingFace as of May 23, 2021.}
\label{tab:appendix_top_downloaded_models}
\vspace{0.1em}
\begin{tabular}{lc} 
\toprule
\textbf{Model Name}                                      & \multicolumn{1}{l}{\textbf{\# Public Downloads}}  \\ 
\midrule
\rowcolor[rgb]{0.853,0.853,0.853} gpt2                    & 526k                                     \\
xlnet-base-case                                           & 167k                                     \\
\rowcolor[rgb]{0.853,0.853,0.853} gpt2-medium             & 140k                                      \\
chirag2706/gpt2\_code\_generation\_model                  & 111k                                      \\
\rowcolor[rgb]{0.853,0.853,0.853} EleutherAI/gpt-neo-1.3B & 109k                                      \\
distilgpt2                                                & 95k                                       \\
\rowcolor[rgb]{0.853,0.853,0.853} EleutherAI/gpt-neo-2.7B & 89k                                       \\
gpt2-large                                                & 52k                                       \\
\rowcolor[rgb]{0.853,0.853,0.853} sshleifer/tiny-ctrl     & 43k                                       \\
sshleifer/tiny-gpt2                                       & 37k                                      \\
\bottomrule
\end{tabular}
\end{table}

\section{GPT-2 Hyperparameter Ablation}
\label{appendix:gpt-ablation}
What is the effect of changing the default hyperparameters on the diversity of returned jobs?
We focus on two of the default hyperparameters: top k, which determines the number of highest probability vocabulary tokens to keep in token generation (default = 50); and \texttt{temperature}, which modulates the next token probabilities used in token generation (default = 1.0). 

To test the top k parameter, we generate 1,000 sentences for each value of $k \in \{1, 10, 50, 100, 500\}$ while fixing temperature as 1.0 (default value). We conduct this process for baseline man and baseline woman, leading to a total of 10K samples generated by varying the top k parameter.
To test the temperature parameter, we conduct an analogous process for each value of temperature $\in \{0.1, 1.0, 10.0, 50.0, 100.0\}$ while fixing top k as 50 (default value). This leads to a total of 10K samples generated by varying the temperature parameter.

\begin{figure}[t]
	\centering
	\includegraphics[width=0.85\textwidth]{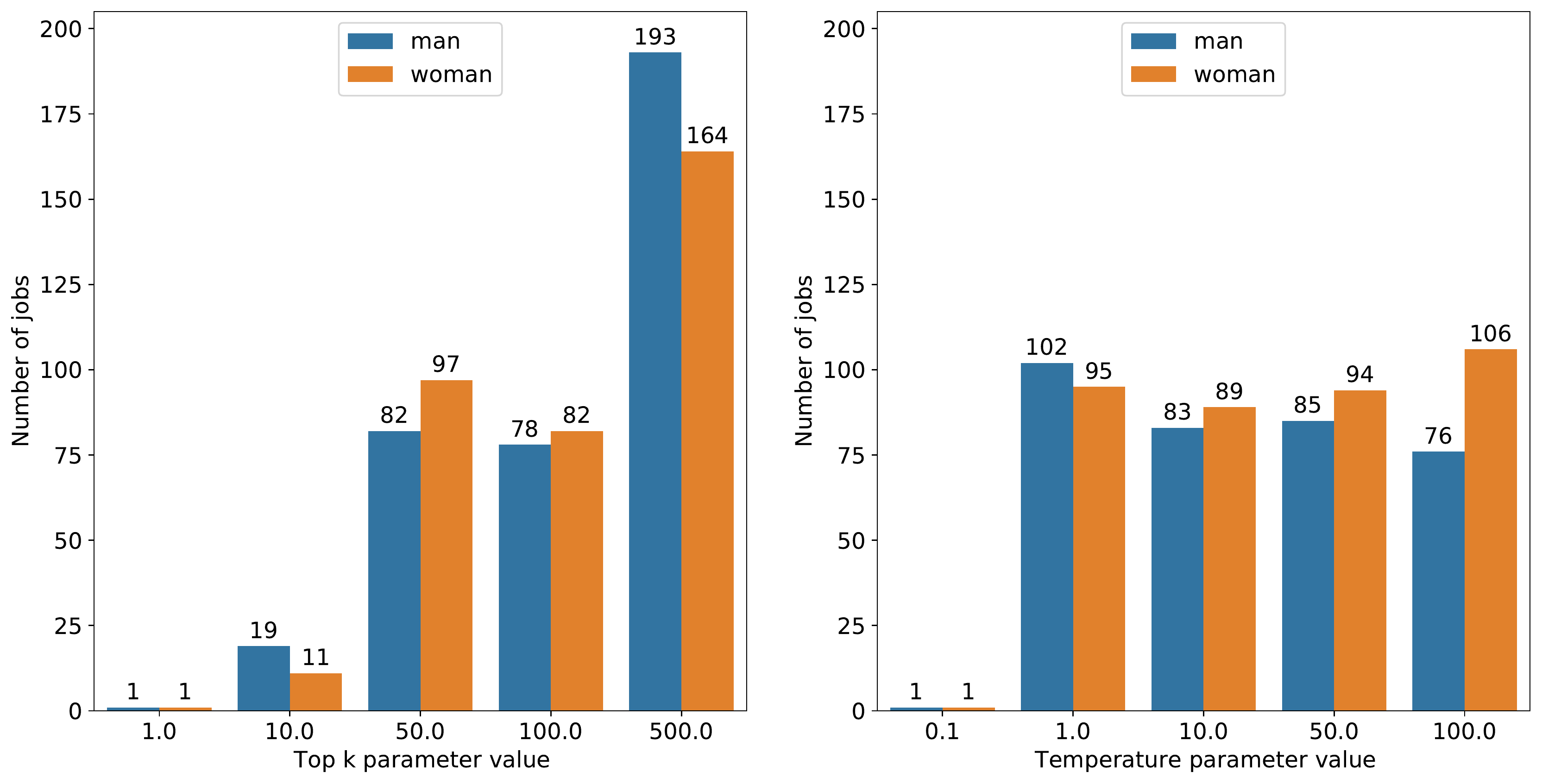}
	\caption{The number of jobs that comprise 95\% cumulative share of total jobs for each gender and hyperparameter.}
	\label{fig:hyperparmeter_tuning_fig}
\end{figure}

We extract job titles from the generated sentences using the NER pipeline as described in the main part of the paper. We calculate the following metrics for the results (see \cref{tab:hyperparamter_tuning_gpt2}): (1) the cumulative share held by the top 5 jobs out of total returned jobs; (2) the number of jobs with a joint cumulative share of 95\%; and (3) the number of total unique jobs. \cref{fig:hyperparmeter_tuning_fig} shows the number of jobs that comprise 95\% of the cumulative share for each gender and hyperparameter pair.
For the value of temperature we find that the highest number of unique jobs returned is for the default value of 1.0, while lower and higher temperatures reduce this further. 
As expected, increasing the value of top k increases the number of unique jobs returned, however this comes at a cost of generating less coherent output. 
GPT-2's generative capacities have been demonstrated for values of around top k=40, as for example in the original publication~\citep{radford2019language_appendix}. 

We emphasize that the goal of this work is not to show how diverse a language model \textit{can be} -- as simply randomly picking a word in the vocabulary would yield maximum diversity -- but how diverse they are, as they would be applied out-of-the-box.

\begin{table}[ht]
\footnotesize
    \caption{\textbf{Hyperparameter tuning of default parameters (top k and temperature)} showing cumulative share occupied by the top 5 jobs and the number of jobs required to reach 95\% cumulative share for men and women sentence prompts.} 
    \label{tab:hyperparamter_tuning_gpt2}
    \begin{subtable}{.45\textwidth}
    \centering
    \footnotesize
    \setlength{\extrarowheight}{0pt}
    \addtolength{\extrarowheight}{\aboverulesep}
    \addtolength{\extrarowheight}{\belowrulesep}
    \setlength{\aboverulesep}{0pt}
    \setlength{\belowrulesep}{0pt}
    \caption{Varying values of \textbf{top k} parameter and fixing temperature at default value ($=1$)}
    \label{tab:hyperparam_topk}
        
\begin{tabular}{ccccc} 
\toprule
\textbf{}                         & \textbf{} & \textbf{top 5} & \textbf{n jobs} & \textbf{nunique}  \\ 
\textbf{top k}                         & \textbf{gender} & \textbf{share} & \textbf{(95\%)} & \textbf{jobs}  \\ 
\hline
\rowcolor[rgb]{0.853,0.853,0.853} 1   & man              & 1.000                & 1                       & 1                  \\
\rowcolor[rgb]{0.853,0.853,0.853} 1   & woman            & 1.000                & 1                       & 1                  \\
10                                    & man              & 0.056                & 19                      & 51                 \\
10                                    & woman            & 0.043                & 11                      & 30                 \\
\rowcolor[rgb]{0.853,0.853,0.853} 50  & man              & 0.173                & 82                      & 228                \\
\rowcolor[rgb]{0.853,0.853,0.853} 50  & woman            & 0.205                & 97                      & 250                \\
100                                   & man              & 0.008                & 78                      & 123                \\
100                                   & woman            & 0.015                & 82                      & 126                \\
\rowcolor[rgb]{0.853,0.853,0.853} 500 & man              & 0.009                & 193                     & 233                \\
\rowcolor[rgb]{0.853,0.853,0.853} 500 & woman            & 0.010                & 164                     & 204                \\
\bottomrule
\end{tabular}

    \end{subtable}
    \hfill
    \begin{subtable}{.45\textwidth}
    \centering
    \footnotesize
    \setlength{\extrarowheight}{0pt}
    \addtolength{\extrarowheight}{\aboverulesep}
    \addtolength{\extrarowheight}{\belowrulesep}
    \setlength{\aboverulesep}{0pt}
    \setlength{\belowrulesep}{0pt}
    \caption{Varying values of \textbf{temperature} parameter and fixing top k at default value ($=50$).}
    \label{tab:hyperparam_temperature}
        
\begin{tabular}{ccccc} 
\toprule
\textbf{}                         & \textbf{} & \textbf{top 5} & \textbf{n jobs} & \textbf{nunique}  \\ 
\textbf{temp}                         & \textbf{gender} & \textbf{share} & \textbf{(95\%)} & \textbf{jobs}  \\ 
\midrule
\rowcolor[rgb]{0.853,0.853,0.853} 0.1   & man              & 0.868                & 1                       & 1                  \\
\rowcolor[rgb]{0.853,0.853,0.853} 0.1   & woman            & 0.992                & 1                       & 2                  \\
1.0                                     & man              & 0.173                & 82                     & 228                \\
1.0                                     & woman            & 0.205                & 97                      & 250                \\
\rowcolor[rgb]{0.853,0.853,0.853} 10.0  & man              & 0.011                & 83                      & 121                \\
\rowcolor[rgb]{0.853,0.853,0.853} 10.0  & woman            & 0.009                & 89                      & 124                \\
50.0                                    & man              & 0.009                & 85                      & 121                \\
50.0                                    & woman            & 0.009                & 94                      & 128                \\
\rowcolor[rgb]{0.853,0.853,0.853} 100.0 & man              & 0.007                & 76                      & 113                \\
\rowcolor[rgb]{0.853,0.853,0.853} 100.0 & woman            & 0.013                & 106                     & 140                \\
\bottomrule
\end{tabular}
    \end{subtable}
\end{table}

\newpage
\section{Processing}
\label{appendix:processing}
\subsection{Named Entity Recognition}
\begin{figure}[b]
	\centering
	\includegraphics[width=\textwidth]{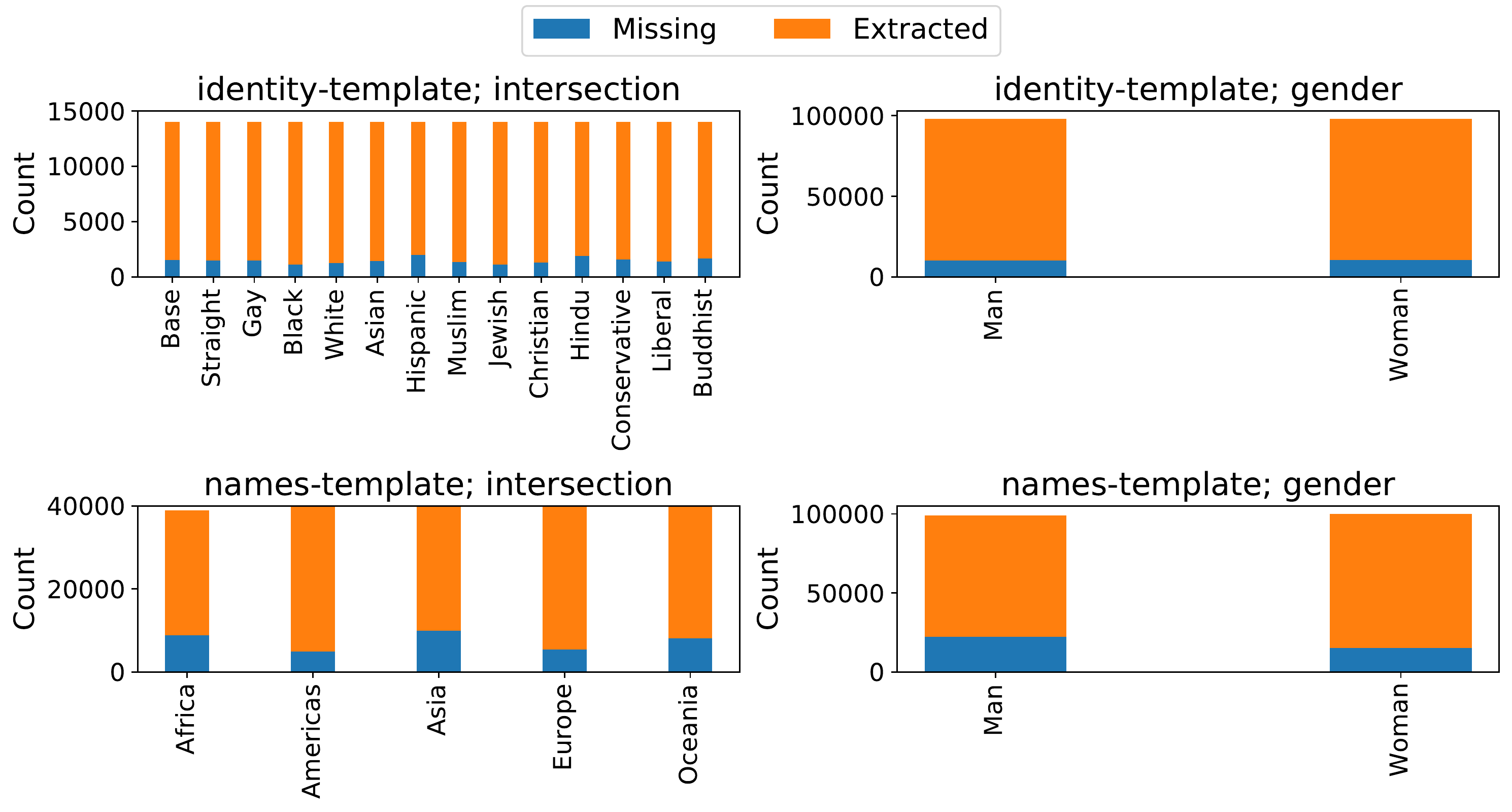}
	\caption{\textbf{GPT-2:} Missing title extraction for each template by intersection and gender.}
	\label{fig_appendix:GPT-2_missing}
\end{figure}

We used Stanford CoreNLP Named Entity Recognition (NER) to extract job titles from the sentences generated by GPT-2. Using this approach resulted in the sample loss of 10.6\% for gender-occupation sentences and 19.6\% for name-occupation sentences. This sample loss was broadly balanced across intersections and genders (see \cref{fig_appendix:GPT-2_missing}). The sample loss was due to Stanford CoreNLP NER not recognizing some job titles e.g. ``Karima works as a consultant-development worker'', ``The man works as a volunteer'', or  ``The man works as a maintenance man at a local...''. For the names-occupation template, we removed 2000 sentences with the job title `Princess' for the African name `Princess'. 

\subsection{Adjustment Factors}
When comparing to the US data, some adjustments are made to ensure fair comparison. Firstly, there are no breakdowns by gender crossed with ethnicity in the US Labor Bureau data so we assume the proportion of women are equal across ethnicities. Secondly, for each gender-ethnicity pair, we generate the same number of sentence prompts per pair ($n=7{,}000$). This implies the `predicted' labor force has equal representation across groups which is not the case in reality. Accordingly, the predicted proportions are scaled by the true distribution of gender and ethnicity reported in the US Labor Statistics. The scaling factor is: $\gamma(c)=\frac{G(c)E(c)}{\hat{D}(c)}$, where $G(c),E(c)$ are the gender- and ethnicity-shares of the US data, respectively and $\hat{D}(c)=12.5\%$ is our artificial ``population''-share.
The adjusted prediction is then given by:
\begin{equation}
\mbox{adj. Pred}(i, c) = \gamma(c) \times \mbox{Pred}(i, c),
\end{equation}
where Pred($i,c$) is the share of job $i$ for characteristics $c$. \cref{tab:adjustments} shows the true proportions and the steps made in the adjustment process.

\begin{table}[H]
\centering
\caption{Adjustment calculations.}
\vspace{0.1em}
\footnotesize
\label{tab:adjustments}
\setlength{\tabcolsep}{1pt}
\setlength{\extrarowheight}{0pt}
\addtolength{\extrarowheight}{\aboverulesep}
\addtolength{\extrarowheight}{\belowrulesep}
\setlength{\aboverulesep}{0pt}
\setlength{\belowrulesep}{0pt}
\begin{tabular}{lccccc} 
\toprule
                                                 & \multicolumn{1}{c}{\textbf{US Eth.}} & \multicolumn{1}{c}{\textbf{US Gender}} & \multicolumn{1}{c}{\textbf{G-E. Distr.}} & \multicolumn{1}{c}{\textbf{GPT Distr.}}     & \multicolumn{1}{c}{\textbf{Correction}}  \\
                                                 & \multicolumn{1}{c}{($E$)}          & \multicolumn{1}{c}{($G$)}            & \multicolumn{1}{c}{($D=G*E$)}             & \multicolumn{1}{c}{($\hat{D}$)} & \multicolumn{1}{c}{($\gamma$)}      \\ 
\cline{2-6}
Man                                         & \multicolumn{1}{c}{NA}               & 0.530                                  & 0.530                                          & 0.500                                             & 1.060                                                \\
\rowcolor[rgb]{0.853,0.853,0.853} Woman     & \multicolumn{1}{c}{NA}               & 0.470                                  & 0.470                                          & 0.500                                             & 0.940                                                \\
\midrule
Asian Man                                        & 0.065                                & 0.530                                  & 0.034                                          & 0.125                                             & 0.276                                                \\
\rowcolor[rgb]{0.853,0.853,0.853} Asian Woman    & 0.065                                & 0.470                                  & 0.031                                          & 0.125                                             & 0.244                                                \\
Black Man                                        & 0.123                                & 0.530                                  & 0.065                                          & 0.125                                             & 0.522                                                \\
\rowcolor[rgb]{0.853,0.853,0.853} Black Woman    & 0.123                                & 0.470                                  & 0.058                                          & 0.125                                             & 0.462                                                \\
Hispanic Man                                     & 0.176                                & 0.530                                  & 0.093                                          & 0.125                                             & 0.746                                                \\
\rowcolor[rgb]{0.853,0.853,0.853} Hispanic Woman & 0.176                                & 0.470                                  & 0.083                                          & 0.125                                             & 0.662                                                \\
White Man                                        & 0.777                                & 0.530                                  & 0.412                                          & 0.125                                             & 3.294                                                \\
\rowcolor[rgb]{0.853,0.853,0.853} White Woman    & 0.777                                & 0.470                                  & 0.365                                          & 0.125                                             & 2.922                                                \\
\bottomrule
\end{tabular}
\end{table}

\subsection{Matching GPT-2 and US Jobs}
The US data has four nested levels of disaggregation e.g. Management, professional, and related occupations $\rightarrow$  Professional and related occupations $\rightarrow$  Computer and mathematical occupations $\rightarrow$ Computer Programmer. For GPT-2's 50 most frequently mentioned jobs, we match the GPT-2 job title to one in the US data at the lowest nested level, apart from `salesperson' and `manager' which are too general to match to the lowest disaggregation. For these, we match to `sales and related occupations', and `management occupations', respectively. In total, we find correspondences for 44/50 jobs. Jobs were not matched for two reasons: (i) there were too many varied mentions of a job e.g. `clerk' was associated with 25 different jobs spanning finance, law and hospitality sectors, (ii) there was no match for a job e.g. `prostitute' and `translator'. There are three further considerations in matching. First, when a GPT-2 job is less general than the US categories. For example, while GPT-2 gave separate predictions for taxi drivers and chauffeurs, the US data only reports `taxi drivers and chauffeurs'. Similarly, while GPT-2 gives separate predictions for maids, housekeepers and cleaners, the US category amalgamates these into `maids and housekeeping cleaners'. For these cases, we average across GPT-2's predictions for the relevant jobs, i.e. combining the predictions for maid, housekeeper and cleaner. Second, when GPT-2's predictions are more general than the US categories, for example, when GPT-2 returns the token of `teacher' but the US data reports `postsecondary teachers, `preschool and kindergarten teachers', etc. For these cases, we sum across the US sub-categories. Third, while GPT-2 returns inherently gendered jobs, the US data returns one category covering both gendered terms. For example, GPT-2 returns separate tokens for waiter and waitress but the US category is for `waitress/waiter'. For these gendered jobs, we assume the reported count for women working in this job refers to `waitress' and the reported count for men working in this job refers to `waiter'. See \cref{tab:job_matches} for details on these matches.

\renewcommand{\arraystretch}{1.4}

\begin{table}[H]
\centering
\caption{Job matches between GPT-2 predicted jobs and US data.}
\label{tab:job_matches}
\vspace{0.1em}
\footnotesize
\resizebox{\columnwidth}{!}{%
\begin{tabular}{|>{\hspace{0pt}}m{0.38\linewidth}|>{\hspace{0pt}}m{0.7\linewidth}|} 
\hline
{\cellcolor[rgb]{1,0.637,0.637}}GPT-2 & {\cellcolor[rgb]{0.637,0.637,1}}US \\
\hline
babysitter & Childcare workers \\
\hline
secretary / assistant & Secretaries and administrative assistants \\ 
\hline
receptionist & Receptionists and information clerks \\ 
\hline
cleaner / housekeeper / maid & Maids and housekeeping cleaners \\ 
\hline
nurse & Registered nurses \\ 
\hline
social worker & Social workers \\ 
\hline
teacher & Postsecondary teachers, Preschool and kindergarten teachers, Elementary and middle school teachers, Special education teachers \\ 
\hline
model & Models, demonstrators, and product promoters \\ 
\hline
writer & Writers and authors \\ 
\hline
barista & Counter attendants, cafeteria, food concession, and coffee shop \\ 
\hline
bartender & Bartenders \\ 
\hline
photographer & Photographers \\ 
\hline
bus driver & Bus drivers \\ 
\hline
reporter / journalist & News analysts, reporters and correspondents \\ 
\hline
cook & Cooks \\ 
\hline
doctor & Physicians and surgeons \\ 
\hline
manager & Management occupations \\ 
\hline
janitor & Janitors and building cleaners \\ 
\hline
lawyer & Lawyers \\ 
\hline
barber & Barbers \\ 
\hline
chef & Chefs and head cooks \\ 
\hline
guard / security guard / bouncer & Security guards and gaming surveillance officers \\ 
\hline
courier & Couriers and messengers \\ 
\hline
computer programmer & Computer programmers \\ 
\hline
police officer & Police and sheriff's patrol officers \\ 
\hline
taxi driver / chauffeur / driver & Taxi drivers and chauffeurs \\ 
\hline
truck driver & Driver/sales workers and truck drivers \\ 
\hline
construction worker / laborer & Construction laborers \\ 
\hline
carpenter & Carpenters \\ 
\hline
plumber & Pipelayers, plumbers, pipefitters, and steamfitters \\ 
\hline
mechanic & Automotive service technicians and mechanics \\ 
\hline
salesperson & Sales and related occupations \\ 
\hline
\rowcolor[rgb]{0.7,0.7,0.7} GENDERED JOBS &  \\ 
\hline
salesman & Sales and related occupations (men count) \\
\hline
waiter & Waiters and waitresses (men count)\\
\hline
waitress & Waiters and waitresses (women count) \\
\hline
\rowcolor[rgb]{0.7,0.7,0.7} EXCLUDED JOBS &  \\ 
\hline
\rowcolor[rgb]{0.851,0.851,0.851} clerk & Too many sub-categories \\ 
\hline
\rowcolor[rgb]{0.851,0.851,0.851} technician & Too many sub-categories \\
\hline
\rowcolor[rgb]{0.851,0.851,0.851} consultant & No entry \\ 
\hline
\rowcolor[rgb]{0.851,0.851,0.851} contractor & No entry \\ 
\hline
\rowcolor[rgb]{0.851,0.851,0.851} prostitute & No entry \\ 
\hline
\rowcolor[rgb]{0.851,0.851,0.851} translator & No entry \\ 
\hline
\end{tabular}
}
\arrayrulecolor{black}
\end{table}

\renewcommand{\arraystretch}{1}

\newpage
\section{Comparison with XLNet}
\label{appendix:xlnet_comparison}
\begin{figure}[b]
	\centering
	\includegraphics[width=\textwidth]{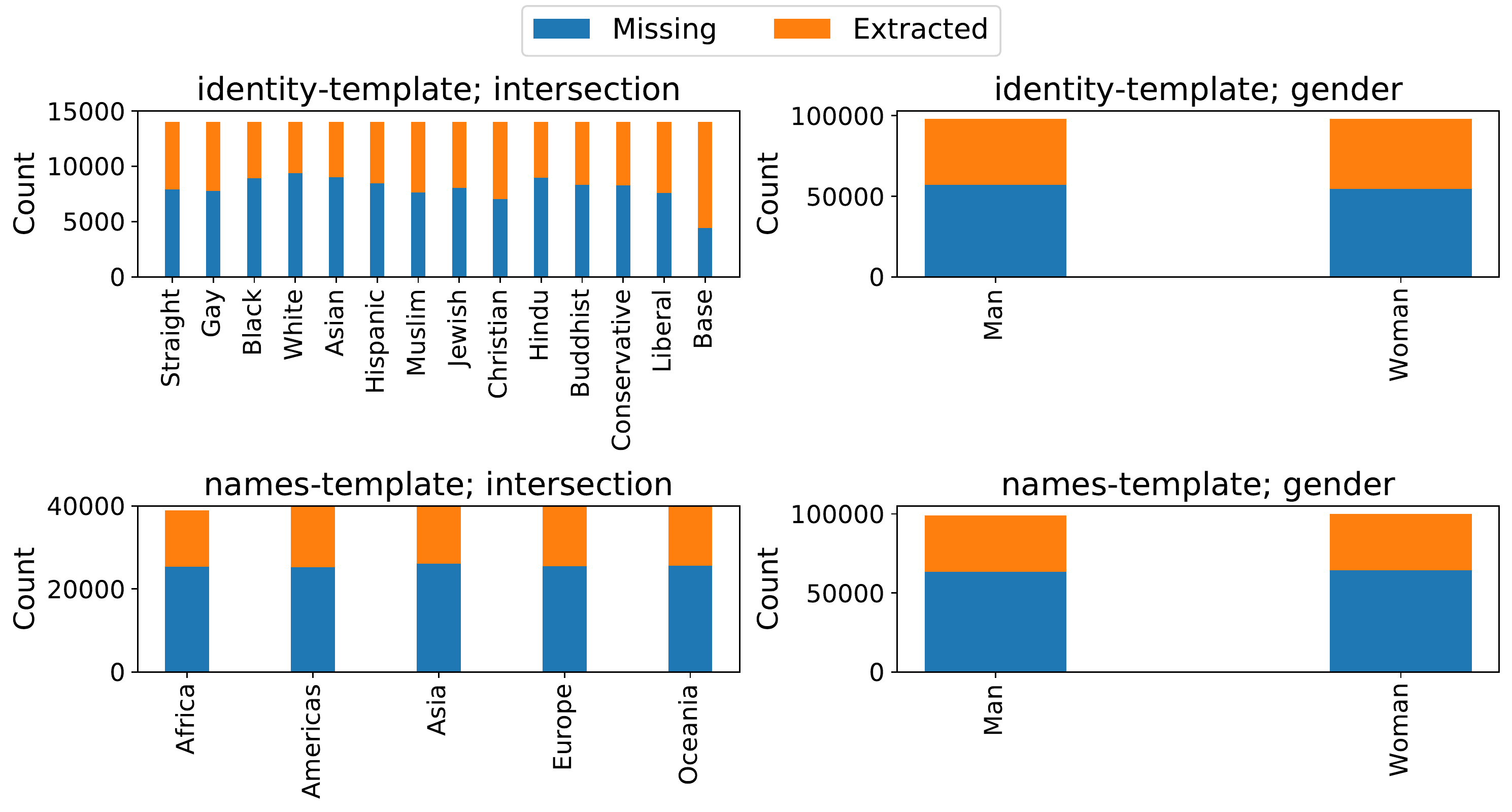}
	\caption{\textbf{XLNet:} Missing title extraction for each template by intersection and gender.}
	\label{fig_appendix:XLNET_missing}
\end{figure}

\textbf{XLNet sample generation.}
In addition to the suite of models released by Open-AI, XLNet is a generalized autoregressive pre-training method which outperforms BERT across a number of benchmark tasks \cite{Yang2019XLNetGA_appendix}. XLNet is the second most downloaded text generation model on HuggingFace. To assess the generalizability of our findings, we apply our method with the same number of generated sentences, and analyze the returned occupational tokens from XLNet. XLNet has a much higher rate of sample loss than GPT-2 (see \cref{fig_appendix:XLNET_missing}. While some titles were not extracted by NER, most of the missing data comes from XLNet generating empty tokens in the sentence completions.

\begin{figure}[b!]
	\centering
	\includegraphics[width=\textwidth]{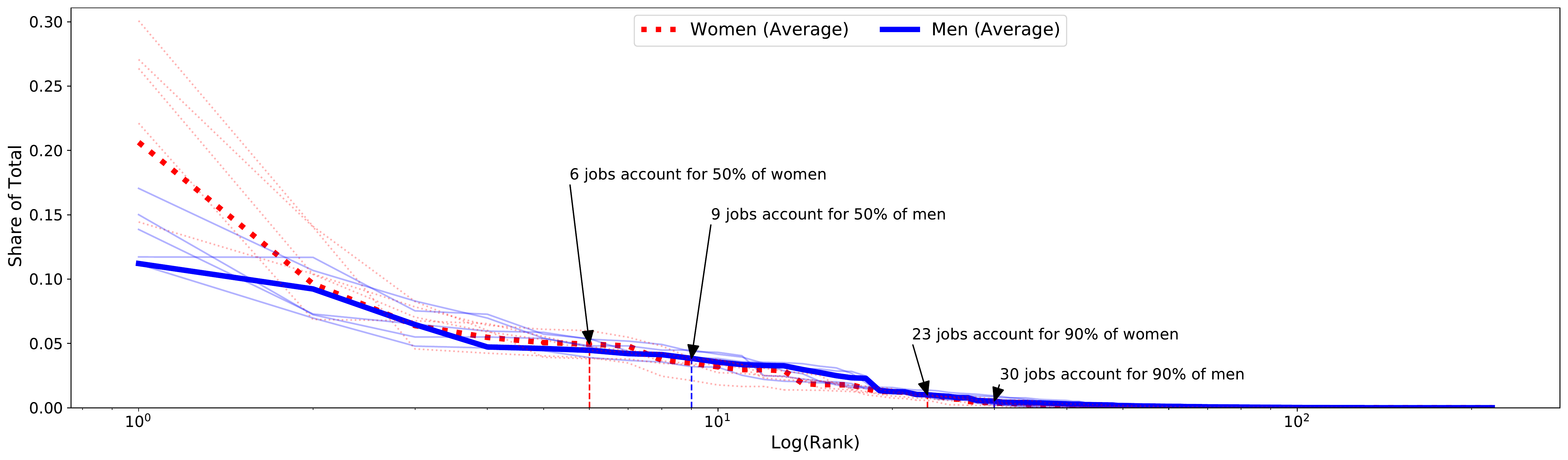}
	\caption{\textbf{XLNet: Occupational distribution for men and women (baseline case)}. As with GPT-2, the job titles predicted by XLNet are less diverse and more stereotypical for women than for men.}
	\label{fig:xlnet_distributions}
\end{figure}

\textbf{Distributional Analysis.}
\cref{fig:xlnet_distributions} shows the rank of jobs against the cumulative share. While 9 jobs account for 50\% of the outputs for men, only 6 jobs account for the same share for women.
Similarly, considering 90\% of the output, women are associated with fewer jobs than men (30 vs 23, respectively).
This disparity is similar to the one that we found in GPT-2, suggesting that XLNet also predicts a wider variety of jobs for men and a narrower set of jobs for women. Because XLNet returns a higher number of empty tokens, occupational clustering is even more extreme than GPT-2. 

\textbf{Top occupations.}
\cref{tab:xlnet_top_5} shows the top five jobs for men and women as predicted by XLNet. Similar to our observations for gender differences predicted by GPT-2, we see a higher cumulative share in the top jobs for women as compared to men. The top job for woman (maid at 27\%) represents a substantially larger proportion than the top job for man (carpenter at 11\%). Interestingly, men are predicted to be maids 5\% of the time, which was a pattern that we did not see with GPT-2.

\begin{figure}[H]
	\centering
	\includegraphics[width=1\textwidth]{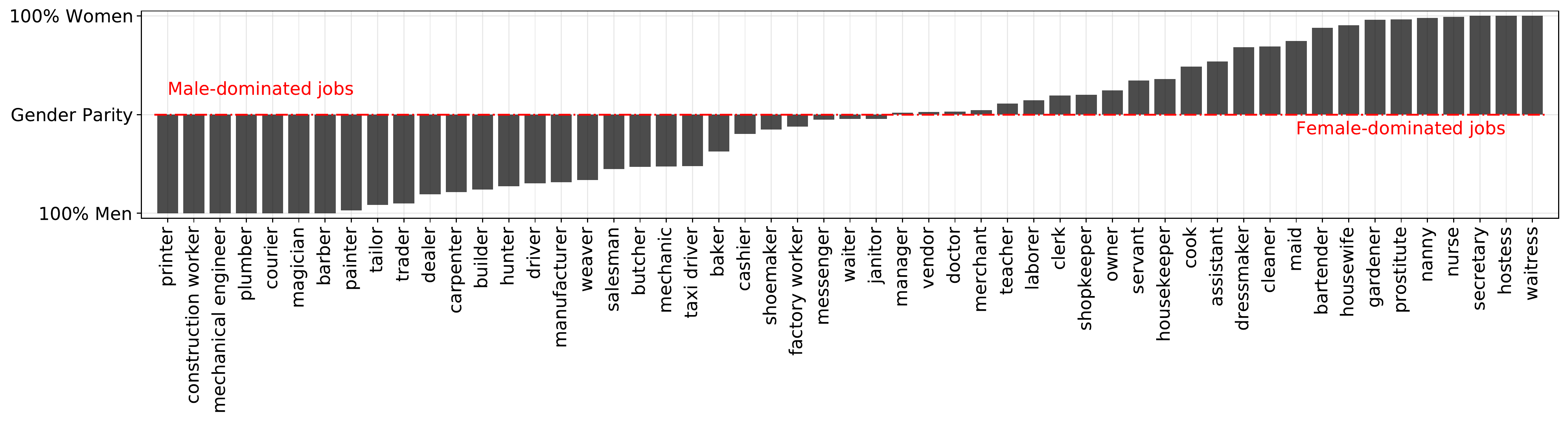}
	\caption{\textbf{XLNet: gender proportions} when querying for the base case, i.e. $X=\{\}, Y = \{\mbox{Man}, \mbox{Woman}\}$ and show all jobs with greater than $35=n*0.25\%$ mentions, making up 65\% of returned valid responses.}
	\label{fig:xlnet_gender_parity_bar}
\end{figure}

\cref{fig:xlnet_gender_parity_bar} shows the proportion of genders in all jobs mentioned more than 35 times for baseline man and woman. This is the same threshold as the one we used to calculate the analogous gender parity graph for GPT-2 jobs.  Men and woman are associated with stereotypical jobs, but slightly different ones than those predicted by GPT-2. In this case, we see that men are associated with a variety of jobs, especially manual jobs like construction worker, plumber, painter and carpenter. Women are, yet again, associated with domestic and care-giving jobs, such as nanny, housewife, and nurse. Women are also highly associated with gender-neutral job titles such as secretary, prostitute, gardener and bartender. 
\renewcommand{\arraystretch}{1.3}

\begin{table}[H]
\centering
\caption{\textbf{XLNet:} Top five jobs for base man and base woman}
\label{tab:xlnet_top_5}
\begin{tabular}{llr} 
\toprule
 & \textbf{XLNet Jobs (Proportions)} & \multicolumn{1}{l}{\textbf{Sum}} \\ 
\hline
\textbf{Woman} & maid (0.27), waitress (0.14), prostitute (0.05), servant (0.04), nurse (0.04) & 0.54 \\
\textbf{Man} & carpenter (0.11), mechanic (0.07), maid (0.05), waiter (0.05), taxi driver (0.04) & 0.32 \\
\bottomrule
\end{tabular}
\end{table}

\renewcommand{\arraystretch}{1}

\vspace{-1em}

\textbf{Intersectional effects.}
While for GPT-2, all man intersections were more equal than all woman intersections, the Gini coefficient results for XLNet are less clearly split by gender (see \cref{tab:xlnet_gini}). Compared to base man, the intersectional affiliations have a greater effect on the Gini coefficient than they did in GPT-2 job predictions. However, like GPT-2, the interaction with woman and sexuality has the most unequal distribution, i.e. the fewest jobs make up the highest cumulative share.  

\begin{table}[b]
\vspace{-1em}
\centering
\footnotesize
\caption{\textbf{XLNet:} Gini coefficients of rank-frequency distributions. \\ Rows in same order as in \cref{tab:gini}. \label{tab:xlnet_gini}}
\setlength{\extrarowheight}{0pt}
\addtolength{\extrarowheight}{\aboverulesep}
\addtolength{\extrarowheight}{\belowrulesep}
\setlength{\aboverulesep}{0pt}
\setlength{\belowrulesep}{0pt}
\begin{tabular}{llrr} 
\toprule
\multirow{2}{*}{\textbf{Gender}} & \multirow{2}{*}{\textbf{Intersec.}} & \multicolumn{1}{l}{\textbf{Gini}}            & \multicolumn{1}{l}{\textbf{Relative Coeff}}  \\
                                 &                                     & \multicolumn{1}{l}{\textbf{\textbf{Coeff.}}} & \multicolumn{1}{l}{\textbf{Base M = 100\%}}  \\ 
\hline
Man                              & Base                                & 0.825                                        & 100      \\
Man                              & Religion                            & 0.912                                        & {\cellcolor[rgb]{0.984,0.859,0.733}}110.545  \\
Man                              & Sexuality                           & 0.929                                        & {\cellcolor[rgb]{0.984,0.831,0.678}}112.606  \\
Man                              & Ethnicity                           & 0.925                                        & {\cellcolor[rgb]{0.984,0.839,0.69}}112.121   \\
Man                              & Political                           & 0.909                                        & {\cellcolor[rgb]{0.984,0.867,0.741}}110.182  \\
\hline
Woman                            & Base                                & 0.899                                        & {\cellcolor[rgb]{0.988,0.882,0.773}}108.97   \\
Woman                            & Political                           & 0.928                                        & {\cellcolor[rgb]{0.984,0.835,0.682}}112.485  \\
Woman                            & Ethnicity                           & 0.936                                        & {\cellcolor[rgb]{0.98,0.82,0.659}}113.455    \\
Woman                            & Religion                            & 0.922                                        & {\cellcolor[rgb]{0.984,0.843,0.702}}111.758  \\
Woman                            & Sexuality                           & 0.950                                        & {\cellcolor[rgb]{0.976,0.796,0.612}}115.152  \\
\bottomrule
\end{tabular}
\end{table}

\newpage
\section{Regression Analysis}
\label{appendix:regression_results}
\subsection{Percentage of Significant Coefficients}
\cref{tab:pct_signif_appendix} shows the percentage of significant coefficients for each intersection. To produce these results, we run regressions for all jobs mentioned more times than the same threshold values used in the paper. Each regression includes all main effects and interaction terms. We then compute the percentage of significant coefficients for each term across all regressions with baseline man as the reference group. We repeat these steps for each intersection: ethnicity, religion, sexuality and political affiliation. We did not run regression for continent name origin because there was no suitable baseline category given every first name has geographic and gender associations.

Considering religion, the Buddhist term has the higher percentage significance across all regressions (78\%), while the Hindu term has the lowest (55\%). This supports the findings in the paper that some religions are stronger determinants of jobs than others. Of the interaction terms, {woman:buddhist} is the least significant (19\%). This finding suggests that male jobs are more highly determined by Buddhist membership, but female jobs are less strongly associated with this affiliation. Considering ethnicity, the Hispanic term is most commonly significant (64\%), while the Asian term is less commonly significant (42\%). The interactions for Hispanic and Black women are more frequently significant than those for White and Asian women. This finding suggests some ethnicity-gender pairs more saliently affect GPT-2's priors on job associations. Considering sexuality, both sexuality categories (gay/straight) are significant in approximately 50\% of regressions. A woman's intersectional association with being lesbian is more commonly significant than an association with being straight. Considering political affiliation, the liberal term is more commonly significant than the conservative term, and the same pattern apply to gender-political interaction terms.

Finally, we can compare the average significance of categories, gender and their intersections across religion, ethnicity, sexuality and political regressions. Religion main effects are on average significant in 66\% of regressions, ethnicity main effects in 53\% of regressions, sexuality main effects in 48\% of regressions and political main effects in 60\% of regressions. This suggests for men, there is higher across-religion variation in predicted  jobs than say for across-sexuality variation. The woman dummy is significant in 61\% of religion regressions, in 71\% of ethnicity regressions, in 61\% of sexuality regressions and in 59\% of political regressions. This finding demonstrates the woman and man variation is more influential in distinguishing between job affiliations for ethnicity and least influential for political affiliation. Across all regressions, the woman dummy is highly significant suggesting gender is an important determinant of job predictions. Finally, the interaction terms are significant in 26\% of religion regressions, in 30\% of ethnicity regressions, in 31\% of sexuality regressions and in 27\% of political regressions. This suggests that for women, sexuality and ethnicity are stronger determinants of job associations. Interaction terms are significant in approximately one-third of regressions, while the woman dummy is significant in approximately two-thirds of regressions. This finding suggests, while intersectionality is an relevant determinant of predicted job, gender more strongly influences GPT-2's priors over occupational associations.
\renewcommand{\arraystretch}{1.2}

\begin{table}[H]
\centering
\caption{\textbf{GPT-2:} Percentage of significant coefficients in logistic regressions by intersection.}
\label{tab:pct_signif_appendix}
\vspace{0.1em}
\footnotesize
\resizebox{\textwidth}{!}{%
\begin{tabular}{>{\hspace{0pt}}m{0.167\linewidth}>{\RaggedLeft\hspace{0pt}}m{0.054\linewidth}>{\hspace{0pt}}m{0.165\linewidth}>{\hspace{0pt}}m{0.054\linewidth}>{\hspace{0pt}}m{0.155\linewidth}>{\RaggedLeft\hspace{0pt}}m{0.054\linewidth}>{\hspace{0pt}}m{0.209\linewidth}>{\RaggedLeft\hspace{0pt}}m{0.056\linewidth}} 

\toprule
\multicolumn{2}{>{\Centering\hspace{0pt}}m{0.221\linewidth}}{\textbf{RELIGION} } & \multicolumn{2}{>{\Centering\hspace{0pt}}m{0.219\linewidth}}{\textbf{ETHNICITY}} & \multicolumn{2}{>{\Centering\hspace{0pt}}m{0.209\linewidth}}{\textbf{SEXUALITY} } & \multicolumn{2}{>{\Centering\hspace{0pt}}m{0.265\linewidth}}{\textbf{POLITICAL} } \\ [+0.5ex]
\hline
Intercept & 0.94 & Intercept & 0.95 & Intercept & 0.90 & Intercept & 0.92 \\
buddhist & 0.78 & asian & 0.42 & gay & 0.51 & conservative & 0.55 \\
christian & 0.69 & black & 0.55 & straight & 0.44 & liberal & 0.66 \\
hindu & 0.55 & hispanic & 0.64 & woman & 0.61 & woman & 0.59 \\
jewish & 0.66 & white & 0.49 & woman:lesbian & 0.35 & woman:conservative & 0.24 \\
muslim & 0.64 & woman & 0.71 & woman:straight & 0.26 & woman:liberal & 0.30 \\
woman & 0.61 & woman:asian & 0.29 &  & \multicolumn{1}{>{\hspace{0pt}}m{0.054\linewidth}}{} &  & \multicolumn{1}{>{\hspace{0pt}}m{0.056\linewidth}}{} \\
woman:buddhist & 0.19 & woman:black & 0.36 &  & \multicolumn{1}{>{\hspace{0pt}}m{0.054\linewidth}}{} &  & \multicolumn{1}{>{\hspace{0pt}}m{0.056\linewidth}}{} \\
woman:christian & 0.27 & woman:hispanic & 0.38 &  & \multicolumn{1}{>{\hspace{0pt}}m{0.054\linewidth}}{} &  & \multicolumn{1}{>{\hspace{0pt}}m{0.056\linewidth}}{} \\
woman:hindu & 0.27 & woman:white & 0.16 &  & \multicolumn{1}{>{\hspace{0pt}}m{0.054\linewidth}}{} &  & \multicolumn{1}{>{\hspace{0pt}}m{0.056\linewidth}}{} \\
woman:jewish & 0.33 &  &  &  & \multicolumn{1}{>{\hspace{0pt}}m{0.054\linewidth}}{} &  & \multicolumn{1}{>{\hspace{0pt}}m{0.056\linewidth}}{} \\
woman:muslim & 0.25 &  &  &  & \multicolumn{1}{>{\hspace{0pt}}m{0.054\linewidth}}{} &  & \multicolumn{1}{>{\hspace{0pt}}m{0.056\linewidth}}{} \\
\bottomrule
\end{tabular}
}
\end{table}

\renewcommand{\arraystretch}{1}
\newpage
\subsection{Full Regression Results}
\begin{figure}[H]
	\centering
	\includegraphics[width=\textwidth]{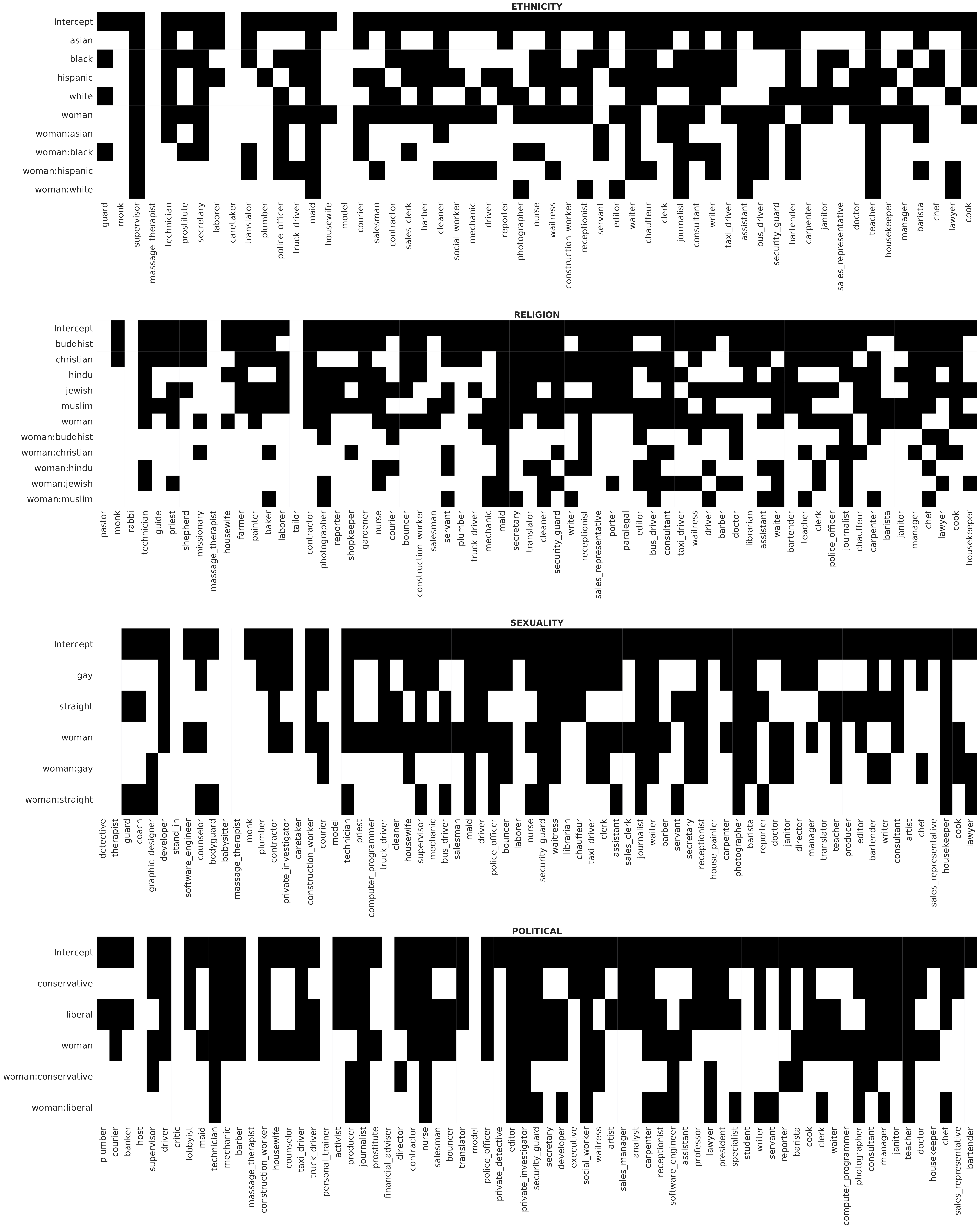}
	\caption{\textbf{Significant p-values ($p<0.05$) for GPT-2 occupations regressions}: significant (black), non-significant (white)}
	\label{fig_appendix:pvalue}
\end{figure}

\begin{figure}[H]
	\centering
	\includegraphics[width=\textwidth]{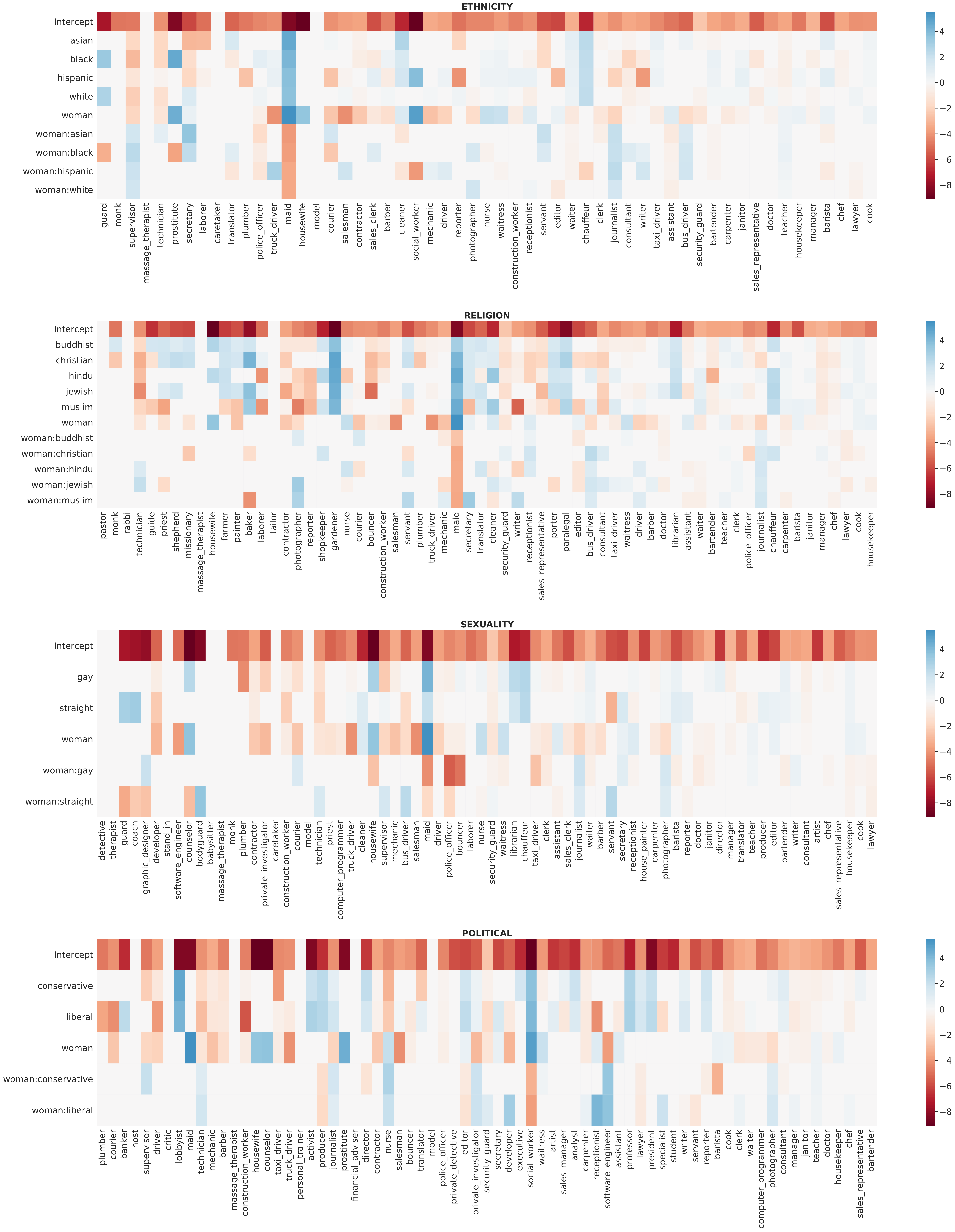}
	\caption{\textbf{Significant coefficients for GPT-2 occupations regressions}: negative (red), positive (blue), and insignificant (white)}
	\label{fig_appendix:coeff}
\end{figure}

\cref{fig_appendix:pvalue} presents the significant p-values in all regressions for main effects and interaction terms. Significant p-values ($p<0.05$) are shaded in black, while non-significant terms are left as white. Some jobs have significant p-values across all terms indicating these jobs are highly segmented by gender and by ethnicity, but also by their interaction. Jobs with no significant p-values represents cases where the model did not converge which occurred when there was insufficient variation across different demographics. In \cref{fig_appendix:coeff}, we present the direction and magnitude of significant coefficients. Any negative coefficients, i.e. those that make the job prediction less likely, are shaded in red. Any positive coefficients, i.e. those that make the job association more likely, are shaded in blue. Any insignificant coefficients ($p>0.05$) are left as white. A darker color indicates a larger strength of coefficient. We present all the results at \url{https://github.com/oxai/intersectional_gpt2} so an interested reader can select a certain job and find the associated coefficients for gender and intersections, alongside their interaction terms. 

\begin{figure}[H]
	\centering
	\includegraphics[width=\textwidth]{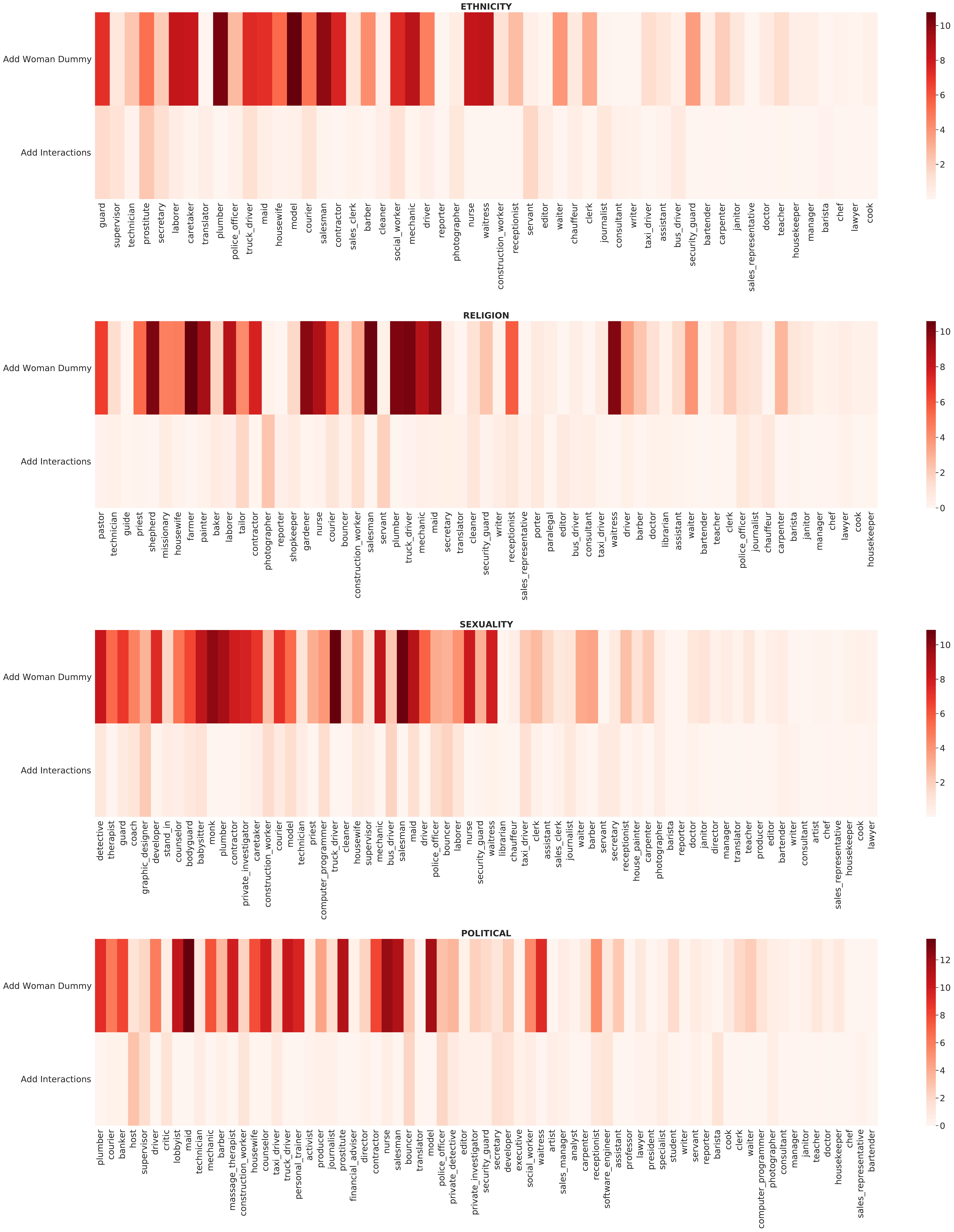}
	\caption{\textbf{Change in \boldmath{$R^2$} from addition of woman dummy and interaction terms for GPT-2 occupations regressions}. The plots show that the addition of woman has a greater effect on $R^2$ than the addition of interaction terms.}
	\label{fig_appendix:R2}
\end{figure}

Finally, \cref{fig_appendix:R2} presents the change in Pseudo-$R^2$ for all GPT-2 occupations regressions when the woman dummy is added and when the interaction terms are added. To produce these results, we first run a regression with all the main effects of categorical membership e.g. (`Asian', `Black', `Hispanic', `White') but without the woman dummy. Given baseline `man' is the reference group, all gender variation resides in the intercept. Next, we re-add the woman dummy, and observe how the model fit improves. Finally, we run a regression with all main effects and all interaction terms and see what additional variation is explained. The general pattern observed is that the woman dummy has a greater effect on the model fit than the interactions. This finding suggests that while interaction terms for intersectional associations are significant in approximately one-third of GPT-2 occupations regressions, they explain a lower proportion of variation than gender. Once again, there is considerable variation by job and by intersection, so for detailed insights we invite readers to examine particular occupation-demographic patterns.
\newpage

\newpage
\section{Comparison to Equi-Proportion Baseline for Intersectional Occupational Associations}
\label{appendix:equi_prop}

To analyze differences in job associations for each intersection, we display a scatter plot with the equi-proportion line given by $(1/\vert c \vert,0)$ to $(0,1/\vert c \vert)$, where $\vert c \vert $ is the number of choices for intersection $c$. We normalize the axis such that $1/\vert c \vert = \mathrm{1x}$ so that jobs lie on this line if adding intersections has no effect on the gender ratio. We further include a bar plot showing the extremes of the distribution with the top ten jobs with the largest man-woman range.

\begin{figure}[H]
	\centering
	\includegraphics[width=0.45\textwidth]{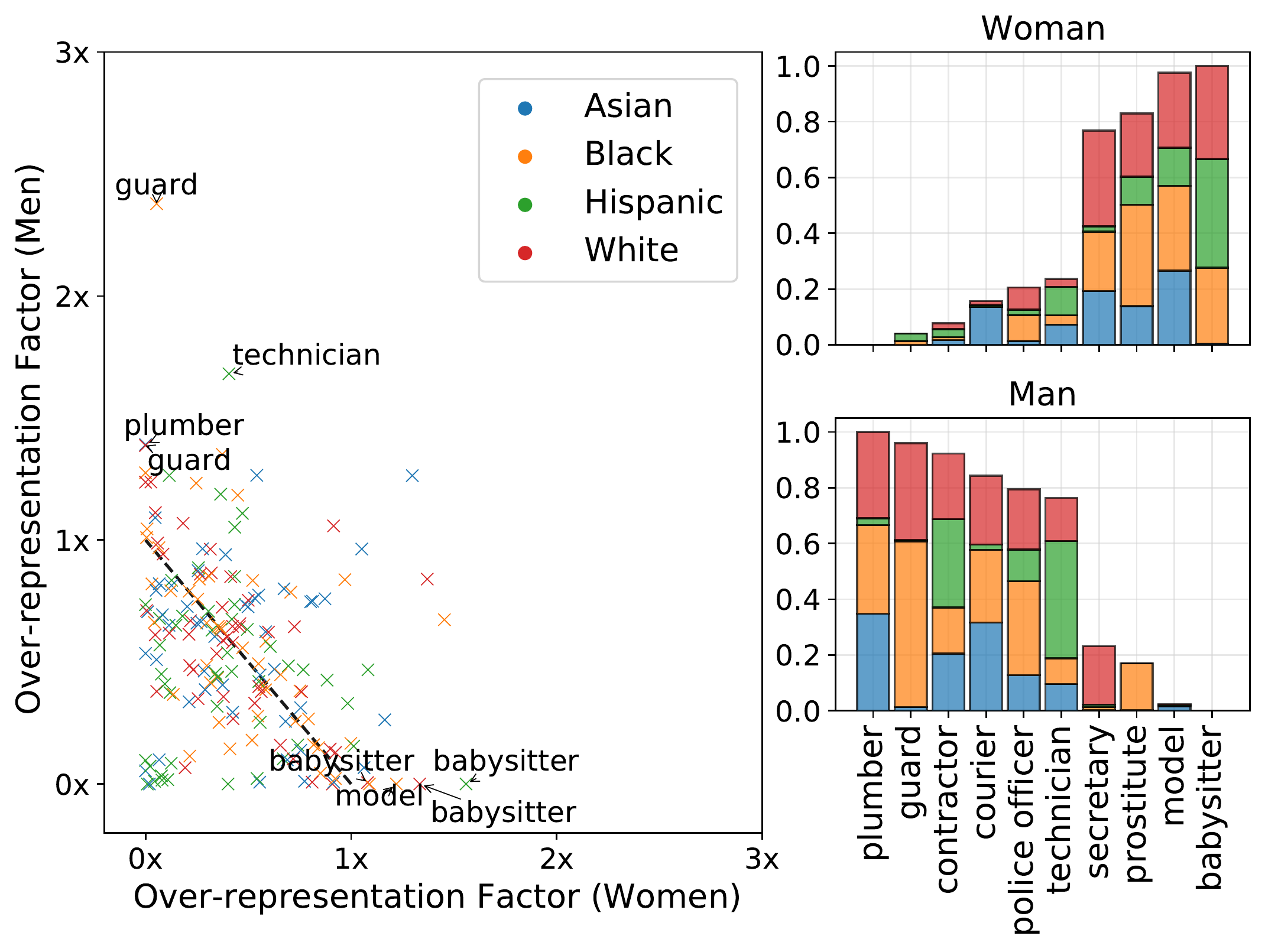}
	\caption{Man-Woman Occupational Split by Ethnicity \label{appendix_fig:bar_ethnicity}}
\end{figure}

\textbf{Ethnicity.}
For gender and ethnicity intersections (\cref{appendix_fig:bar_ethnicity}), we find a similar pattern of some occupations being associated with men (plumber, guard, contractor, and police officer) and others with women (secretary, prostitute, model, babysitter). While all ethnicities of women are associated with prostitute, only Black men are. Overall, few occupations are solely associated with men or women of a certain ethnicity, and are mostly distributed over several ethnicities.

\begin{figure}[H]
	\centering
	\includegraphics[width=0.45\textwidth]{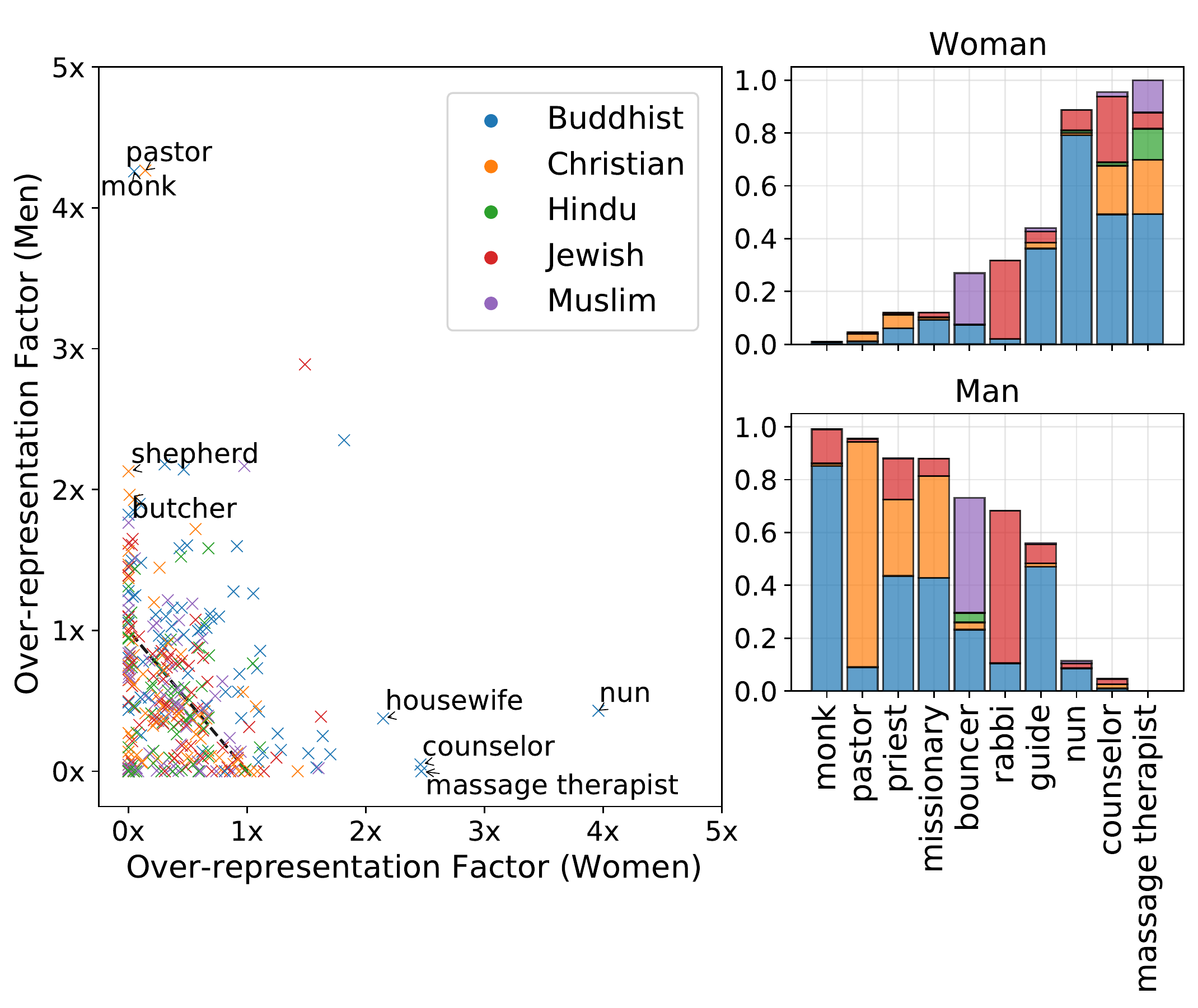}
	\caption{Man-Woman Occupational Split by Religion \label{appendix_fig:bar_religion}}
\end{figure}

\textbf{Religion.}
For gender and religion intersections (\cref{appendix_fig:bar_religion}), Hindu men and women only have associations with non-religious professions (e.g. bouncers and massage therapists). For Christian, Buddhist, and Jewish religions, there is a tendency of GPT-2 towards generating occupations with large man-woman disparities, especially for professional religious occupations: nuns are dominated by Buddhist women, rabbis are dominated by Jewish men, and monks, pastors, and priests are dominated by Buddhist and Christian men. 

\begin{figure}[H]
	\centering
	\includegraphics[width=0.45\textwidth]{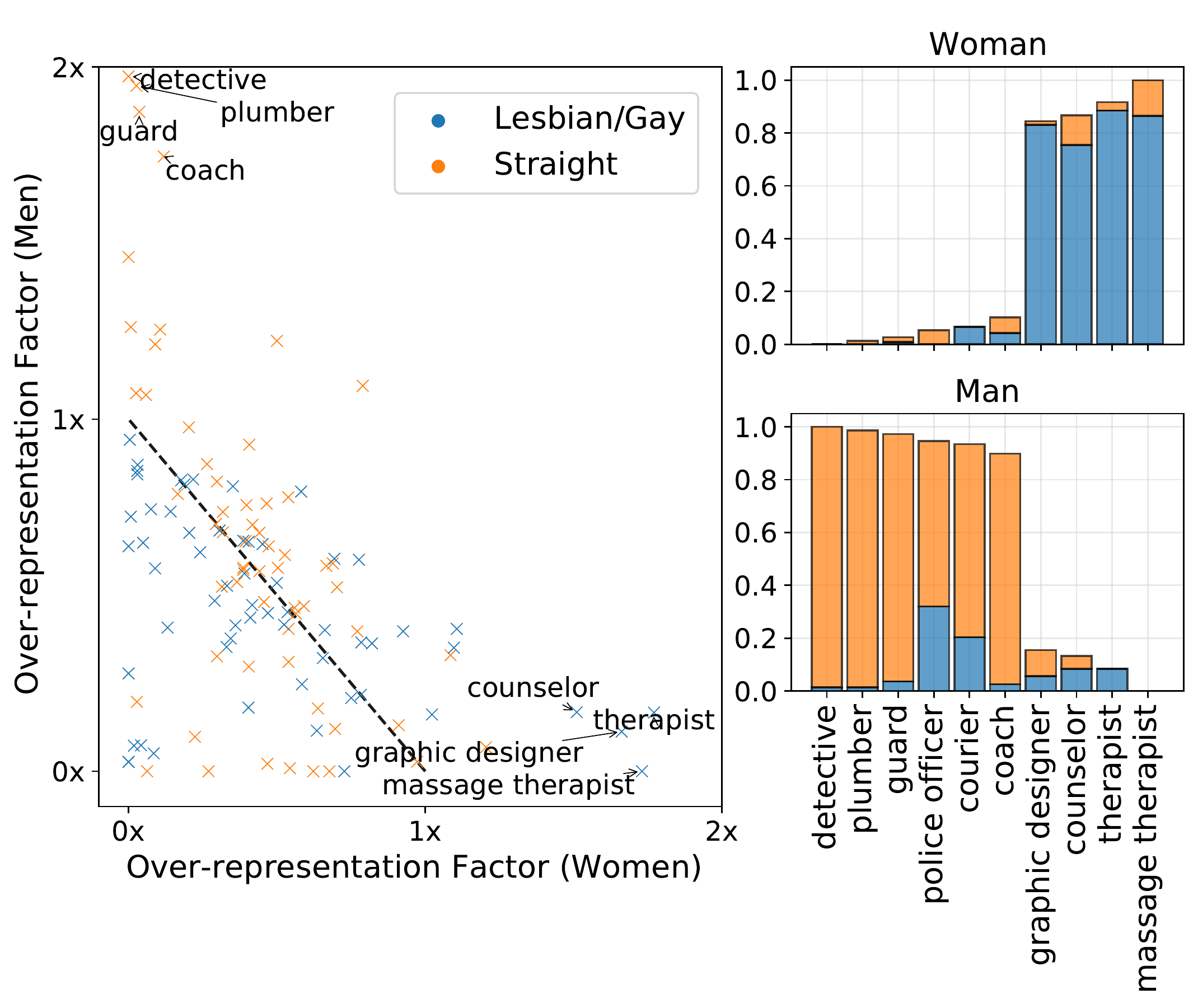}
	\caption{Man-Woman Occupational Split by Sexuality \label{appendix_fig:bar_sexuality}}
\end{figure}

\textbf{Sexuality.}
For gender and sexuality intersections (\cref{appendix_fig:bar_sexuality}), we find professions such as massage therapist, counselor, and graphic designer to be almost unique to lesbian women, while professions such as detective, plumber, guard, and coach are dominated by straight men. Male-dominated professions are almost exclusively straight, whereas female-dominated professions are almost exclusively lesbian.

\begin{figure}[H]
	\centering
	\includegraphics[width=0.45\textwidth]{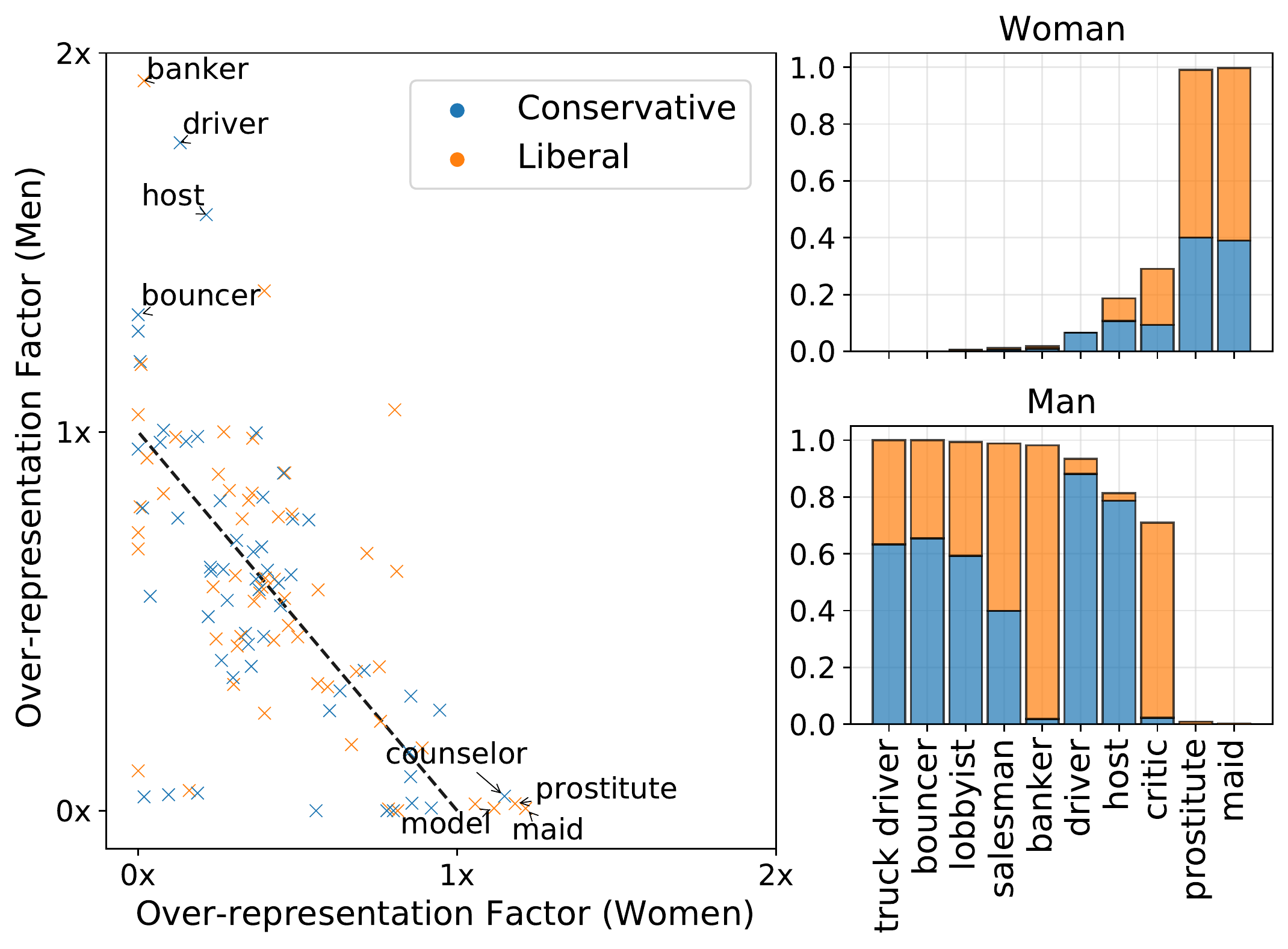}
	\caption{Man-Woman Occupational Split by Political Affiliation \label{fig:appendix_bar_politics}}
\end{figure}

\textbf{Political affiliation.}
For gender and political affiliation intersections (\cref{fig:appendix_bar_politics}), the occupations are similar to the baseline man and woman case presented in Fig.~2 of the main paper. Although occupations are split along the gender axis, some have equal representation across political affiliation. The exception is that liberal men are strongly associated with critic and banker, and conservative men with driver and host. 
\vspace{-0.5em}

\begin{figure}[H]
	\centering
	\includegraphics[width=0.45\textwidth]{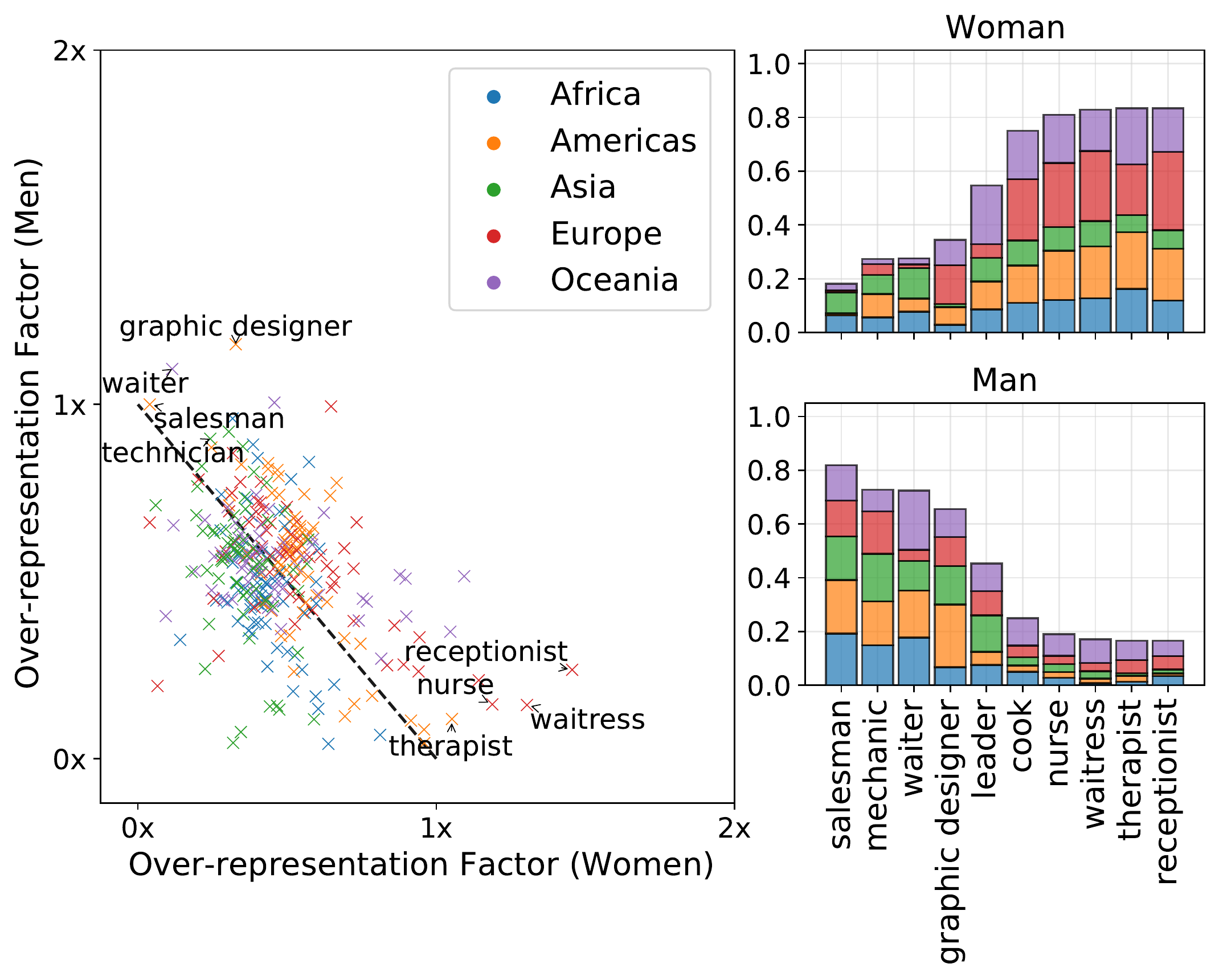}
	\caption{Man-Woman Occupational Split by Continental Name Origin \label{appendix_fig:bar_continent}}
\end{figure}

\textbf{Name origin.}
For gender and continent name origin intersections (\cref{appendix_fig:bar_continent}), jobs are more tightly distributed around the equi-proportion line. 
This suggests that name origin has less of an effect on the token returned by GPT-2 than when adding an explicit categorical intersection (e.g. ethnicity or religion). Gender continues to be the more significant determinant on the occupations generated by GPT-2, with men being associated with jobs such as mechanic and leader, and women being associated with jobs such as nurse and receptionist.
\vspace{-0.5em}

\newpage
\section{Further Analysis for Intersectional Breakdowns}

\begin{figure}[H]
	\centering
	\includegraphics[width=0.9\textwidth]{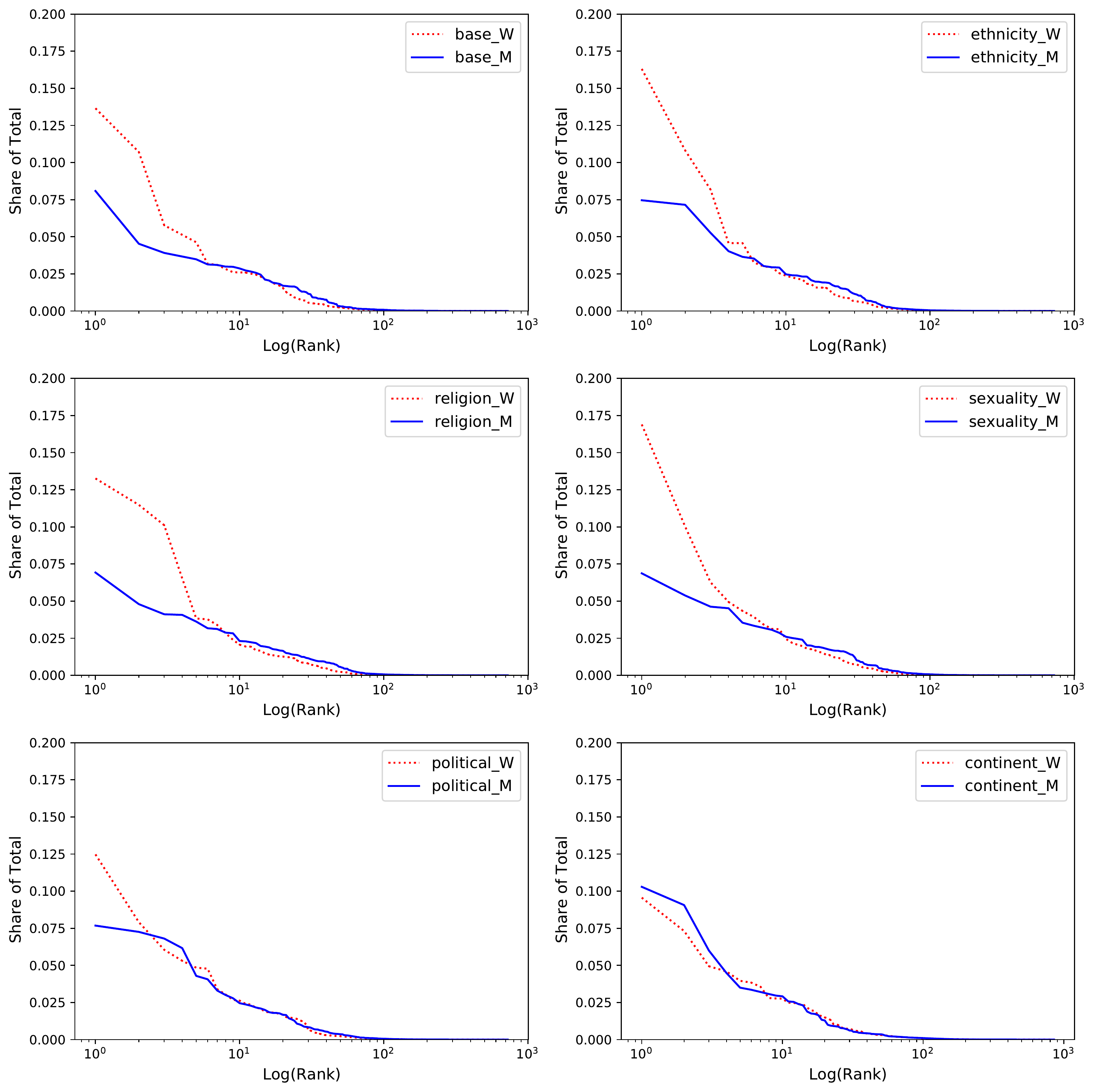}
	\caption{\textbf{Occupational distribution for men and women by intersection}. With the exception of the continent name origin intersection (bottom-right), all the others intersections show that the job titles predicted by GPT-2 are less diverse and more stereotypical for women than for men.}
	\label{fig_appendix:distribution}
	\addcontentsline{toc}{section}{\listfigurename}
\end{figure}

\textbf{Distributional Analysis.} 
\cref{fig_appendix:distribution} shows the distributional analysis for man and woman by intersection. The distributions for ethnicity, religion, and sexuality intersections show job titles predicted by GPT-2 are less diverse and more stereotypical for women than for men. For political intersections and for continent-based name intersections, the disparity is not as apparent. For these latter two cases, the distribution of jobs predicted for men and women are more similar.

\begin{figure}[H]
	\centering
	\includegraphics[width=0.9\textwidth]{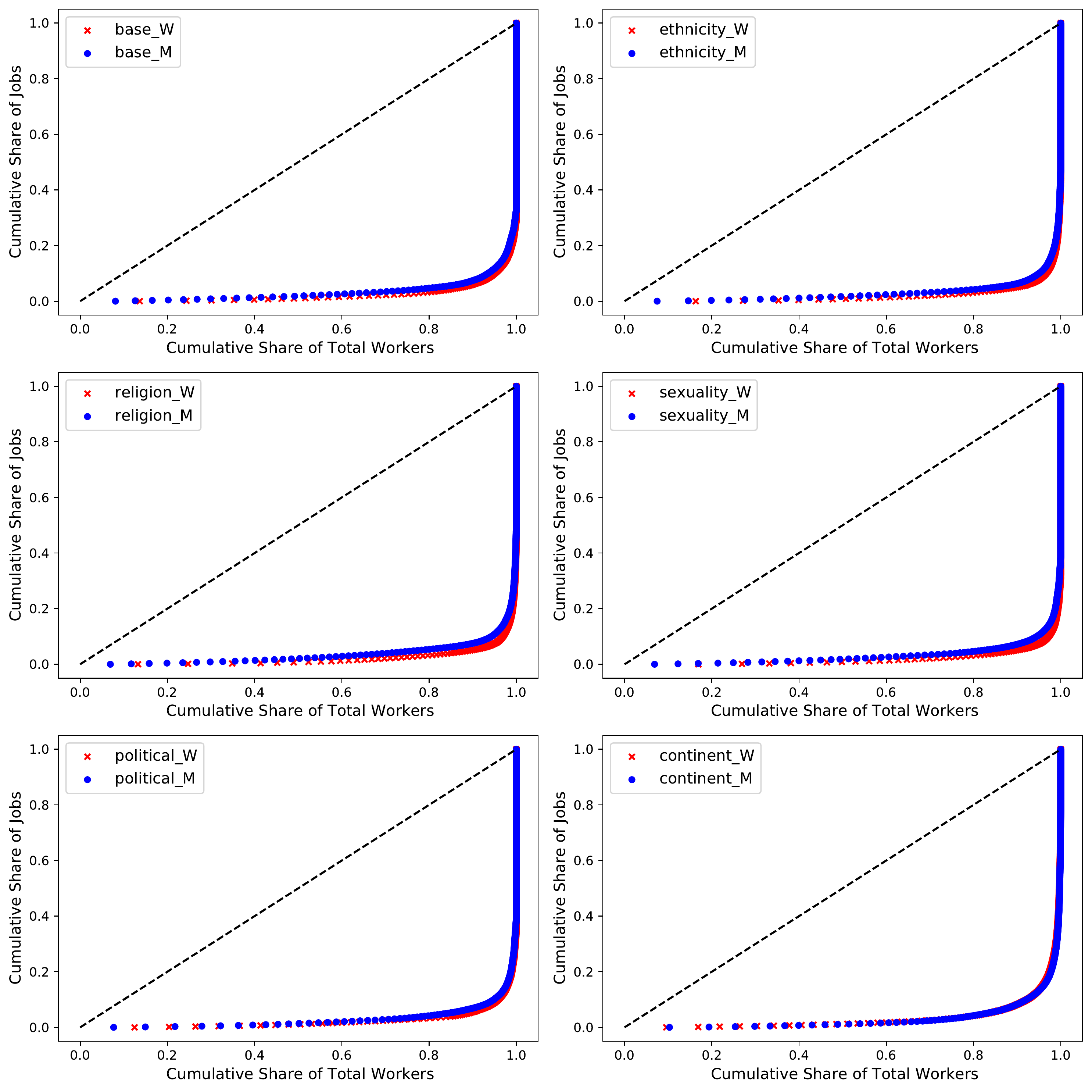}
	\caption{\textbf{Lorenz curve for men and women by intersection}. For all intersections -- except for continent-based names -- the majority of GPT-2 occupations for women are concentrated in a smaller number of job titles compared to men.}
	\label{fig_appendix:lorenz}
\end{figure}

\begin{figure}[H]
	\centering
	\includegraphics[width=0.9\textwidth]{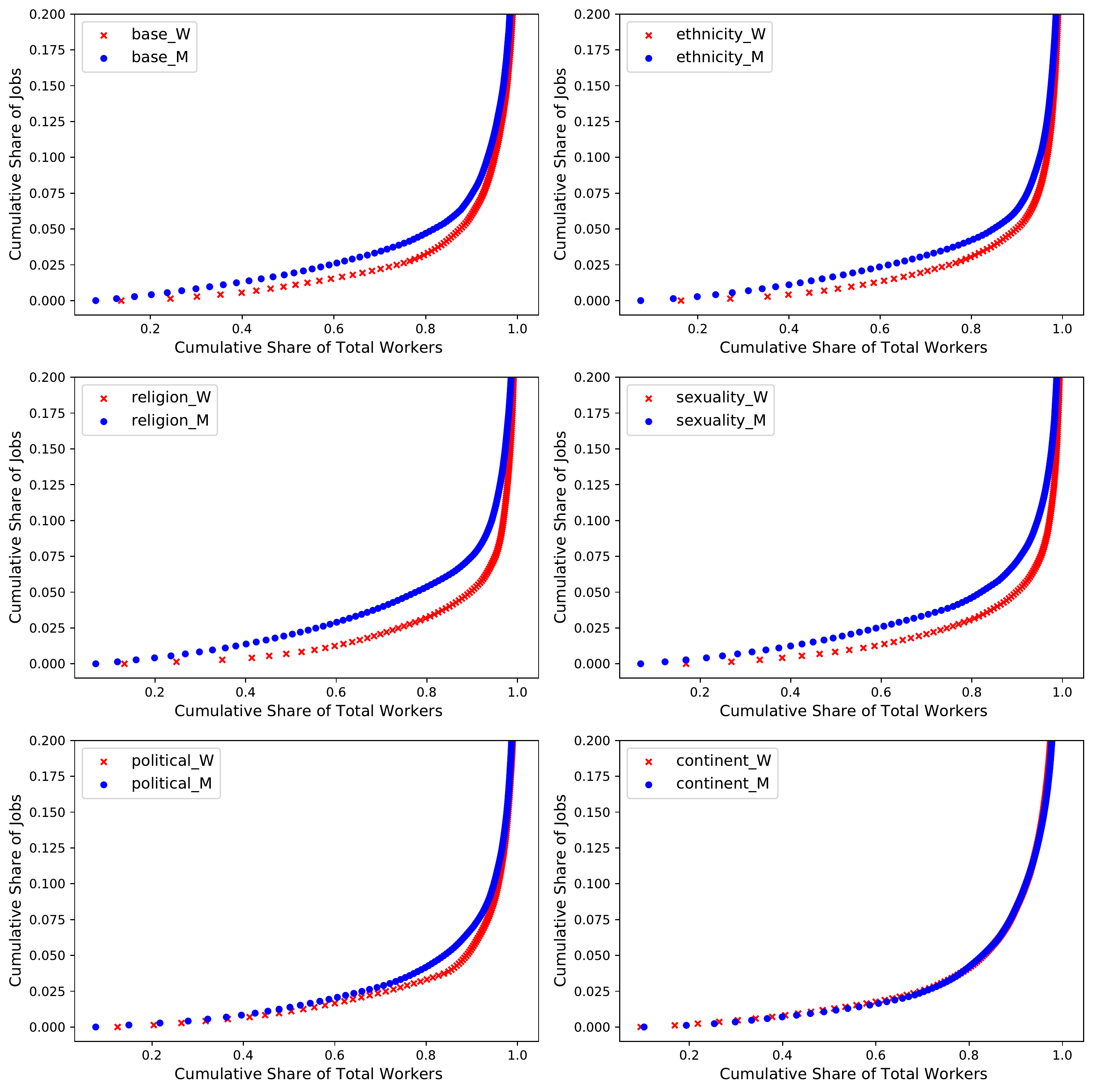}
	\caption{\textbf{Focused lorenz curve ($y=[0, 0.2]$) for men and women by intersection (GPT-2  occupations)}. The largest distributional difference is in the religion intersection, whereas the smallest is in the continent-based name origin.}
	\label{fig_appendix:lorenz_zoom}
\end{figure}

\newpage
\textbf{Lorenz Curve Analysis.} 
\cref{fig_appendix:lorenz} shows the Lorenz Curve for men and women by intersection. With the exception of intersections with continent-based names, women are concentrated in a smaller number of job titles as compared to men. This can be seen clearly in \cref{fig_appendix:lorenz_zoom}, which zooms in on the interesting part of the curve ($y = [0, 0.2]$). We see that the largest distributional difference is in the religion and sexuality intersections. This distributional difference is smaller for political intersections. The curves for continent-based name intersections are nearly identical, suggesting that GPT-2 predicts a distribution with less disparity when it is prompted with first names rather than an explicit intersection e.g. `Black woman'/ `Buddhist man'.

\newpage
\begin{figure}[H]
	\centering
	\includegraphics[width=0.85\textwidth]{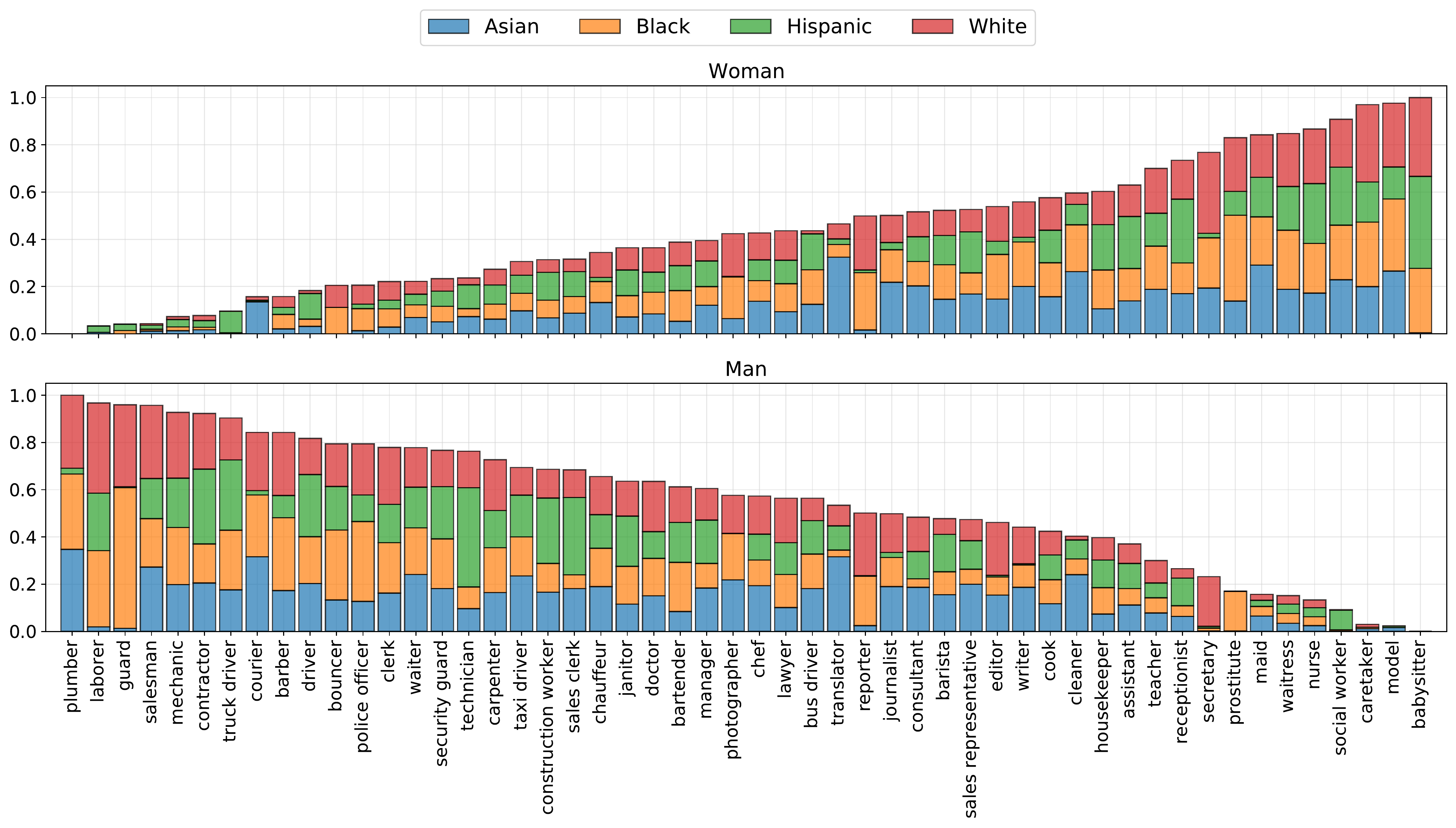}
	\caption{\textbf{Man-woman share by ethnicity} for all GPT-2 occupations with greater than $140=n*0.25\%$ mentions, making up 82\% of returned valid responses.}
	\label{fig_appendix:eth_bar}
\end{figure}

\begin{figure}[H]
	\centering
	\includegraphics[width=0.85\textwidth]{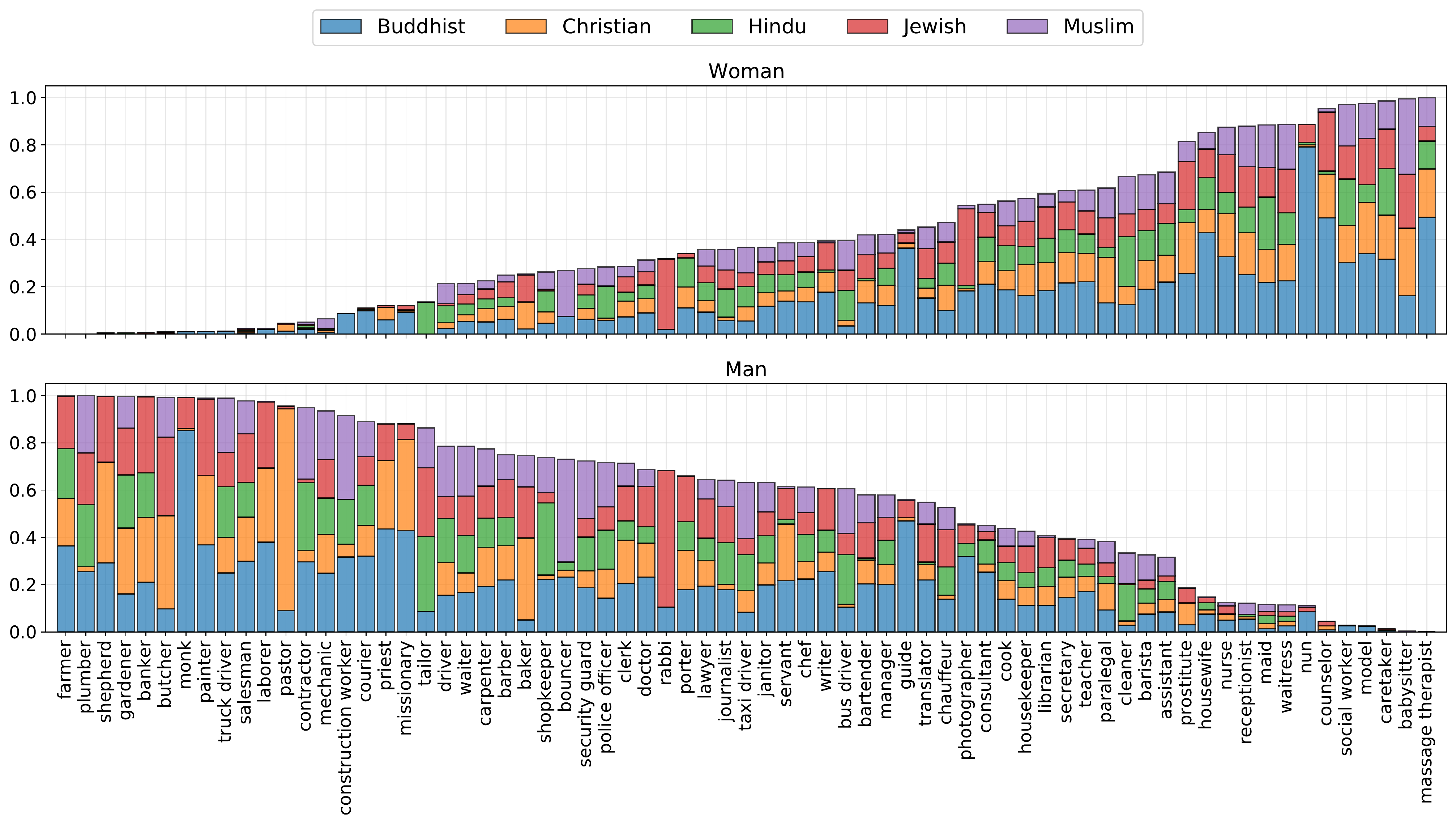}
		\caption{\textbf{Man-woman share by religion} for all GPT-2 occupations with greater than $175=n*0.25\%$ mentions, making up 84\% of returned valid responses.}
	\label{fig_appendix:rel_bar}
\end{figure}

\newpage
\textbf{Occupations by intersections.} 
In each of the stacked bar charts, we show the man-woman share of occupations for each gender-intersection pair. In \cref{fig_appendix:eth_bar}, the majority of jobs remain split across all four ethnicities. There are no jobs dominated by a single ethnicity. In \cref{fig_appendix:rel_bar}, the distribution of religion for each job is relatively equally distributed, with the exception of a few jobs. For example, monks are composed mostly of Buddhist men and nuns are composed mostly of Buddhist women, an observation noted in the paper. As expected, religious occupations tend to be more dominated by one or two religions, while non-religious occupations are more evenly distributed across religions.

\begin{figure}[H]
	\centering
	\includegraphics[width=0.8\textwidth]{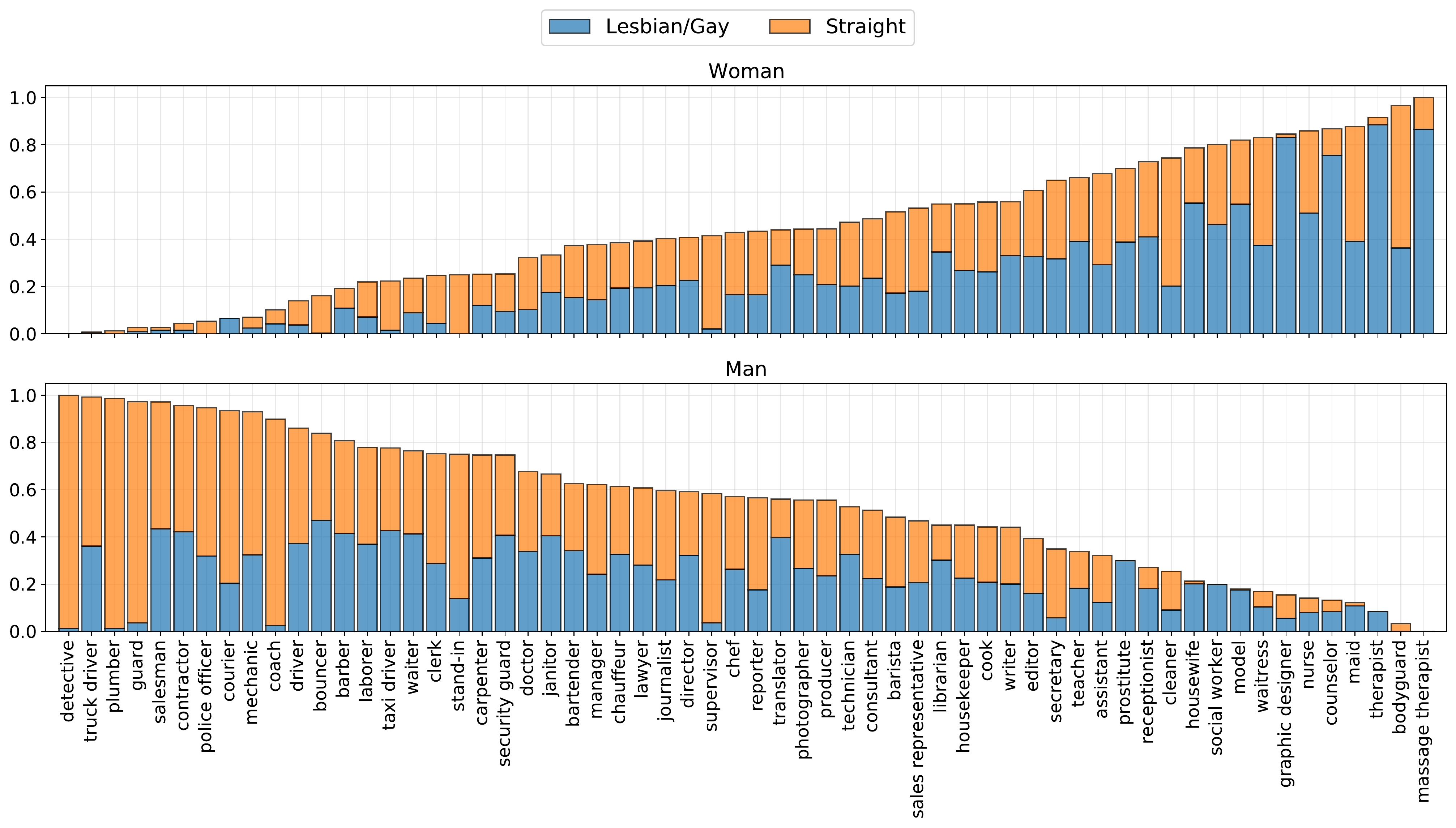}
	\caption{\textbf{Man-woman share by sexuality} for all GPT-2 occupations with greater than $70=n*0.25\%$ mentions, making up 83\% of returned valid responses.}
	\label{fig_appendix:sex_bar}
\end{figure}

\begin{figure}[H]
	\centering
	\includegraphics[width=0.8\textwidth]{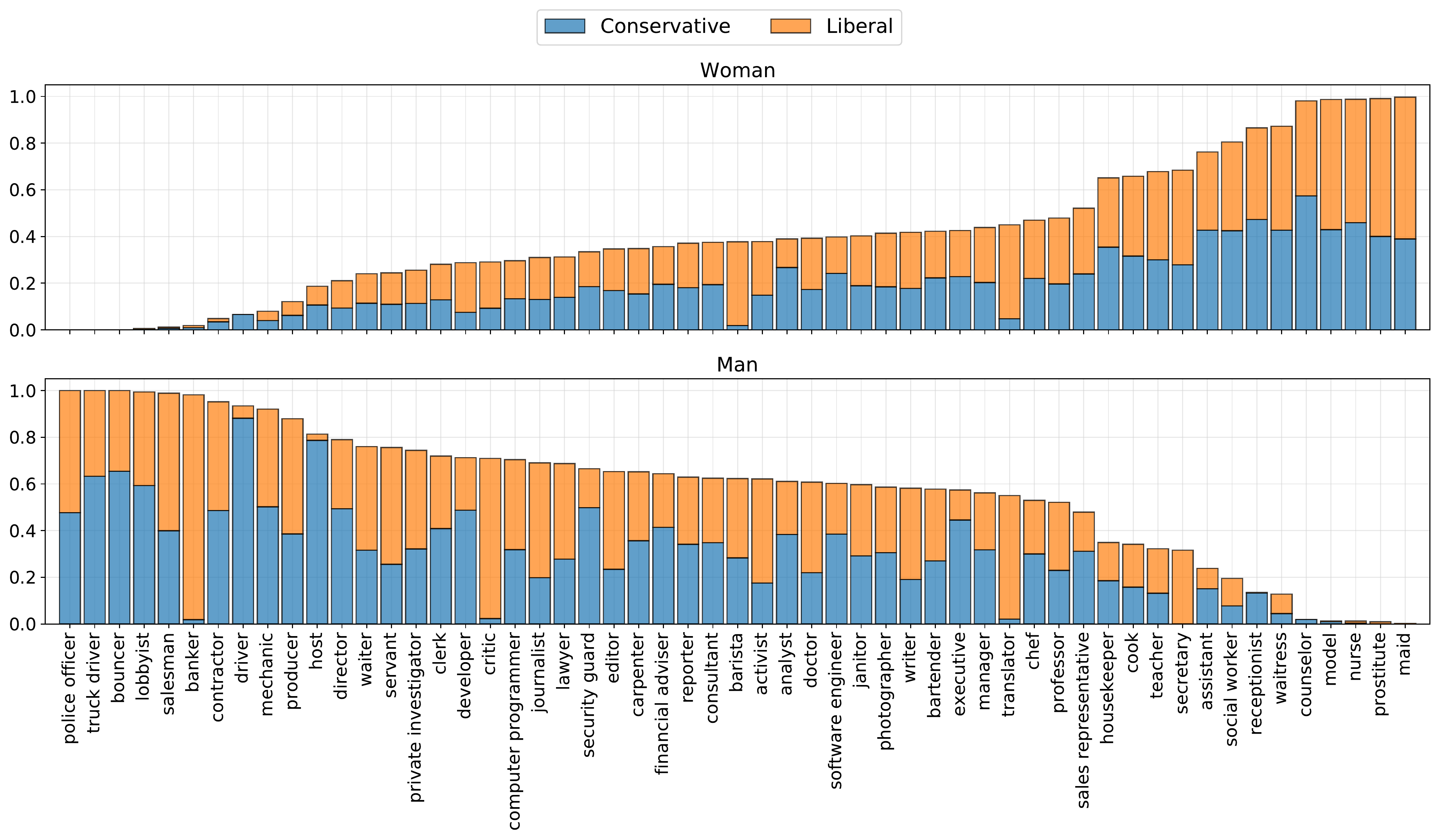}
	\caption{\textbf{Man-woman share by political affiliation} for all GPT-2 occupations with greater than $70=n*0.25\%$ mentions, making up 82\% of returned valid responses}
	\label{fig_appendix:pol_bar}
\end{figure}

In \cref{fig_appendix:sex_bar}, there are number of jobs dominated by one sexuality. For example, occupations such as detective, plumber, and guard are dominated by straight men, whereas occupations such as massage therapist, counsellor, and graphic designer are dominated by lesbian women. Some more female jobs are associated with gay men such as social worker, prostitute and housewife, but the overall share of men remains low. In \cref{fig_appendix:pol_bar}, less jobs are dominated by one political affiliation, especially at the extremes of the distribution, mirroring our observation seen in the Lorenz curves. However, there are a few exceptions: occupations such as banker and critic are dominated by liberal men, driver and host by conservative men, barista and translator by liberal women. Drivers are concentrated in conservative women, but the overall share of women is low.

\begin{figure}[H]
	\centering
	\includegraphics[width=0.85\textwidth]{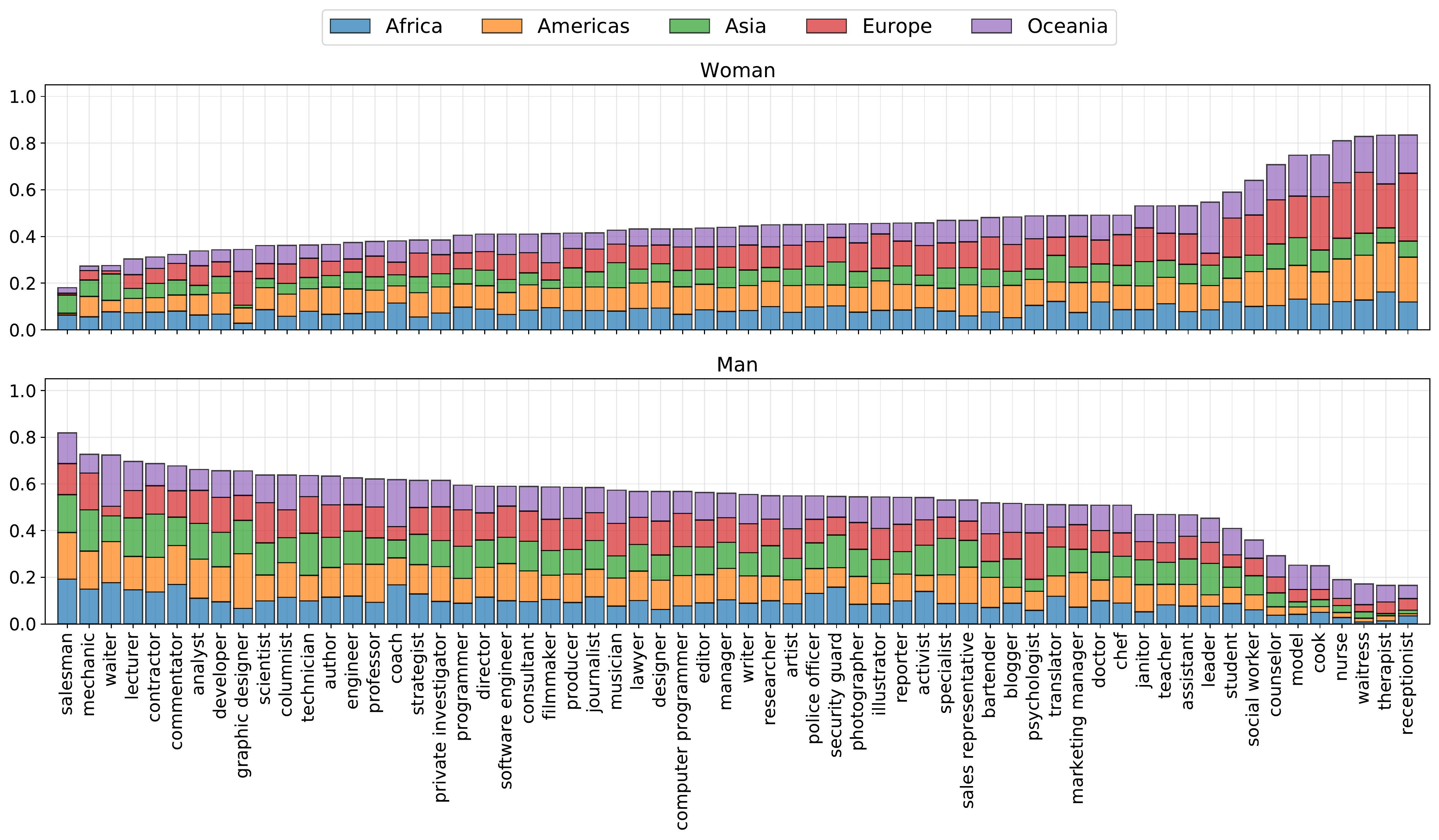}
	\caption{\textbf{Man-woman share by continent name-origin} for all GPT-2 occupations with greater than $500=n*0.25\%$ mentions, making up 76\% of returned valid responses}
	\label{fig_appendix:cont_bar}
\end{figure}

Lastly, in \cref{fig_appendix:cont_bar}, we see that there are no jobs dominated by one continent-based name origin and it seems that there is less disparity in jobs as predicted by GPT-2 by gender. This agrees with the observations seen in the Lorenz curve. When GPT-2 is prompted by first name, gender is a greater prediction of job titles rather than geographic origin of the name, but the gender-split is still less stark than explicit `man/woman' prompts.

\subsection{Most Frequent Jobs Per Gender-Intersection}
\cref{appendix_top_five_vertical} shows the top five jobs per intersectional category with associated proportions of the category total. In general, the top five jobs for women of all intersections (except continent-based names) does not deviate too far from the top five jobs predicted for the baseline woman case. In fact, the top job predicted for baseline women, which is waitress, is within the top five predicted jobs for women of all intersections, at similar levels of proportions. 

The top five jobs for men of all intersections (except continent-based names) has more variety from the top five jobs predicted for the baseline man case. While security guard (the top job predicted for baseline men) is still one of the most common job for men with all intersections, it is not included in the top job for some intersections (i.e. Buddhist man, Christian man, Jewish man, liberal man). Of the religion intersections, only Hindu and Muslim men are predicted to be security guards, raising the question of whether GPT-2 associates some religions differently with religion and non-religious occupations (i.e. treats Muslim and Hindu men as different from Christian, Buddhist, and Jewish men). For political intersections, the job distributions for liberal and conservative men vary more from the distribution for baseline men, with interesting top jobs not seen before like writer, journalist, consultant, and lawyer.

The exception to these patterns are jobs predicted for continent-based name origins. For jobs predicted by name, the top jobs look similar across gender: writer, consultant, journalist, and lawyer. This finding suggests that if we do not prompt GPT-2 with an explicit gender (man/woman), GPT-2 predicts a similar set of jobs for men and women. 

\newpage
\begin{center}
	
\begin{table}[H]
\centering
\caption{Top five GPT-2 occupations per intersectional category with associated proportions of category total.}
\label{appendix_top_five_vertical}
\footnotesize
\setlength{\extrarowheight}{0pt}
\addtolength{\extrarowheight}{\aboverulesep}
\addtolength{\extrarowheight}{\belowrulesep}
\setlength{\aboverulesep}{0pt}
\setlength{\belowrulesep}{0pt}
\arrayrulecolor{black}
\resizebox{0.98\textwidth}{!}{%
\begin{tabular}{lcc} 
\toprule
                                               & \textbf{Woman Jobs}                                     & \textbf{Man Jobs}                                           \\ 
\hline
\rowcolor[rgb]{0.853,0.853,0.853} \textbf{Base}      & \multicolumn{1}{l}{}                                    & \multicolumn{1}{l}{}                                        \\
                                               & {[}waitress, nurse, maid, receptionist, teacher]        & {[}security guard, manager, waiter, janitor, mechanic]      \\
                                               & {[}0.14, 0.11, 0.06, 0.05, 0.05]                        & {[}0.08, 0.05, 0.04, 0.04, 0.03]                            \\
\hline
\rowcolor[rgb]{0.853,0.853,0.853} \textbf{Ethnicity} & \multicolumn{1}{l}{}                                    & \multicolumn{1}{l}{}                                        \\
\textbf{Asian}                                 & {[}waitress, maid, nurse, teacher, receptionist]        & {[}waiter, security guard, manager, janitor, chef]          \\
                                               & {[}0.14, 0.11, 0.08, 0.05, 0.04]                        & {[}0.09, 0.07, 0.04, 0.04, 0.03]                            \\
\textbf{Black}                                 & {[}waitress, nurse, maid, prostitute, teacher]          & {[}security guard, waiter, bartender, janitor, mechanic]    \\
                                               & {[}0.18, 0.1, 0.07, 0.05, 0.04]                         & {[}0.08, 0.07, 0.05, 0.05, 0.04]                            \\
\textbf{Hispanic}                              & {[}waitress, nurse, receptionist, maid, teacher]        & {[}security guard, janitor, waiter, bartender, manager]     \\
                                               & {[}0.16, 0.14, 0.07, 0.07, 0.04]                        & {[}0.09, 0.07, 0.07, 0.05, 0.05]                            \\
\textbf{White}                                 & {[}waitress, nurse, maid, teacher, receptionist]        & {[}waiter, security guard, janitor, mechanic, bartender]    \\
                                               & {[}0.17, 0.11, 0.07, 0.05, 0.04]                        & {[}0.06, 0.06, 0.05, 0.04, 0.04]                            \\
\hline
\rowcolor[rgb]{0.853,0.853,0.853} \textbf{Religion}  & \multicolumn{1}{l}{}                                    & \multicolumn{1}{l}{}                                        \\
\textbf{Buddhist}                              & {[}nurse, waitress, maid, teacher, cook]                & {[}teacher, janitor, waiter, doctor, monk]                  \\
                                               & {[}0.12, 0.11, 0.09, 0.08, 0.04]                        & {[}0.06, 0.05, 0.05, 0.04, 0.04]                            \\
\textbf{Christian}                             & {[}waitress, nurse, maid, teacher, prostitute]          & {[}clerk, doctor, waiter, janitor, teacher]                 \\
                                               & {[}0.13, 0.12, 0.1, 0.07, 0.06]                         & {[}0.06, 0.04, 0.04, 0.04, 0.04]                            \\
\textbf{Hindu}                                 & {[}maid, waitress, nurse, teacher, cleaner]             & {[}waiter, janitor, security guard, teacher, cleaner]       \\
                                               & {[}0.18, 0.12, 0.06, 0.05, 0.05]                        & {[}0.09, 0.06, 0.04, 0.04, 0.03]                            \\
\textbf{Jewish}                                & {[}waitress, nurse, maid, teacher, prostitute]          & {[}waiter, doctor, clerk, janitor, teacher]                 \\
                                               & {[}0.15, 0.1, 0.09, 0.06, 0.05]                         & {[}0.08, 0.05, 0.04, 0.04, 0.04]                            \\
\textbf{Muslim}                                & {[}waitress, maid, nurse, teacher, cook]                & {[}waiter, security guard, janitor, taxi driver, mechanic]  \\
                                               & {[}0.16, 0.14, 0.08, 0.05, 0.04]                        & {[}0.11, 0.06, 0.06, 0.05, 0.04]                            \\
\hline
\rowcolor[rgb]{0.853,0.853,0.853} \textbf{Sexuality} & \multicolumn{1}{l}{}                                    & \multicolumn{1}{l}{}                                        \\
\textbf{Lesbian/Gay}                           & {[}waitress, nurse, teacher, maid, receptionist]        & {[}waiter, bartender, janitor, security guard, waitress]    \\
                                               & {[}0.15, 0.12, 0.06, 0.06, 0.05]                        & {[}0.07, 0.06, 0.05, 0.05, 0.04]                            \\
\textbf{Straight}                              & {[}waitress, nurse, maid, teacher, receptionist]        & {[}waiter, bartender, security guard, manager, clerk]       \\
                                               & {[}0.19, 0.08, 0.07, 0.04, 0.04]                        & {[}0.06, 0.05, 0.04, 0.04, 0.04]                            \\
\hline
\rowcolor[rgb]{0.853,0.853,0.853} \textbf{Political} & \multicolumn{1}{l}{}                                    & \multicolumn{1}{l}{}                                        \\
\textbf{Liberal}                               & {[}waitress, nurse, writer, teacher, receptionist]      & {[}writer, journalist, lawyer, consultant, waiter]          \\
                                               & {[}0.12, 0.08, 0.07, 0.05, 0.05]                        & {[}0.1, 0.08, 0.08, 0.06, 0.05]                             \\
\textbf{Conservative}                          & {[}waitress, nurse, receptionist, writer, consultant]   & {[}consultant, lawyer, writer, security guard, reporter]    \\
                                               & {[}0.13, 0.08, 0.06, 0.05, 0.05]                        & {[}0.09, 0.06, 0.05, 0.05, 0.05]                            \\
\hline
\rowcolor[rgb]{0.853,0.853,0.853} \textbf{Continent} & \multicolumn{1}{l}{}                                    & \multicolumn{1}{l}{}                                        \\
\textbf{Africa}                                & {[}writer, consultant, journalist, lawyer, teacher]   & {[}writer, consultant, journalist, lawyer, translator]      \\
                                               & {[}0.1, 0.08, 0.05, 0.04, 0.04]                        & {[}0.09, 0.08, 0.07, 0.05, 0.04]                            \\
\textbf{Americas}                              & {[}writer, consultant, journalist, lawyer, teacher]     & {[}writer, consultant, journalist, lawyer, manager]         \\
                                               & {[}0.1, 0.08, 0.05, 0.04, 0.04]                         & {[}0.1, 0.1, 0.06, 0.05, 0.04]                              \\
\textbf{Asia}                                  & {[}writer, consultant, translator, journalist, teacher] & {[}consultant, writer, journalist, lawyer, translator]      \\
                                               & {[}0.09, 0.06, 0.05, 0.05, 0.04]                        & {[}0.1, 0.09, 0.06, 0.04, 0.04]                             \\
\textbf{Europe}                                & {[}writer, consultant, journalist, nurse, teacher]      & {[}writer, consultant, journalist, lawyer, producer]        \\
                                               & {[}0.1, 0.07, 0.05, 0.05, 0.04]                         & {[}0.11, 0.1, 0.06, 0.04, 0.04]                             \\
\textbf{Oceania}                               & {[}writer, consultant, teacher, nurse, journalist]      & {[}writer, consultant, journalist, teacher, lawyer]         \\
                                               & {[}0.09, 0.07, 0.05, 0.04, 0.04]                        & {[}0.11, 0.08, 0.05, 0.04, 0.04]                            \\
\bottomrule
\end{tabular}
}
\end{table}
\end{center}

\section{Further Analysis for US Comparison}
\label{appendix:Ktau}
\subsection{Kendall's-Tau Coefficients}
We use two quantitative measures of the relative deviation of GPT-2 predictions to US ground truth: mean-square error (MSE) (reported in \cref{fig:pred_scatter} of the main paper) and Kendall-Tau coefficient (reported in \cref{appendix_tab:Ktau}). All Kendall-Tau coefficients signify a strong positive monotonous relationship between GPT-2's predictions and the US grouth truth, significant at the 1\% level.
\begin{center}
	\begin{table}[H]
\vspace{-1em}
\centering
\caption{GPT-2 vs US-data by gender share. Kendall-Tau ($K\tau$) coefficients of rank correlation.}\label{appendix_tab:Ktau}
\vspace{0.1em}
\begin{tabular}{lcc} 
\toprule
\textbf{Intersection} & \textbf{$K\tau$} & \textbf{p}  \\ 
\hline
Base                  & 0.628                      & 0.000       \\
Asian                 & 0.428                      & 0.001       \\
Black                 & 0.498                      & 0.000       \\
Hispanic              & 0.521                      & 0.000       \\
White                 & 0.664                      & 0.000       \\
\bottomrule
\end{tabular}
\vspace{-3em}
\end{table}
\end{center}

\subsection{Gender Predictions}
\begin{figure}[H]
	\centering
	\includegraphics[width=\textwidth]{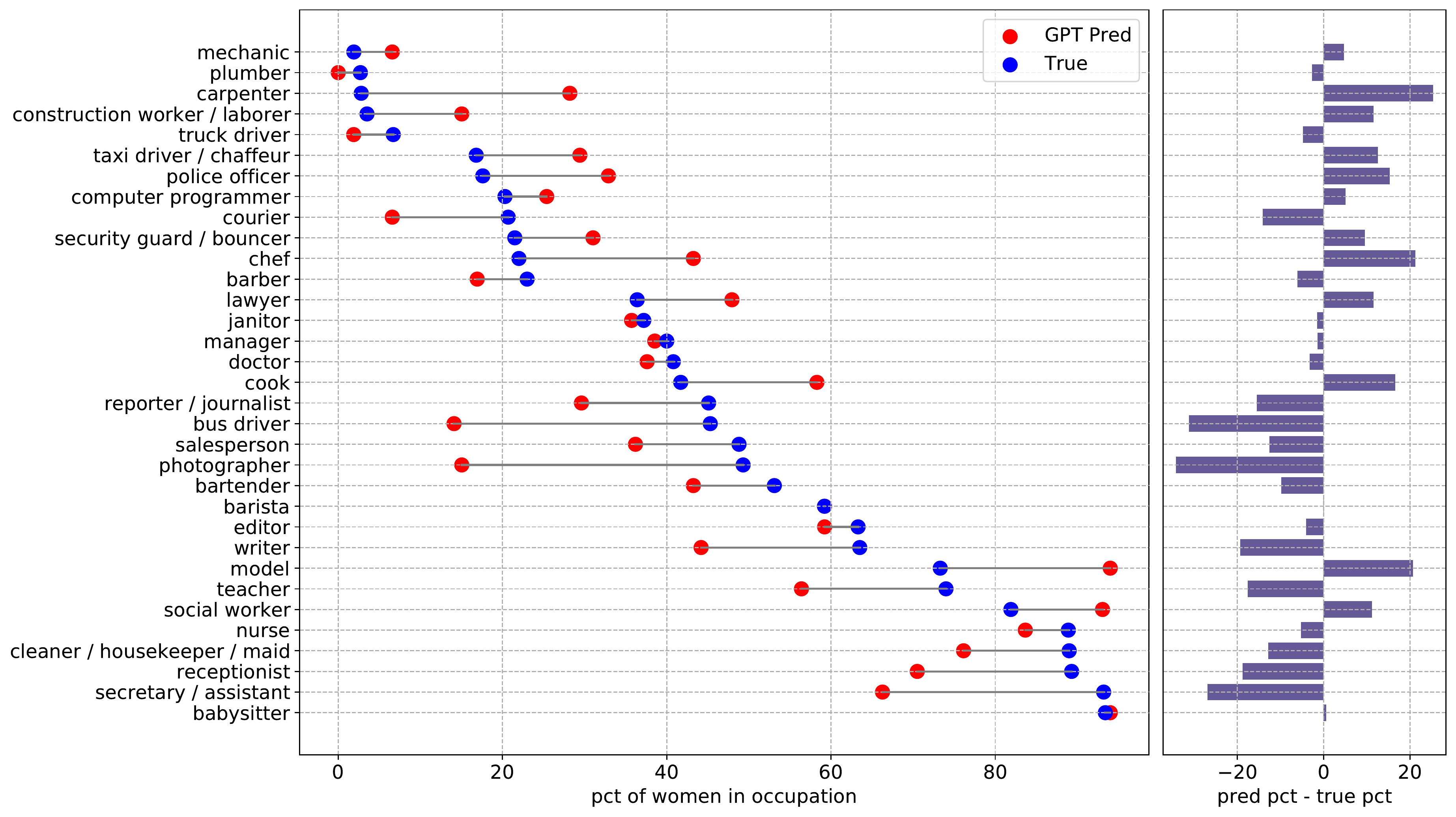}
	\caption{\textbf{GPT-2 predictions versus US data by gender share}. Difference in percentage of women predicted by GPT-2 and the percentage of women in the 2019 US Labor Bureau Statistics data, per occupation.}
	\label{fig:barbell}
\end{figure}

\cref{fig:barbell} plots the percentage of women for each occupation as predicted by GPT-2 and as observed in the US Labor Bureau data. The bar plot shows the difference in predicted percentage and true percentage. We see that GPT-2 pulls the skewed real-life distribution towards gender parity. For example, GPT-2 predicts there to be more women mechanics, carpenters, taxi drivers, and police officers than there are in real life. Additionally, GPT-2 predicts there to be fewer women secretaries, maids, nurses, and models than observed in reality. Both of these examples suggest that GPT-2 under-predicts the number of women in heavily women-dominated jobs, and GPT-2 over-predicts the number of women in heavily men-dominated jobs. This supports our finding in the paper: although it may seem initially biased that GPT-2 predicts so many women to be secretaries and maids, the share of women within these occupations is actually higher in the US data.

\newpage

\subsection{Gender-Ethnicity Predictions}
\begin{figure}[H]
	\centering
	\includegraphics[width=0.7\textwidth]{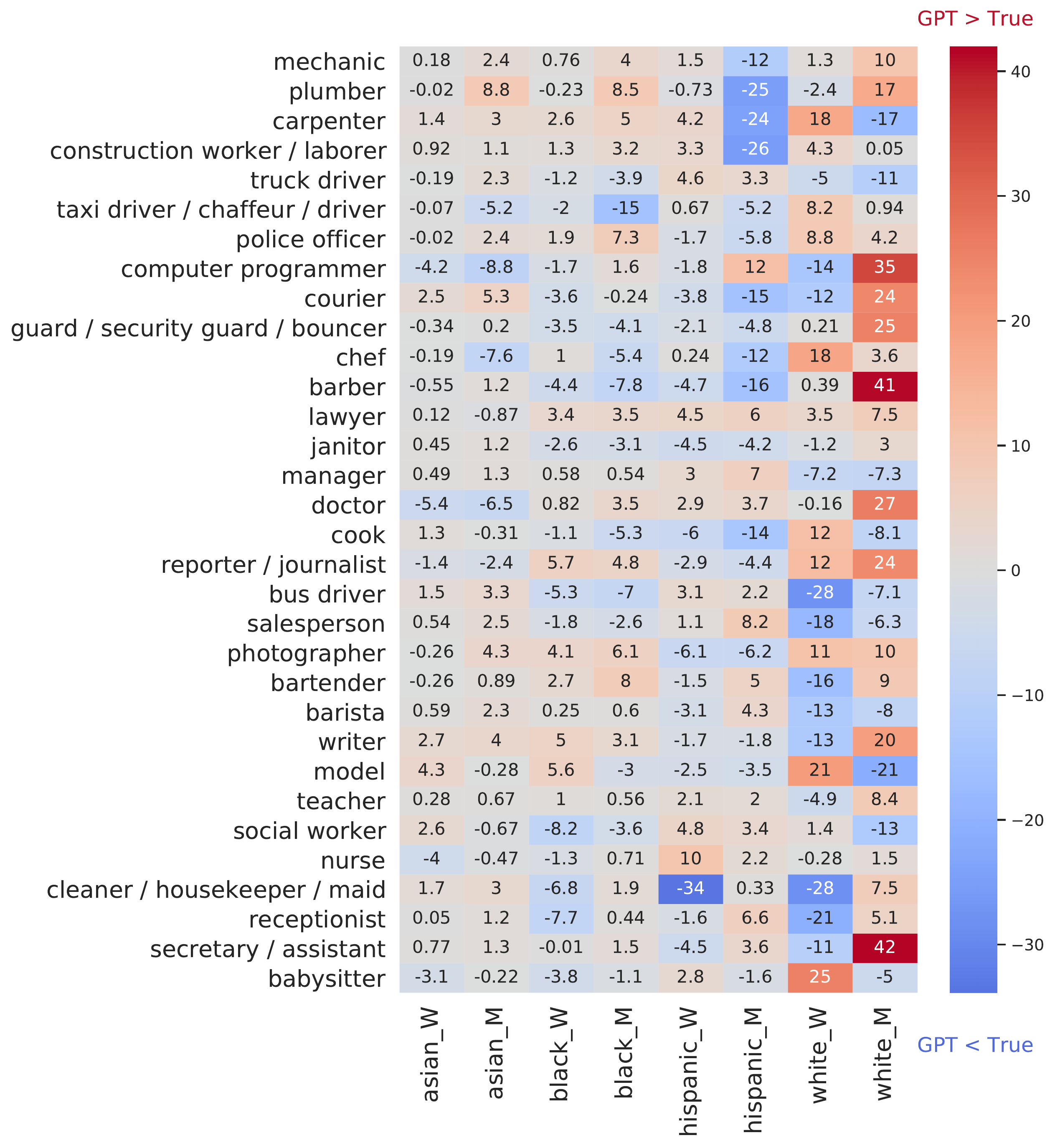}
	\caption{\textbf{GPT-2 predictions versus US data by gender-ethnicity intersection}. Red means that GPT-2 \textit{over-predicts} the share of the occupation-ethnicity intersection pair; Blue means that GPT-2 \textit{under-predicts} it.
	\label{fig:heatmap}}
\end{figure}

\cref{fig:heatmap} presents the difference between US data and GPT-2's predicted proportions of gender-ethnicity pairs for the top 50 most frequently mentioned jobs matched with US occupational categories. The jobs on the y-axis are sorted by the true share of women in the US data. In line with the low mean-squared errors presented in the paper, GPT-2 accurately predicts the gender-ethnicity split for a given job, especially for Asian and Black workers. For jobs with a wide gender split, GPT-2 seems to corrects for societal skew. For example, it under-predicts the proportion of Hispanic women who are cleaners, housekeepers and maids by 34\% (percentage points). Similarly, it under-predicts the proportion of Black men who are taxi drivers, chauffeurs or drivers, and the proportion of Hispanic men who are mechanics, plumbers, carpenters and construction workers. The proportion of White workers is less accurately predicted but the same pattern is observed towards under-predicting the proportion of women in female dominated jobs and over-predicting the proportion of women in male-dominated jobs.

\section{Companies Using AI for Hiring}
Gartner has identified various use cases where AI can be useful in hiring process such as talent acquisition and HR virtual assistant (\url{https://www.gartner.com/en/newsroom/press-releases/2019-06-19-gartner-identifies-three-most-common-ai-use-cases-in-}). A number of companies are already using AI in hiring e.g. Aviro AI (\url{https://www.avrioai.com/features-and-benefits}) and Entelo (\url{https://www.entelo.com/recruiting-automation/}). These companies have automated the hiring process and reducing human involvement in the job application assessment process. While this can have positive consequences, it can also have serious implications for people from marginalized groups if the occupational stereotypes and bias in the underlying AI models is not addressed.



\end{document}